\documentclass[12pt]{graphicsclass}
\usepackage[utf8]{inputenc}
\usepackage{palatino}
\usepackage{amsmath}
\usepackage{amssymb} 
\usepackage{graphicx}
\graphicspath{ {images/} }
\usepackage[labelfont=bf]{caption}
\usepackage{setspace}
\usepackage[margin=1in]{geometry}
\usepackage[hidelinks]{hyperref}
\usepackage{xcolor}
\usepackage{overpic}
\usepackage[all]{nowidow}

\usepackage{epigraph}

\newcommand{\myepigraph}[2]{\epigraph{\textit{#1}}{--- #2}}

\newcommand{\nonengepigraph}[3]{
  \epigraph{
    \begin{spacing}{1.25}
    \textit{#1}
    \end{spacing}
    \par\vspace{1ex}
    \textcolor{gray}{\textit{(#2)}}
    \vspace{2ex}
  }{--- #3}
}

\usepackage[printonlyused,withpage]{acronym}

\usepackage{tcolorbox}

\usepackage{lettrine}

\usepackage{fancyhdr}
\usepackage{booktabs}
\usepackage{natbib}

\usepackage{booktabs}
\usepackage{amsmath}
\usepackage{enumerate}
\usepackage{multirow}
\usepackage{bbding}
\usepackage{mathtools}
\usepackage{subfig}

\newlength\lengtha 
\newlength\lengthb 
\newlength\lengthc 

\DeclarePairedDelimiterX{\infdivx}[2]{(}{)}{%
  #1\;\delimsize\|\;#2%
}

\usepackage{graphicx}

\newcommand{\robertazs}[0]{RoBERTa\textsubscript{Zero-shot}}

\newcommand{\conceptmax}[0]{\textsc{ConceptMax}}
\newcommand{\conceptinject}[0]{\textsc{ConceptInject}}

\newcommand{\pep}[0]{\textsc{Pep-3k}}

\newcommand{\ccd}[0]{CC$\Delta$}
\newcommand{\ler}[0]{LER}
\newcommand{\mlp}[0]{GloVe+MLP}

\usepackage{amsmath}
\usepackage{amssymb}
\usepackage{booktabs}
\usepackage{subfig}
\usepackage{graphicx}
\usepackage{enumerate}
\usepackage[noabbrev]{cleveref}

\usepackage{booktabs}
\usepackage{amsmath}
\usepackage{hyperref}
\usepackage{graphicx}
\usepackage{subfig}
\usepackage{nicematrix,tikz}
\usepackage{stackengine}
\usepackage{authblk}
\usepackage{enumitem,kantlipsum}
\usepackage{arydshln}
\usepackage{diagbox}
\usepackage{multirow}
\usepackage{soul}
\usepackage{colortbl}

\definecolor{LightGoldenrod}{RGB}{255,236,139}
\definecolor{amber}{rgb}{1.0, 0.69, 0.4}
\definecolor{BabyPink}{rgb}{0.96, 0.76, 0.76}
\definecolor{LightBlue}{RGB}{173,216,230}
\definecolor{LightGreen}{RGB}{144,238,144}
\definecolor{cellgray}{gray}{0.8}

\DeclareRobustCommand{\hly}[1]{{\sethlcolor{amber}\hl{#1}}}

\DeclareRobustCommand{\hlb}[1]{{\sethlcolor{LightBlue}\hl{#1}}}
\DeclareRobustCommand{\ccell}[1]{{\setul{-2ex}{-2ex}\sethlcolor{cellgray}\hl{#1}}}
\DeclareRobustCommand{\hlg}[1]{{\sethlcolor{LightGreen}\hl{#1}}}

\newcommand{\para}[1]{\vspace{4pt}\noindent\textbf{#1.}~}

  {\list{}{\leftmargin=#1\rightmargin=#1}\item[]}%
  {\endlist}

\usepackage{tikz}
\usetikzlibrary{tikzmark}

\pgfkeys{/tikz/.cd,
      set fill color/.code={\renewcommand{\fillcol}{#1}},
      set border color/.code={\renewcommand{\bordercol}{#1}},
      left offset/.store in=\leftoff,
      right offset/.store in=\rightoff,
      above offset/.store in=\aboveoff,
      below offset/.store in=\belowoff,
      left offset=-0.02,
      right offset=0.12,
      above offset=0.325,
      below offset=-0.1,
      below right offset/.store in=\belowrightoff,
      below right offset={\rightoff,\belowoff},
      above left offset/.store in=\aboveleftoff,
      above left offset={\leftoff,\aboveoff},
}%

\usepackage{makecell}

\newcommand{\xia}{X\&V}

\usepackage{amssymb}%
\usepackage{pifont}%

\newtheorem{definition}{Definition}

\usepackage{float}

\usepackage[T1]{fontenc}

\title{
\linespread{0.5}

{\baselineskip=1.5em
The Validity of Coreference-based Evaluations of Natural Language Understanding
\\~\\[2.5em]
}

{\large
Ian Porada \\[1.5em]
Doctor of Philosophy \\[1.5em]
November 2025 \\[1.5em]
School of Computer Science \\[2.5pt]
McGill University \\[2.5pt]
Montreal, Quebec \\[1em]

\begin{figure}[h]
    \centering
    \includegraphics[width=5.5cm]{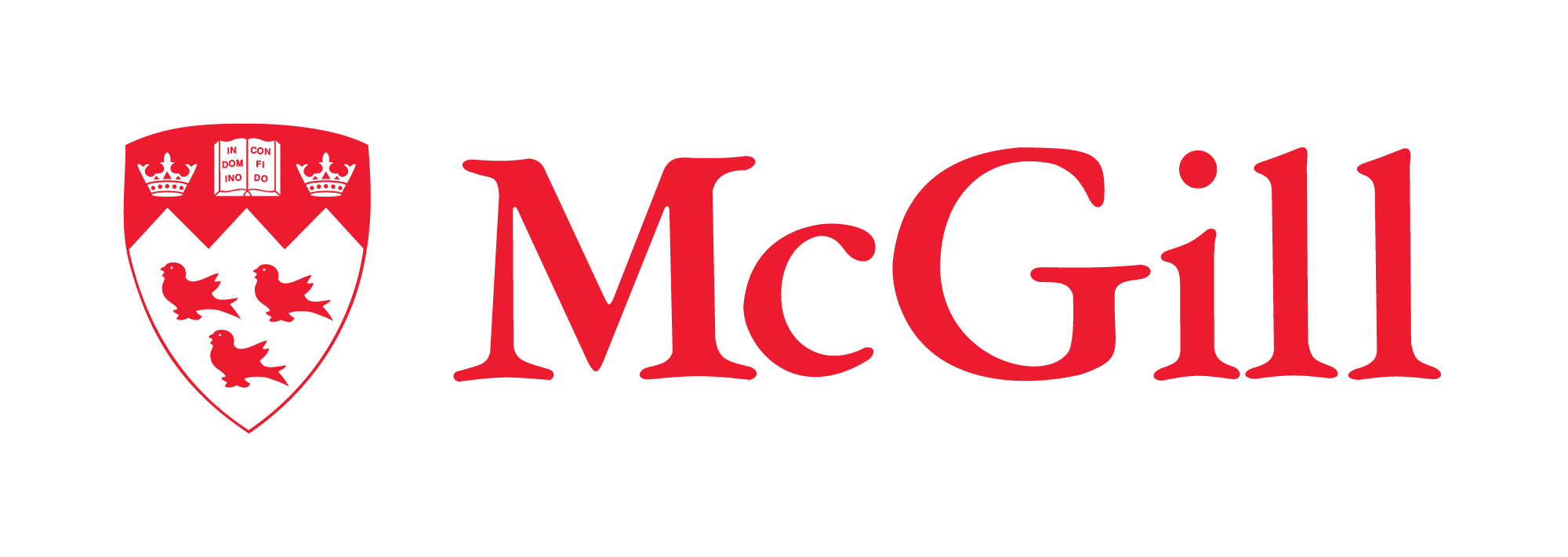}
\end{figure}

\textit{A thesis submitted to McGill University in partial fulfillment of the \\[2.5pt] requirements of the degree of Doctor of Philosophy in Computer Science} \\
}
}

\author{\textcopyright \ Ian Porada 2025}
\date{}

\begin{document}
 \maketitle

\newpage

\chapter*{Defense Committee}

{
\begin{center}
\setlength{\arrayrulewidth}{0.5mm}
\setlength{\tabcolsep}{20pt}
\renewcommand{\arraystretch}{0.65}
\begin{tabular}{rl}
 Jackie Chi Kit Cheung, Ph.D. & \ School of Computer Science,\\
\emph{Supervisor} & \ McGill University \\[2em]

Brendan Gillon, Ph.D. & \ Department of Linguistics,\\
\emph{Internal Examiner} & \ McGill University \\[2em]

Golnoosh Farnadi, Ph.D. & \ School of Computer Science,\\
\emph{Internal Member} & \ McGill University \\[2em]

Vincent Ng, Ph.D. & \ Department of Computer Science,\\
\emph{External Examiner} & \ The University of Texas at Dallas %
\end{tabular}
\end{center}
}

\newpage
\topskip0pt
\vspace*{\fill}
\begin{center}
    \vspace{19em}
    \textcopyright \ Copyright 2025 \\[-0.5em]
    by \\[-0.5em]
    Ian Porada \\[19em]

    \textit{Where noted, certain materials have been reproduced from the Association for Computational Linguistics (ACL) Anthology under the Creative Commons 3.0 BY-NC-SA license.}
\end{center}
\vspace*{\fill}

\newpage
\topskip0pt
\vspace*{\fill}
\begin{center}
    \textit{To my family.}
\end{center}
\vspace*{\fill}

\chapter*{Abstract}
\label{sec:engAbstract}
\addcontentsline{toc}{section}{\nameref{sec:engAbstract}}

An aspirational goal of the field of natural language processing (NLP) is to develop computer systems that can perform tasks that require understanding natural language.
Imagine, for instance, a computer program that could seamlessly execute users’ English-language instructions, even when they involve complex reasoning across long, nuanced contexts.
To make progress towards this goal, we as a field need methods for evaluating systems' natural language understanding capabilities.
Since at least the 1960s, one approach has been to evaluate systems based on a complex but fundamental component of understanding natural language: the ability to resolve coreferences.
In brief, coreference is the phenomenon of multiple linguistic expressions referring to the same entity in the world---e.g., a name and a pronoun that refer to the same person, as in the sentence ``\textit{Robert} woke up at 9 a.m. before \textit{he} went to school.''
Despite the existence of numerous evaluations of systems' ability to resolve coreferences, however, what broader conclusions can be realistically drawn from these evaluations are not clear.
Popular benchmarks such as OntoNotes and its derivatives offer large-scale, standardized tests but suffer from issues of measurement validity, making it difficult to know what conclusions about systems' capabilities truly generalize.

In this thesis, I refine our understanding as to what conclusions we can reach from coreference-based evaluations by expanding existing evaluation practices and considering the extent to which evaluation results are either converging or conflicting.
First, I analyze standard coreference evaluations and show that their design often leads to non-generalizable conclusions due to issues of measurement validity—including contestedness (multiple, competing definitions of coreference) and convergent validity (evaluation results that rank models differently across benchmarks).
Second, I propose and implement a novel evaluation focused on testing systems’ ability to infer the relative plausibility of events, a key aspect of resolving coreference.
Through this extended evaluation, I find that contemporary language models demonstrate strong performance on standard benchmarks---improving over earlier baseline systems within certain domains and types of coreference---but remain sensitive to the evaluation conditions: they often fail to generalize in ways one would expect a human to be capable of when evaluation contexts are slightly modified.
Taken together, these findings clarify both the strengths, such as improved accuracy over baselines on widely used evaluations, and the limitations of the current NLP paradigm, including weaknesses in measurement validity, and suggest directions for future work in developing better evaluation methods and more genuinely generalizable systems.

\chapter*{Acknowledgements}
\label{sec:ded}
\addcontentsline{toc}{section}{\nameref{sec:ded}}

Firstly, I am grateful to my advisor, Jackie CK Cheung, for the years of guidance. Jackie has a talent for motivating and framing research problems that I can only describe as a superpower. In combination with his considerate patience (maybe his second superpower), this made for a supportive and ambitious environment without which I am not sure I could have finished my thesis.

I am also grateful to the additional members of my supervisory committee, Timothy J. O'Donnell and Siva Reddy, for their essential feedback on my research proposal. It was incredibly helpful to get their outside feedback periodically throughout the years.

The initial direction of this thesis started from a collaboration that further included Ali Emami, Paul Trichelair, Adam Trischler, and Kaheer Suleman. I am especially indebted to Ali who taught me many early-career research skills when I first joined the lab.

I also had many useful collaborations along this journey. Alessandro Sordoni provided necessary mentorship and ideas for experiments related to evaluating language models, but more importantly helped me learn to code more quickly. Alexandra Olteanu was instrumental in connecting the work to measurement theory, and I'm especially thankful to her for helping guide my research direction. Lastly, Xiyuan Zou was a great collaborator helping to implement models for the re-evaluation experiments. I will miss being able to discuss ideas with these three.

During my studies, I was lucky to be supported by a fellowship from the Fonds de recherche du Québec (FRQ). All experiments were conducted with compute resources provided by Mila, Microsoft, and the Digital Research Alliance of Canada (DRAC). I am incredibly grateful for the technical support provided by the Mila and DRAC teams, in particular Olexa Bilaniuk who is a computing wizard.

Finally, I would like to thank my colleagues for useful discussions that directly contributed to this research: Rahul Aralikatte, Akshatha Arodi, Kushal Arora, Ines Arous, Yu Bai, Meng (Caden) Cao, Aishik Chakraborty, Ziling Cheng, Andre Cianflone, Bonaventure Dossou, Jules Gagnon-Marchand, Sitao Luan, Yu Lu Liu, Cesare Spinoso-Di Piano, Martin Pömsl, Abhilasha Ravichander, Andrei Romascanu, Mingde (Harry) Zhao, and Chenxuan (Tracy) Zhou---in addition to all the members of McGill NLP and Mila. My tenure as a student spanned the COVID-19 pandemic, but these individuals helped keep the research going even when the university buildings were closed. I am excited to see all of the great things they will accomplish. 

\chapter*{Statement of Contribution}
\label{sec:contributions}
\addcontentsline{toc}{section}{\nameref{sec:contributions}}

The core experiments and findings presented in this thesis were published in refereed conference proceedings, although these results are uniquely restructured and presented in this thesis as a coherent body of work. For all of the respective publications other than \citet{porada-etal-2024-controlled}, I am the sole first author and only student author. \citet{porada-etal-2024-controlled} was prepared in collaboration with Xiyuan Zou with detailed contributions outlined below. The respective publications are as follows:

\defcitealias{porada-cheung-2024-solving}{Solving the Challenge Set without Solving the Task: On Winograd Schemas as Test of Pronominal Coreference Resolution. In \textit{Proceedings of the 28th Conference on Computational Natural Language Learning (CoNLL 2024)}. Ian Porada and Jackie CK Cheung.}

\defcitealias{porada-etal-2024-challenges}{Challenges to Evaluating the Generalization of Coreference Resolution Models: A Measurement Modeling Perspective. In \textit{Findings of the Association for Computational Linguistics: ACL 2024}. Ian Porada, Alexandra Olteanu, Kaheer Suleman, Adam Trischler, and Jackie CK Cheung.}

\defcitealias{porada-etal-2024-controlled}{A Controlled Reevaluation of Coreference Resolution Models. In \textit{Proceedings of the 2024 Joint International Conference on Computational Linguistics, Language Resources and Evaluation (LREC-COLING 2024)}. Ian Porada, Xiyuan Zou, and Jackie CK Cheung.}

\defcitealias{porada-etal-2021-modeling}{Modeling Event Plausibility with Consistent Conceptual Abstraction. In \textit{Proceedings of the 2021 Conference of the North American Chapter of the Association for Computational Linguistics: Human Language Technologies (NAACL-HLT 2021)}. Ian Porada, Kaheer Suleman, Adam Trischler, and Jackie CK Cheung.}

\defcitealias{porada-etal-2022-pre}{Does Pre-training Induce Systematic Inference? How Masked Language Models Acquire Commonsense Knowledge. In \textit{Proceedings of the 2022 Conference of the North American Chapter of the Association for Computational Linguistics: Human Language Technologies (NAACL-HLT 2022)}. Ian Porada, Alessandro Sordoni, and Jackie CK Cheung.}

\begin{itemize}
    \item \citetalias{porada-cheung-2024-solving}
    \item \citetalias{porada-etal-2024-challenges}
    \item \citetalias{porada-etal-2024-controlled}
    \item \citetalias{porada-etal-2021-modeling}
\end{itemize}

\section*{Contribution to original knowledge}

To summarize the contributions that this thesis makes to the existing body of knowledge in the field, these contributions are as follows:

\begin{enumerate}
    \item I make a novel connection between coreference-based evaluations of NLP systems to the framework of measurement validity, and, based on this connection, highlights limitations in the validity of existing coreference-based evaluations. Namely, definitions of coreference being inconsistent and results being contradictory.
    \item I present a novel evaluation of the consistency of systems' ability to consistently infer the semantic plausibility of events. This reveals the original conclusion that language-model-based systems produce inconsistent inferences across a semantic hierarchy. 
\end{enumerate}

\section*{Contribution of authors}
\label{sec:contribution-of-authors}

All chapters in this thesis were originally drafted by myself and edited based on feedback and suggestions from my advisor and the additional authors of the aforementioned publications.

All experiments were implemented by myself with the following exception: experiments in Chapter~\ref{chap:supervised-comparison} were implemented jointly by myself and Xiyuan Zou. I wrote an initial framework based on which Zou implemented the encoder-based architectures, and I implemented the decoder-based architecture. The evaluations were jointly implemented by both of us, and we have equal-contribution authorship on the final publication which included results of experiments presented in this chapter.

\tableofcontents
\listoffigures %
\addcontentsline{toc}{section}{\listfigurename}
\listoftables
\addcontentsline{toc}{section}{\listtablename}

\newcommand{\listacronymname}{List of Acronyms}
\chapter*{\listacronymname}
\begin{acronym}[AMR]
 \acro{NLP}{natural language processing}
 \acro{CR}{coreference resolution}
 \acro{WSC}{Winograd Schema Challenge}
 \acro{MM}{Measurement Model}
 \acro{MLE}{maximum likelihood estimation}
 \acro{MDL}{Minimum Description Length}
 \acro{LDA}{Latent Dirichlet Allocation}
 \acro{PCR}{pronominal coreference resolution}
 \acro{CCD}[\ccd{}]{Concavity Delta}
 \acro{LER}[\ler{}]{Local Extremum Rate}
 \acro{PMI}{pointwise mutual information}
 \acro{MRR}{mean reciprocal rank}
 \acro{AMR}{Abstract Meaning Representation}

\end{acronym}
\addcontentsline{toc}{section}{\listacronymname}

\newpage 
\pagenumbering{arabic} %

\chapter{Introduction}
\label{chap:introduction}

\nonengepigraph{Wenn man in der gewöhnlichen Weise Worte gebraucht, so ist das, wovon man sprechen will, deren Bedeutung.}{If words are used in the ordinary way, what one intends to speak of is their reference.}{\citet{frege1892sense}}

\lettrine{T}{he} concept \textit{natural language} can be thought of as an abstraction used to describe the systems, such as English and French, that humans employ in the processes of thinking and communication. Importantly, all natural languages are believed to incorporate discrete, symbolic units \citep{hockett1960origin,fromkin2017introduction}. That is to say, any utterance in natural language can be associated with a fixed set of distinguishable, abstract elements such as phonemes (discrete units of sound; e.g., in English, individual consonant and vowel sounds) and morphemes (discrete units of syntax; e.g., in English, roots, prefixes, and suffixes) \citep{twaddell1935defining,nida1948identification}. These discrete units are \textit{realized} in the physical instantiation of natural language, e.g., as sounds articulated by the larynx, written marks on paper, or pixels on a screen. From a computer science perspective, the relationship between these abstract units and their physical instantiation is analogous to the object-oriented programming concept of instantiation of an abstract class: phonemes and morphemes can be alluded to an abstract specification, while the raw, physical form can be taken to be the instantiation of these discrete units in a specific modality.

\begin{figure}[t]
    \centering
    \includegraphics[width=8cm]{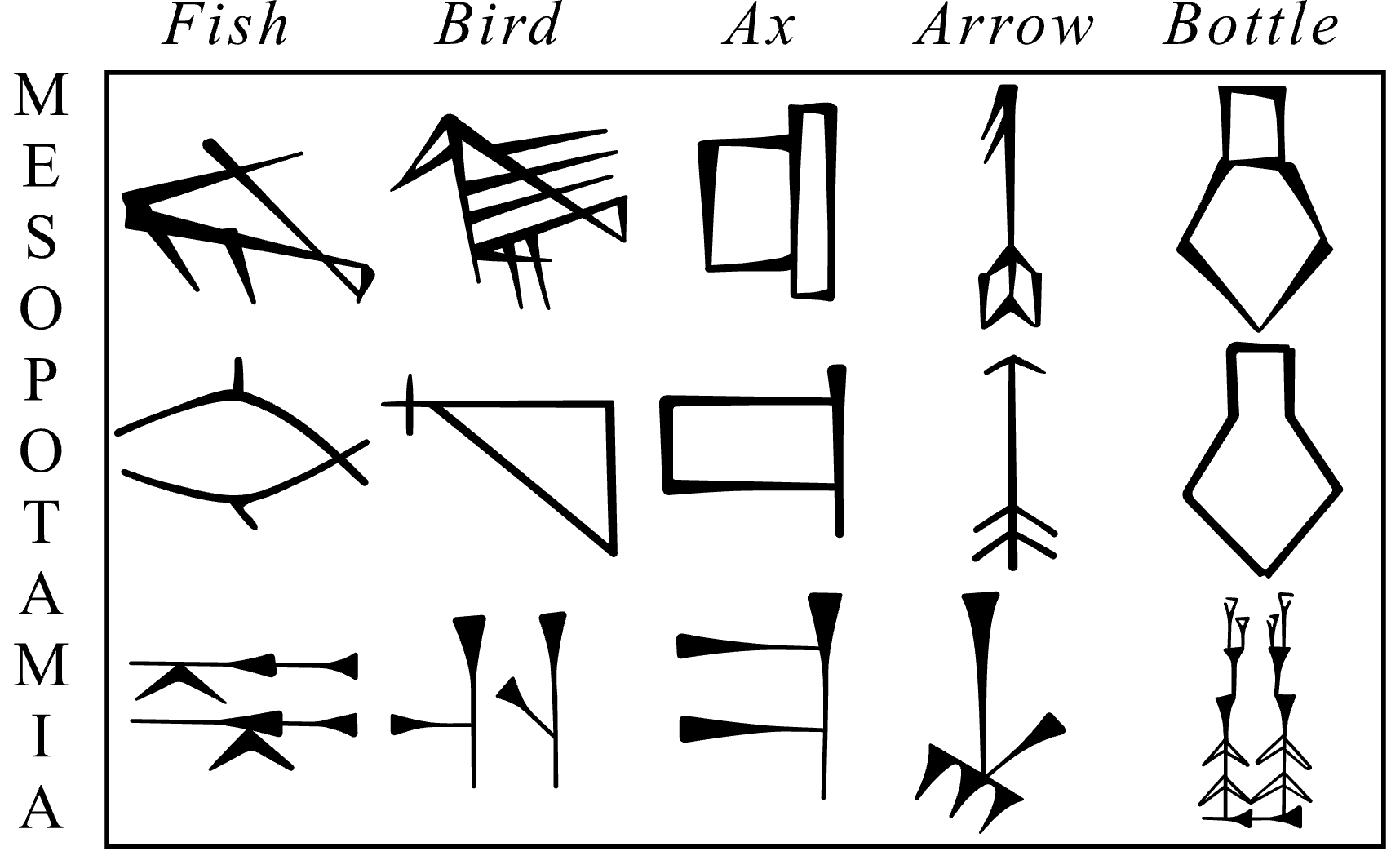}
    \caption[The evolution of cuneiform written characters.]{The evolution of cuneiform written characters. Reproduced from \citet{maspero1916recueil} (Creative Commons Attribution-Share Alike 4.0 International license). \vspace{1em}}
    \label{fig:mesopotamia}
\end{figure}

Beyond their physical realizations, these discrete units can also be associated with the meaning encoded by natural language. This association can be seen clearly in early writing systems such as cuneiform (Figure~\ref{fig:mesopotamia}): the written pictographic form corresponding to each morpheme has an obvious association with physical concepts, allowing readers to associate the physical realization of morphemes with objects in the world \citep{walker1990reading}. And yet, even in such cases, meaning itself is abstract and unobservable; what we can more directly study, however, are the inferential properties of language utterances, as opposed to any direct representation of meaning.

Consider, as an example, that the written text ``Robert woke up'' can be interpreted by a reader beyond the visual stimulus of the letters to indicate something about the world: that there is some person Robert and that this person completed a specific action of awaking. Performing tasks generally associated with understanding natural language requires accessing this encoded information. In short, someone cannot really be said to understand the text ``Robert woke up'' unless they infer the existence of the person Robert and associate Robert with an action of awaking.

This thesis concerns how one might evaluate computational systems based on their ability to make appropriate inferences from the literal, realized form of text. And, furthermore, how one might use such evaluation to measure the extent to which a given computational system can appropriately capture the meaning expressed by a given language utterance.

Studying computational systems that are intended to make inferences based on the realized form of natural language has far-reaching implications both in aiding our scientific understanding of language as a natural phenomenon and in developing \ac{NLP} systems that can be applied to complex tasks which implicitly require such methods---consider the aspirational goal of a computer program capable of executing natural language instructions which require inferences over long, complex contexts, for instance.

\begin{figure}[t]
    \centering
    \includegraphics[width=15cm]{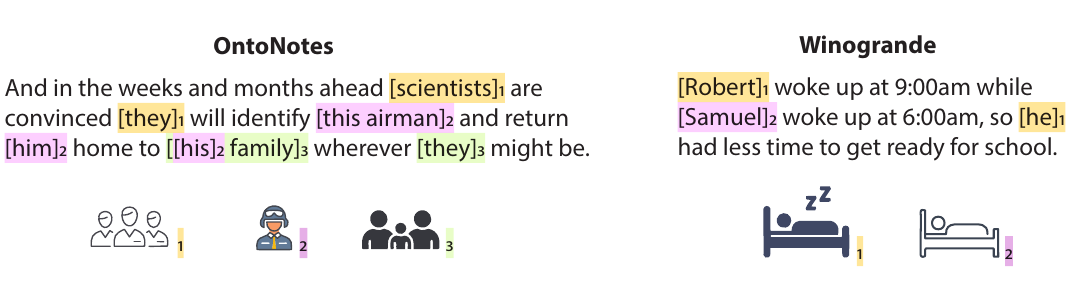}
    \caption[Demonstrative examples of coreferences annotated in two popular datasets.]{Demonstrative examples of coreferences annotated in two popular datasets: OntoNotes~\citep{marcus2011ontonotes} and Winogrande~\citep{winogrande_2020}. Square brackets mark coreferring linguistic expressions, and subscripts indicate the entities to which the expressions refer. \vspace{1em}}
    \label{fig:coref-examples-intro}
\end{figure}

But, how can one practically evaluate a computational system intended to make appropriate inferences based on natural language? As alluded to previously, the literal meaning encoded by natural language, broadly referred to as semantics, is an unobservable theoretical concept that cannot be measured directly. Therefore, a popular approach to evaluating such systems, which spans the history of NLP as a field, has been to evaluate these systems based on the extent to which they can sufficiently infer coreferences. Coreference is broadly the phenomenon of two or more linguistic expressions referring to the same entity, concept, or event. (See Figure~\ref{fig:coref-examples-intro} for examples of coreference annotated in popular research datasets.) The task of \textit{inferring}, or equivalently \textit{resolving}, coreferences is commonly referred to as \ac{CR}~\citep[e.g.,][]{bagga1998evaluation,ng-2005-machine,SUKTHANKER2020139} and is considered a necessary component of representing the semantics of language~\citep{hobbs1978resolving}. Therefore, if one can show that a given system either does not or cannot sufficiently infer coreference relations, it follows that said system is limited in its capacity to infer representations of semantics.

To give a concrete example, consider the passage taken from an instance of the popular OntoNotes dataset that is presented earlier in Figure~\ref{fig:coref-examples-intro}:

\begin{quote}
    \textit{And in the weeks and months ahead scientists are convinced they will identify this airman and return
him home to his family wherever they might be.}
\end{quote}
A complete interpretation of this passage should somehow encode that the nominal expression ``this airman'' and the mention of the pronoun ``him'' are coreferring in that both expressions refer to the same entity---some particular airman in the world.

From the 1960s to today, a wealth of existing work has functioned on this premise in order to use CR as an intermediate task to evaluate computational, semantics-inferring systems~\citep[e.g.,][]{Charniak1968CARPSAP,chinchor-marsh-1998-appendix,levesque_winograd_2012,pradhan-etal-2012-conll,emami-etal-2019-knowref}. The phenomenon of coreference has the useful characteristic that it is often operationalized as a simple binary relationship between spans of text (informally, ``do these two spans refer to the same entity or not?'') which is accessible in that it can be explicitly annotated and systems' inferences can be scored automatically. The majority of evaluation work follows from the basic premises of such a formulation.

So, then, what can we conclude from the extensive literature of coreference-based evaluations? While coreference resolution has served as a central benchmark for evaluating systems' ability to make appropriate inferences based on meaning, there are persistent concerns about its sufficiency and validity. In particular, although OntoNotes has long been the de facto standard dataset for evaluating coreference systems, its limitations in domain coverage and annotation consistency have been widely acknowledged~\citep{pradhan-etal-2012-conll}. Efforts such as the \textit{Coreference Resolution Beyond OntoNotes} (CORBON) workshop series~\citep{ws-2017-coreference} have explicitly aimed to push the field beyond OntoNotes. However, it remains unclear to what extent these efforts have yielded meaningful progress, as evaluations continue to produce conflicting or inconclusive findings about systems' abilities. In other words, coreference remains a contested evaluation target for meaning-based inference, motivating the need to critically assess existing practices.

In this thesis, I address these concerns by examining the design of coreference evaluation setups and the conclusions they support. Specifically, I analyze existing evaluation practices, identify sources of inconsistency or misalignment between definitions and metrics, and propose novel extensions to address these limitations. I focus in particular on the question of whether current evaluation settings allow us to characterize the generalization behavior of contemporary NLP systems, such as large language models, with respect to coreference resolution.

\textit{I find that, in the case of coreference-based evaluations, we are often limited in terms of the conclusions we can reach due to the evaluation setup itself. In particular, evaluation results may be affected by inconsistent definitions of the same concept or incorrect assumptions as to what is being measured. Nonetheless, more restricted evaluation settings indicate that the dominant paradigm in the field of natural language processing (NLP)---namely, language models as NLP systems---leads to systems that may be relatively strong, outperforming previous baselines, but fragile, exhibiting poor generalization to novel or varied contexts. That is, contemporary systems are improvements over previous baseline methods but in certain cases do not generalize in ways that one would expect of a system capable of understanding natural language at a human level.}

This conclusion can also be stated more succinctly as the following \textbf{thesis statement}:
\begin{tcolorbox}
\vspace{1.2em}
\begin{quote}
    \textit{Conclusions drawn from coreference-based evaluations are often limited by assumptions and design choices that restrict their generalizability; nevertheless, more controlled evaluation settings reveal that large language models can achieve high performance on standard benchmarks yet remain fragile, exhibiting poor generalization to novel or varied contexts when inferring coreference relations.}
\end{quote}
\vspace{0em}
\end{tcolorbox}

As is evident from the structure of this thesis statement, this thesis sets out to establish two closely related, central points. For the purposes of introduction, I will refer to these as \textbf{Point 1} and \textbf{Point 2}. The first point, \textbf{Point 1}, is that existing coreference-based evaluations are limited in the generality of the conclusions they allow one to reach. The second point, \textbf{Point 2}, is that certain, restricted coreference-based evaluations may nonetheless have the potential to indicate something about language models as possible systems for inferring appropriate representations of the semantics of natural language: that such systems may themselves be limited in terms of how well they generalize.

\paragraph{Point 1: Existing coreference-based evaluations are limited in the generality of the conclusions they allow one to reach.}

Chapters \ref{chap:challenge-set-assumption} to \ref{chap:supervised-comparison} are primarily aimed at establishing \textbf{Point 1} and in doing so connect this thesis to the vast literature of existing coreference-based evaluations. These evaluations most commonly function as follows: first, human annotators annotate coreference relations in a corpus of natural language; then, the extent to which a system is capable of inferring coreference is measured as the agreement between the respective system's inferences and the human annotations. I will refer to this style of evaluation as \textit{canonical} evaluations because it represents the most common form of evaluation of coreference resolution in the literature.

In order to establish \textbf{Point 1}, I rely on the existing framework of measurement validity~\citep[e.g.,][]{yun2002estimating,jacobs2021measurement}. This is a framework developed in the quantitative social sciences used for discussing the conceptual validity of a measurement by analyzing if said measurement accurately reflects what is purportedly being measured. As I will show in the case of existing coreference-based evaluations, taking the perspective of this framework reveals clear limitations in terms of the conclusions one can reach. These limitations are evidenced by factors such as inconsistent definitions of what is being measured and by the fact that the relative performance rankings of models can vary substantially between evaluation datasets.

\paragraph{Point 2: Restricted coreference-based evaluations indicate that language models may be limited as systems of inferring representations of semantics.}

This second point is recurringly suggested throughout the first three core chapters, but the aforementioned problems of validity prevent the point from being concluded based on a single, coherent argument. In Chapter \ref{chap:semantic-plausibility}, the final core chapter of this thesis, I build on the groundwork established in the preceding three chapters and present a novel evaluation which more directly supports \textbf{Point 2}. To do so, I focus on a specific aspect of semantics believed to be central to resolving coreferences: inferring the relative plausibility of events~\citep[\textit{i.a.}]{hobbs1978resolving,denis-baldridge-2008-specialized,zhang-etal-2019-knowledge}. This aspect can be demonstrated using the instance from the Winogrande dataset presented in Figure~\ref{fig:coref-examples-intro}:
\begin{quote}
    \textit{Robert woke up at 9:00am while Samuel woke up at 6:00am, so he had less time to get ready for school.}
\end{quote}
Consider your initial interpretation of this sentence. Who had less time to get ready for school? Equivalently, who is the pronoun ``he'' referring to?

Resolving the pronoun ``he'' to ``Robert'' amounts to inferring the event ``Robert had less time to get ready for school.'' This is in contrast to another possible resolution of ``he'' to ``Samuel'' which would have instead amounted to the inference ``Samuel had less time to get ready for school.'' Yet, in the case of this alternate interpretation, one could argue that the resulting event is a \textit{less plausible} inference because it was Samuel who woke up earlier and thus most plausibly had more time to get ready. In this way, coreference resolution can depend on inferring the relative plausibility of events expressed by possible interpretations of the original text.

Focusing on controlled evaluations of this more restricted aspect of coreference resolution has the advantage of sidestepping certain limitations observed in canonical evaluations. Specifically, by using carefully designed, simple test instances, one can shift attention away from contested or overly broad definitions of coreference and instead evaluate whether systems make the appropriate inferences that coreference resolution is intended to test. This emphasis on the inferential properties of resolution targets the core functional goal of the task—tracking entity reference in context without the need to agree on a universal definition of coreference phenomena across all linguistic contexts.

Based on the novel evaluation that I conduct in this more restricted case, I provide convergent evidence that the representations inferred by modern language models may be limited. In particular, I present evidence that language models and related systems fail to generalize in ways one would expect of an appropriate system when linguistic expressions are substituted with abstractions that should not alter the relative plausibility of events. For example, consider again the instance from the Winogrande dataset presented in Figure~\ref{fig:coref-examples-intro}. One would not expect the resolution of the coreference between the mentions ``Robert'' and ``he'' to change if the name Robert was replaced with a more abstract expression such as ``my friend.'' However, I present evidence that language models may be overly sensitive to such perturbations in textual form.

\paragraph{Uniting Point 1 and Point 2.}

Considering again the overall thesis statement, these two overarching points work in conjunction with one another. Establishing the first helps to clarify what it means for an evaluation to be valid, and establishing the second is in turn based on designing evaluations that attempt to satisfy certain aspects of validity. Namely, contestedness, which refers specifically to disagreements about how the construct itself should be defined, and convergent validity, which refers to the degree to which different evaluations or measures of the same construct yield consistent, mutually reinforcing results.

The positioning of this thesis within the NLP literature is unique in that I ultimately consider two distinct perspectives that are typically studied independently in terms of what it means to draw a general conclusion. On the one hand, I consider more traditional machine learning conceptualizations of generalization in that results generalize to samples from some underlying test distribution~\citep[e.g.,][]{hastie2009model}. Such is a common conceptualization of generalization in coreference-based evaluation and NLP or machine learning more broadly. On the other hand, I consider measurement theory from the quantitative social sciences as a tool for understanding what is being measured and whether the measurement results are valid in that they apply generally, beyond the constraints of a single experiment~\citep[e.g.,][]{adcock_collier_2001,jacobs2021measurement}. In this way, the results of this thesis may be of interest not only to researchers studying coreference, but to researchers of complex natural language phenomena more broadly. Relatedly, connecting these two conceptualizations of \textit{generalization} has the potential to benefit other areas of NLP or machine learning research as we as a field work towards understanding the real-world implications of our research findings.

Finally, I would like to note that it would be too presumptuous to try to definitively conclude whether any given evaluation is or is not valid, and to do so is not a goal of this thesis. Rather, the recurring theme of this thesis is that certain evaluations may fail certain criteria of validity to some degree; and, one may be able to design or restrict evaluations to overcome such limitations to an extent. It is exactly the framework of measurement validity that provides a new lens for discussing what these failures might be.

\section{Overview}
\label{sec:structure-overview}

\begin{figure}[t]
    \centering
    \includegraphics[width=15cm]{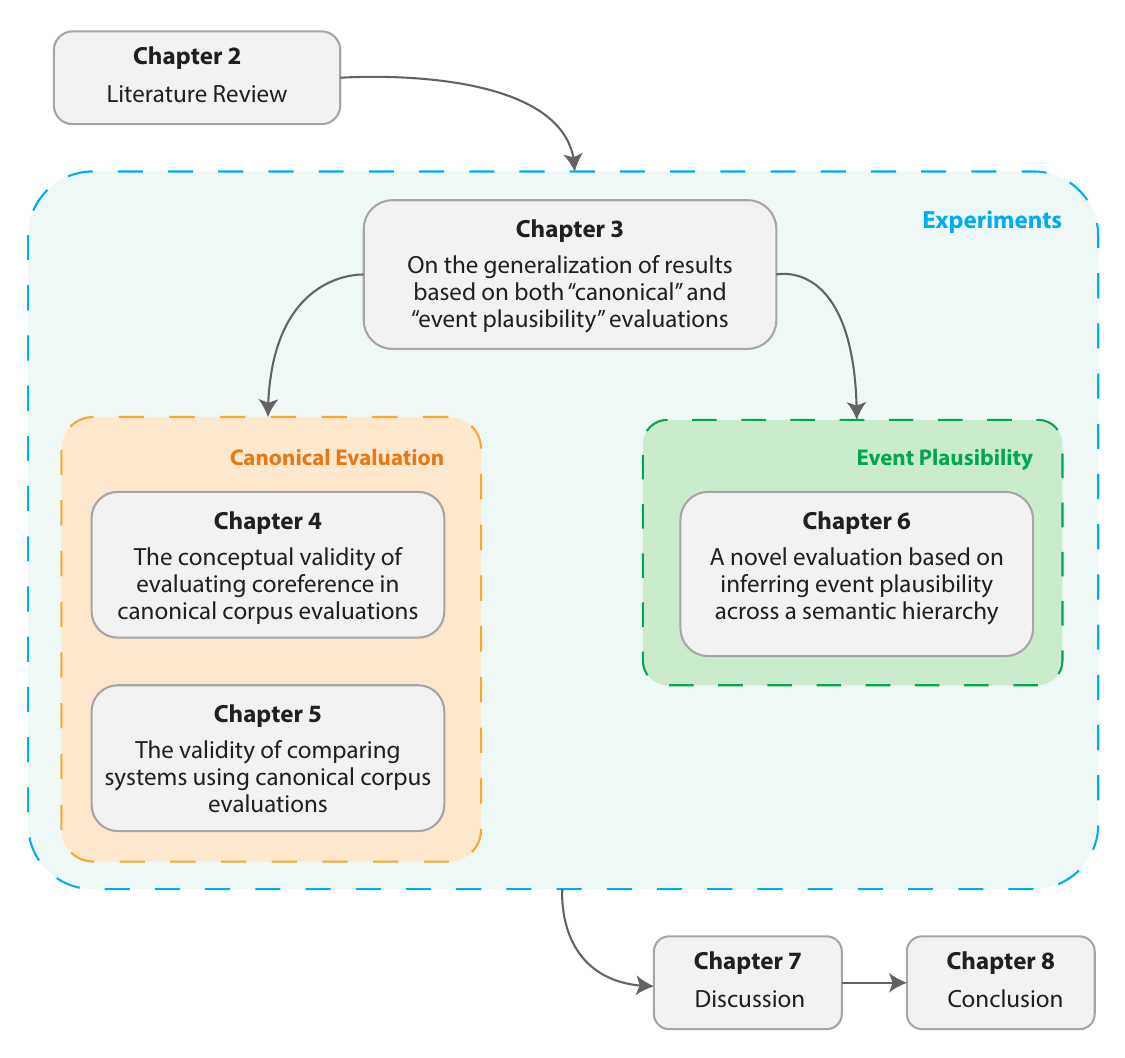}
    \caption[A visualized outline of the structure of this thesis.]{A visualized outline of the structure of this thesis. The body of the thesis focuses on empirical experiments. Certain experiments concern either a) \textit{canonical corpus evaluations} or b) targeted evaluations of an important aspect of resolving coreference, inferring \textit{event plausibility}, as outlined in Section~\ref{sec:structure-overview}. \vspace{1em}}
    \label{fig:chapters-overview}
\end{figure}

This section provides a high-level overview of the structure of this thesis focusing on the experiments presented in each chapter. See Figure~\ref{fig:chapters-overview} for a visualization of this structure.

\paragraph{Chapter~\ref{chap:challenge-set-assumption}}

Following a literature review in Chapter~\ref{chap:literature-review}, Chapter~\ref{chap:challenge-set-assumption} introduces and compares two types of coreference-based evaluations that together represent the majority of such evaluation in the broader literature and will be the focus of this thesis. These two types are, as alluded to previously, a) ``canonical'' evaluations and b) controlled evaluations based on event plausibility. This comparison achieves two things: first, it highlights some initial limitations of coreference-based evaluations and gives a sense of how the framework of measurement validity helps to discuss these limitations; and, second, it connects the more canonical style of coreference-based evaluation that will be the focus of the next two chapters with the more controlled style of evaluation that will be the focus of the final chapter of this thesis.

The specific finding of this first set of experiments is that conclusions are not always consistent between these two types of evaluations. In particular, language-model-based approaches that have led to improvements in a particular style of evaluation do not necessarily transfer to other evaluations.

\paragraph{Chapter~\ref{chap:measurement}}

Then, in Chapter~\ref{chap:measurement}, I describe more thoroughly the limitations of existing evaluations and how the perspective of measurement validity helps to unpack these limitations. Using this framework, I provide evidence that evaluations often reflect idiosyncrasies of the experimental setup rather than the ability of a system to generalize some consistent concept of CR across datasets. Specifically, I show that across seven canonical datasets, measurements intended to reflect the generalization of the ability to perform CR and often correlated with differences in how coreference is both defined and operationalized. This chapter serves to more formally introduce and apply concepts from the area of measurement validity to coreference-based evaluations in NLP.

\paragraph{Chapter~\ref{chap:supervised-comparison}}

Building on these findings in Chapter~\ref{chap:supervised-comparison}, I make progress towards standardizing the comparison of state-of-the-art approaches to CR. I systematically evaluate five existing CR systems and control for certain design decisions including the underlying architecture used by each. I conclude that controlling for the choice of this underlying architecture reduces most, but not all, of the increase in performance reported in the past five years. That is, with more extensive baselines I am able to show that improvements in the modeling of coreference may not be due to the factors that were originally believed. This result further highlights the limitations of existing evaluation practices by showing again that evaluation results can be contradictory.

\paragraph{Chapter~\ref{chap:semantic-plausibility}}

I then focus more narrowly on novel evaluations of inferring event plausibility, which is in certain cases an aspect of coreference resolution, in order to avoid certain limitations identified in the preceding chapters. In Chapter~\ref{chap:semantic-plausibility}, the final core chapter of this thesis, I present a novel evaluation designed to assess the consistency of language models in inferring the plausibility of events. Specifically, this evaluation considers the extent to which models are consistent in inferring events across a semantic taxonomy. This evaluation reveals that language models are markedly inconsistent in terms of the inferences made. For example, contemporary systems may lead to the inference that ``a person breathing'' is highly plausible while ``a dentist breathing'' is not. I find this inconsistency persists even when language models are softly injected with lexical knowledge. Together with the vocabulary of measurement validity introduced in the preceding chapters, this novel evaluation suggests limitations of modern language models in the constrained setting considered.

\paragraph{Discussion and Conclusion}

Taken together, these works provide new perspectives on coreference resolution evaluations as used for evaluating computational models of semantics. Through different perspectives, these experiments provide insight into current success and limitations all the while emphasizing the importance of considering generalization ability in context, beyond singular, scalar metrics. The thesis closes with a discussion of the implications of the findings designed to be accessible by a more general audience in Chapter~\ref{chap:discussion} and a conclusion in Chapter~\ref{chap:conclusion}.

\chapter{Literature Review}
\label{chap:literature-review}

\nonengepigraph{Eadem sunt, quae sibi mutuo substitui possunt, salva veritate.}{Terms which can be substituted for one another wherever we please without altering the truth of any statement are the same or coincident.}{\citet{leibniz1989general}}

\vspace{15pt}

This chapter reviews the literature necessary for understanding the main content of this thesis. Naturally, this includes an overview of existing work related to the evaluation of coreference resolution (\S\ref{sec:lit-review-coreference-resolution}). In addition, I overview necessary related work in the following areas: measurement validity (\S\ref{sec:lit-review-measurement-validity}), the framework which I use to analyze the validity of evaluation practices; language modeling (\S\ref{sec:lit-review-language-modeling}), as I will focus primarily on the evaluation of language models as NLP systems capable of drawing inferences; and, finally, modeling semantic plausibility (\S\ref{sec:lit-review-semantic-plausibility}), as Chapter~\ref{chap:semantic-plausibility} focuses on inferring the semantic plausibility of events as a subproblem of inferring coreference relations.

While this section aims to provide the necessary background for the experiments of this thesis, additional details related to relevant datasets or models are presented alongside the corresponding descriptions of experimental methodology.

\section{Coreference Resolution}
\label{sec:lit-review-coreference-resolution}

Necessarily, coherent discourse in natural language, beyond some sufficient length, will make repeated references to the same entities~\citep{hobbs1979coherence}. The broad phenomenon of multiple linguistic expressions referring to the same discourse entity is called coreference. A \textit{discourse entity}, also known as a \textit{discourse referent}, refers to an entity introduced in a natural language discourse~\citep{karttunen1976discourse,webber2016formal}. Informally, a discourse entity is any \textit{thing} being discussed in the discourse, such as a person, object, idea, or event. The discourse entity is a theoretical concept without a precise, operational definition used to make the distinction between entities in the real world versus entities that exist in a reader's model of a discourse. For instance, I can reference the fictional character \textit{Robert} within a discourse regardless of whether there exists a corresponding person in the literal world of the reader.

Coreference resolution (CR) is then the task of identifying those expressions that are coreferring in that they refer to the same discourse entity~\citep[][\textit{i.a.}]{winograd_1972,kantor1977management,hirst1981anaphora}.
This task has been studied extensively~\citep[\textit{e.g.,}][]{SUKTHANKER2020139,poesio2023computational} and is considered a core part of representing the semantics of natural language~\citep{hobbs1978resolving}. The basic premise of the historical argument for the importance of coreference is an intuitive one: that if a reader cannot infer coreferences they cannot be said to understand the meaning of a natural language utterance because they cannot perform tasks generally agreed to require understanding. An obvious such task is answering the human interpretable question ``what does this specific pronoun refer to?'' Indeed, the broad task of question answering is in general believed to depend on resolving coreferences~\citep{vicedo2008coreference}.

\begin{figure}[t]
    \centering
    \includegraphics[width=16cm]{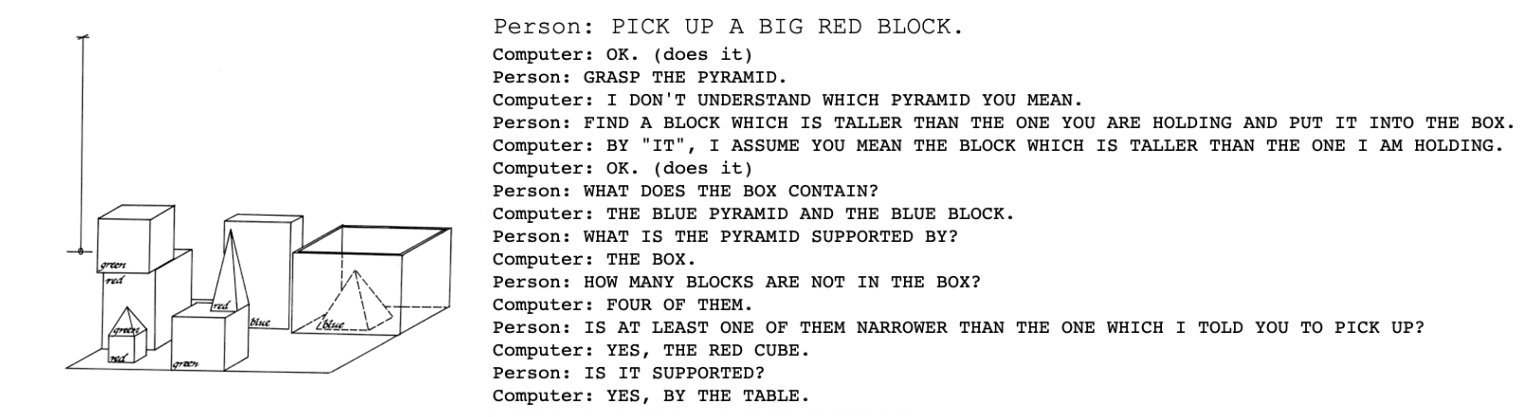}
    \caption[The SHRDLU program proposed by Terry Winograd.]{The SHRDLU program proposed by Terry Winograd which was designed to interact with a virtual world of 3D objects based on natural language instructions. On the left is a visualization of the virtual world of objects, and on the right an example interaction between a person and the program. Figure reproduced from \citet{winograd_1972}. (M.I.T. Project MAC, Defense Technical Information Center, Public Domain)
    \vspace{1em}}
    \label{fig:shrdlu-example}
\end{figure}

Figure~\ref{fig:shrdlu-example} presents an early example of an explicit model of coreferences designed as part of a larger system by \citet{winograd_1972}. Winograd's SHRDLU system was designed to manipulate blocks in a virtual 3D environment based on a natural language dialogue with a user. As part of this interaction, it was necessary that the system was able to resolve pronominal coreferences in the user's utterances in order to determine which block to move. This pronoun resolution relied on a rule-based parser.

\citet{hobbs1978resolving} famously demonstrated that many pronominal coreferences in natural language can be resolved based on simple heuristics such as the syntactic distance between a pronoun and a candidate antecedent (a candidate antecedent being an earlier expression that could possibly be coreferring with the pronoun). This approach is often referred to as \textit{Hobbs' naive algorithm} because it relied on naive heuristics. Hobbs also noted, however, that many coreferences rely on sophisticated knowledge about the world not easily accessible from surface-level heuristics. He proposed a theoretical model of how such references could be resolved, but the model was too abstract to be practically realizable as it depends on full semantic representations and ensuring references satisfy semantic constraints without complete definitions of how these dependencies can be realized. This theoretical model is sometimes referred to as \textit{Hobbs' semantic algorithm}, although it is not an algorithm in the standard sense because it was not described in a way that could be directly implemented by a computer even with modern NLP approaches.

Following these early explorations of computational systems and coreference, evaluations became more standardized. In the remainder of this section, I will survey relevant literature both in terms of evaluating and building computational systems capable of inferring coreferences.

\subsection{Coreference-based Evaluations}

Standardized benchmarking of coreference resolution dates back to the MUC~\citep{grishman-sundheim-1996-design} and ACE~\citep{chen2006exploration} sets of shared tasks which evaluated systems' ability to perform coreference resolution on restricted sets of linguistic expressions. These evaluations were ``restricted'' in the sense that only a fixed set of entity types was annotated for coreference in natural language corpora. Part of the stated motivation for focusing on restricted sets of entity types is that doing so avoided highly-disputed cases.

Following these initial series of shared tasks, the OntoNotes project~\citep{hovy-etal-2006-ontonotes} then represented the largest endeavor to annotate coreferences in natural language at the time of its publication. This annotation was used as part of the CoNLL 2011 shared task~\citep{pradhan-etal-2011-conll}, and later the CoNLL 2012 shared task~\citep{pradhan-etal-2012-conll} evaluations. These shared tasks expanded the evaluation of coreference relations to ``unrestricted'' linguistic expressions in the sense that the expressions were not restricted to a particular semantic class or type of entity.

To give a sense of this \textit{restricted} versus \textit{unrestricted} distinction, the ACE dataset was annotated only for coreference between person, organization, geopolitical entity, facility, and location referring expressions; on the other hand, OntoNotes was annotated for any coreference between essentially any noun phrase (ignoring restrictions on how the literal span of a noun phrase was annotated) and even certain verb phrases.

The OntoNotes project inspired several more recent datasets of coreference annotations, and corresponding benchmarks, which I will outline in more detail in Chapters \ref{chap:challenge-set-assumption} and \ref{chap:measurement}. Notable more recent datasets include PreCo \citep{chen-etal-2018-preco}, the largest scale dataset of coreference annotations to date which was annotated by trained university students, Phrase Detectives~\citep{yu-etal-2023-aggregating} which crowdsourced annotations through online games, the GUM corpus~\citep{10.1007/s10579-016-9343-x}, which includes annotations of many anaphoric phenomena and is iteratively improved, LitBank~\citep{bamman-etal-2020-annotated} which consists of coreference annotated in English literature by experts, and ARRAU~\citep{Uryupina_Artstein_Bristot_Cavicchio_Delogu_Rodriguez_Poesio_2020}, a dataset of multiple anaphoric phenomenon annotated by experts.

These mentioned datasets largely focus on English-language corpora, but also include annotations of other languages at smaller scales such as the inclusion of Chinese and Arabic corpora in the OntoNotes dataset. The recent CRAC shared tasks~\citep{crac-2022-crac,crac-2023-crac} have focused on evaluation of coreference across multiple datasets which include natural language discourses spanning over twelve languages \citep{11234/1-5896}.

Precisely how coreference has been annotated differs across these datasets and will be discussed in more detail in Chapter~\ref{chap:measurement}. In general, coreference has been annotated as a discrete relationship between linguistic expressions, although some annotations have diverged from such a formulation---for instance, in the annotation of inherently ambiguous coreferences for which there is no obvious, discrete resolution \citep[e.g.,][]{bamman-etal-2020-annotated,yuan-etal-2023-ambicoref}. Such ambiguity is often used intentionally as a literary device \citep{timmerman2002robert}. In a similar vein, \citet{recasens2011identity} challenged the discrete formulation of coreference by introducing a more granular typology of near-identity relations.

\paragraph{The Winograd Schema Challenge.}

\citet{levesque_winograd_2012} presented the \ac{WSC} as an alternate challenge set of coreference resolution problems designed to require world knowledge and commonsense reasoning abilities. In contrast to the aforementioned ``canonical'' evaluations of coreference which focus on annotations of coreference within existing corpora of natural language discourse, the WSC focuses on targeted examples specifically designed for the purposes of the evaluation itself. The WSC has inspired what is essentially a parallel line of research to canonical coreference resolution evaluations which instead focuses on specific challenge sets. Because the WSC challenge set was proposed to test systems allegedly capable of solving pronominal coreference resolution problems, this and related challenge sets have therefore been classified as a coreference resolution evaluation in the literature. For instance, WSC datasets are classified as a coreference resolution task in the popular Super-Natural-Instructions collection of tasks~\citep{wang-etal-2022-super}.

The WSC consists of a collection of problems where each individual problem is called a Winograd schema. Winograd schemas are minimal sentence pairs that differ in the resolution of a definite pronoun. At its most general, a Winograd schema is a sentence that includes \textit{two entity mentions}, \textit{an ambiguous pronoun}, and \textit{a special word} such that changing just this special word can change the resolution of the pronoun. Consider the following popular example of a Winograd schema:

\begin{itemize}
    \item[] \textit{\underline{The trophy} did not fit into \underline{the suitcase} because \underline{it} was too \underline{small}.}
    \item[] \textit{\underline{The trophy} did not fit into \underline{the suitcase} because \underline{it} was too \underline{large}.}
\end{itemize}

In this example, the minimal pair of sentences differ by the final word which changes the resolution of the pronoun. Because resolving the pronoun in each sentence is designed to depend on commonsense knowledge about the world, Winograd schemas have become a popular diagnostic of commonsense reasoning ability. The design of these schemas was dependent on manual curation and the expertise of the dataset creator.

The original WSC consisted of 273 problems, but there are many related variations that I will outline where they are used in the experiments of this thesis. One notable variant of the WSC is the Winogrande dataset~\citep{sakaguchi2021winogrande}. This dataset consists of thousands of instances inspired by Winograd schemas created and refined by crowdworkers. In contrast to the original WSC which consists of pronoun resolution problems, Winogrande was created as a fill-in-the-blank, cloze-style task based on the observation that Winograd schemas can often be reformulated in this alternate form. (Nonetheless, instances from the Winogrande dataset are often presented as pronominal coreference resolution problems including in the original publication itself.)

\subsection{Evaluation Metrics}

\begin{figure}[t]
    \centering
    \includegraphics[width=15cm]{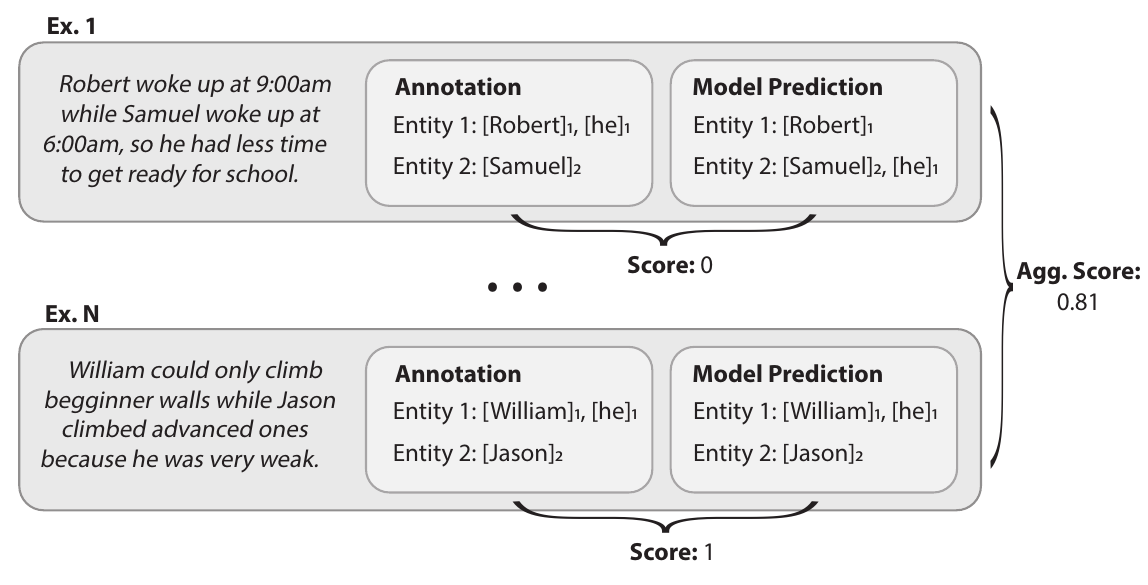}
    \caption[An illustration of how model performance is most commonly evaluated using the Winogrande dataset.]{An illustration of how model performance is most commonly evaluated using the Winogrande dataset~\citep{winogrande_2020} as an example. For each instance in the dataset, predicted coreferences are elicited from the model being evaluated and compared against coreferences annotated by a human annotator. Accuracy is then calculated across all examples in the test set. (Certain details regarding data formatting are simplified for illustrative purposes. The spelling of ``beginner'' is as this example appears in the original dataset.) \vspace{1em}}
    \label{fig:agg-stat-example}
\end{figure}

Evaluation in coreference resolution benchmarking generally functions by averaging the agreement between human-annotated coreferences and those predicted by the model being tested across an entire test set of instances. See Figure~\ref{fig:agg-stat-example} for an example of how CR is typically scored using the Winogrande dataset, for instance.

When instances have two possible resolutions, as in Winograd schemas, accuracy is the most common evaluation metric, although precision, recall, and F1 score have also been used for evaluation of these instances~\citep{emami-etal-2018-generalized}.

In other cases, where multiple coreferences in a discourse are being evaluated simultaneously, the problem of coreference resolution is generally considered a clustering problem. In the case of coreference being annotated as the clustering of linguistic expressions, popular evaluation metrics are MUC~\citep{vilain-etal-1995-model}, B-cubed~\citep{bagga-baldwin-1998-entity-based}, and CEAF~\citep{luo-2005-coreference}. In addition, CoNLL F1 score was proposed by \citet{denis2009global} as a method of summarizing model performance by averaging these three scores and became a standard practice following its adoption in the CoNLL 2012 shared task~\citep{pradhan-etal-2012-conll}. This standard, while pervasive, is disputed and not based on any empirical motivation~\citep{moosavi-strube-2016-coreference}.

To give some intuition for these evaluation metrics: MUC is often described as a method of evaluating coreference based on \textbf{links} between linguistic expressions, B-cubed a method of evaluating based on \textbf{mentions} (where a mention is another way of saying a coreferring linguistic expression which is almost always a span of words), and CEAF---in particular CEAF\textsubscript{e} which is the most commonly used version of the metric---a method of evaluating based on \textbf{discourse entities}. That being said, all three metrics are often highly correlated when calculated on popular evaluation test sets~\citep{moosavi-strube-2016-coreference}.

To give a sense of how these evaluation metrics function, I will outline the original MUC metric below. (Other metrics will be introduced in more detail where they are used within the experiments of this thesis.)

\begin{figure}
    \centering
    \includegraphics[width=0.65\linewidth]{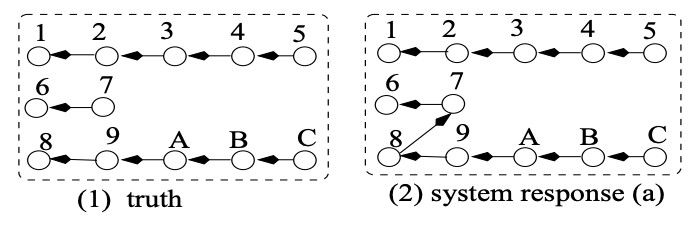}
    \caption[A demonstrative example of an annotated coreference-based clustering of linguistic expressions.]{A demonstrative example of an annotated clustering of linguistic expressions (``truth'') and a system-predicted clustering (``system response''). Each circle represents a referring linguistic expression, and circles in the same chain are considered coreferring as annotated (truth) or predicted (system response). Reproduced from \citet{vilain-etal-1995-model}. (Creative Commons Attribution-NonCommercial-ShareAlike 3.0 International License)}
    \label{fig:muc-example}
\end{figure}

For a given discourse entity $i$, let $S_i$ be the set of linguistic expressions annotated in a discourse as referring to $i$. Then, let $\text{p}(S_i)$ be the set $S_i$ partitioned into subsets that are predicted as coreferring by the system being evaluated. For instance, if $S_i = \{ 1, 2, 3 \}$ and the predicted clustering is only $\{1, 2 \}$, then $\text{p}(S_i) = \{\{1, 2 \}, \{ 3 \} \}$. 

MUC recall is then calculated as
\begin{align}
    \text{Recall}_{\mathrm{MUC}} = \frac{\sum_{i} \left( |S_i| - |\text{p}(S_i)| \right)}{\sum_{i} (|S_i| - 1)}
\end{align}
and precision can be calculated by switching the annotated and system-predicted clusters and using the same formula.

How, then, can this metric be interpreted? Intuitively, this MUC metric is calculating recall as follows: imagine all of the annotated coreferences were lined up in a sequential order as in Figure~\ref{fig:muc-example}. Consider the coreference between a linguistic expression and the sequentially next coreferring expression to be a ``link.'' MUC recall is then the ratio of annotated ``links'' that are predicted by the system, and precision is the ratio of system-predicted links that were actually annotated as coreferring.

Note that MUC, and in fact all related clustering metrics, do not consider the actual semantic content or relative position of a linguistic expression. For this reason, the accuracy of determining referring expressions may be reported independently. A relatively recent approach is to consider the overlap of predicted expressions with syntactic constituents in the discourse \citep{moosavi-etal-2019-using}.

\subsection{Coreference Resolution Systems}

Early systems explicitly designed to perform the task of CR were based on manual heuristics such as the word-distance between mentions and grammatical feature agreement \citep{raghunathan-etal-2010-multi}; however, more recent advances came as the result of incorporating fine-tuned language models~\citep{joshi-etal-2019-bert,https://doi.org/10.48550/arxiv.2210.14698}. For an outline of the now standard language model finetuning process, I defer to \citet{joshi-etal-2019-bert}.

In the remainder of this section, I will overview coreference resolution systems at a high level. I will go into more of the details of modern language models in \S\ref{sec:lit-review-language-modeling}.

\paragraph{Supervised Approaches}

Supervised approaches, which are those that are trained using explicitly labeled training instances, are most commonly based on either ``mention ranking'' \citep{lee-etal-2017-end} or ``shift-reduce'' \citep{webster-curran-2014-limited} architectures. At the time of the experiments of this thesis, popular mention ranking models are the state-of-the-art ``Word-level Coreference'' \citep{dobrovolskii-2021-word} and LingMess \citep{otmazgin2022lingmess} systems. The shift-reduce model of \citet{https://doi.org/10.48550/arxiv.2211.12142} was the model with the highest reported performance on the OntoNotes test set, the most popular dataset for evaluating CR, at the time of its publication. This model directly finetunes a large language model to output pairs of co-referring expressions for each sentence in a discourse. More details are provided for these models when they are compared in the relevant experiments.

One perspective is that the supervised finetuning of a language model for an explicit task, such as CR as in this case, might be viewed as an upper-bound of the model's capacity to encode representations necessary for solving said task~\citep{tenney-etal-2019-bert,zhu-etal-2022-predicting}. This perspective means if that a model cannot be finetuned to perform the task of CR, then the representations encoded by the model are inherently insufficient for representing coreference relations.

\paragraph{Unsupervised Approaches}

On the other hand, unsupervised approaches are those that are not trained on explicitly labeled instances of coreference.

\citet{doi:10.1073/pnas.1907367117} found that the latent representations induced by language models are correlated with coreference relations annotated in natural corpora, but only demonstrated this to be the case on limited subsets of annotations absent strong baselines.

\citet{le2023large} demonstrated that large-scale language models prompted with instructions to explicitly perform coreference resolution in a very specific format perform competitively in the setting where all coreferring linguistic expressions are provided in the model input and the expected task is only to cluster these expressions based on the entity to which they refer. In the setting where such expressions are not provided in the input, language models were found to perform worse than models trained to perform the task of coreference resolution but better than older, rule-based systems.
In contrast, \cite{zhu-etal-2024-large} found a negative result using an alternative prompt of a different format.

Meanwhile, \citet{gan-etal-2024-assessing} performed a manual evaluation the ability of prompted language models to explicitly perform coreference resolution and noted that in many cases the models appeared to successfully resolve coreference through manual inspection, but that this success did not translate to high accuracy scores in standard benchmarks.

\begin{figure}[t]
    \centering
    \includegraphics[width=13cm]{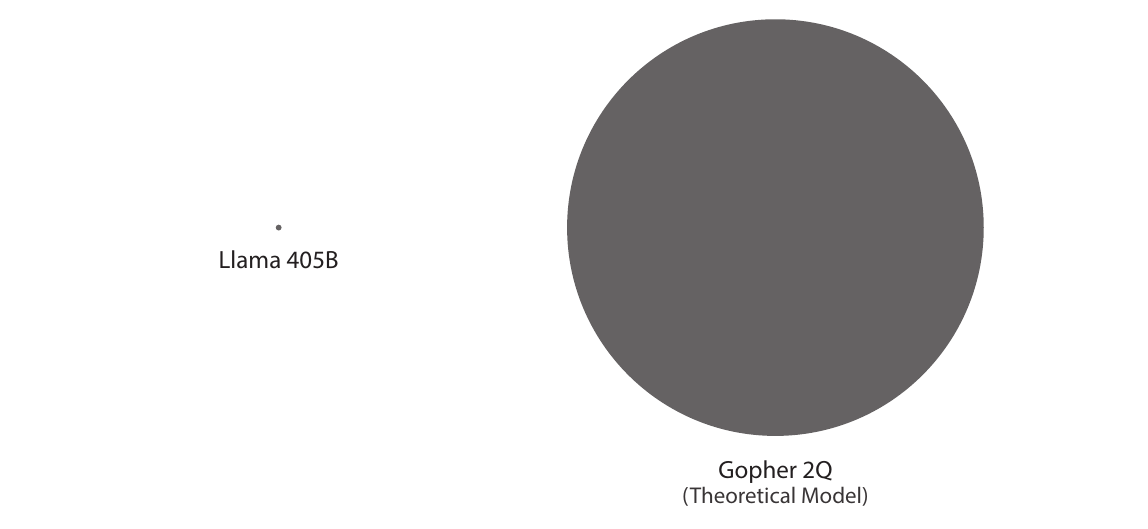}
    \caption[A visualization of the model size needed to achieve human baseline accuracy on the Winogrande test set.]{A visualization of the model size needed to achieve human baseline accuracy on the Winogrande test set as estimated by \citet{li-etal-2022-systematic}. The area of the circle on the left represents the total number parameters of the best performing open-weights language model as of 2024, LLama 3.1 405 billion~\citep{dubey2024llama}. Each parameter in this model is a 16-bit precision floating point decimal. The total file size of the model is $\sim$800 GB, and the model was trained using a cluster of 16,000 H100 Nvidia GPUs. By comparison, the area of the circle on the right represents the relative number of parameters to which a Gopher model is estimated to need to be scaled to achieve baseline human accuracy on the Winogrande test set (2 quadrillion parameters). \vspace{1em}}
    \label{fig:gopher}
\end{figure}

More recently, large language models are considered the current state of the art for representing the semantics of language, but there currently is not a consensus as to what extent these models appropriately encode coreferences. \citet{li-etal-2022-systematic} estimate that based on the empirical performance of the Gopher language model across various scales~\citep{rae2021scaling}, it would require at least 2 quadrillion parameters for a such a model to achieve human baseline accuracy on the artificial Winogrande test set (visualized in Figure~\ref{fig:gopher}).

\paragraph{Generalization of Coreference Resolution Systems}

Studies that consider the generalization of systems across multiple test datasets without additional training data have largely evaluated performance by aggregate performance metrics such as CoNLL F1~\citep{guha-etal-2015-removing,zeldes-zhang-2016-annotation}. Performance decreases when models are evaluated on corpora not included in the training dataset~\citep{zhu-etal-2021-ontogum}.

\citet{moosavi-strube-2017-lexical} showed that training and evaluating on OntoNotes likely leads to overfitting due to a high train/test overlap. Similarly, \citet{subramanian-roth-2019-improving} show that models trained on OntoNotes struggle to generalize to novel named entities. They present an adversarial training method to help overcome this issue and improve generalization. Additionally, studies have improved generalization by incorporating explicit linguistic features~\citep{moosavi-strube-2018-using,otmazgin2022lingmess}.

\section{Measurement Validity}
\label{sec:lit-review-measurement-validity}

In order to study how well coreference-based evaluations are measuring what is intended, I rely on existing methods for understanding the validity of measurements from the quantitative social sciences~\citep[][\textit{i.a.}]{black_doing_1999,adcock_collier_2001}.

Measurement is the process of quantifying some concept~\citep[e.g.,][]{pap2002handbook}. In the case of measuring theoretical concepts, such as the ability to resolve coreferences, these theoretical concepts are abstract, unobservable concepts which cannot be directly measured. Instead, the quantification of these concepts depends on inferring their value based on some other, observable concepts.

\begin{figure}
    \centering
    \includegraphics[width=0.95\linewidth]{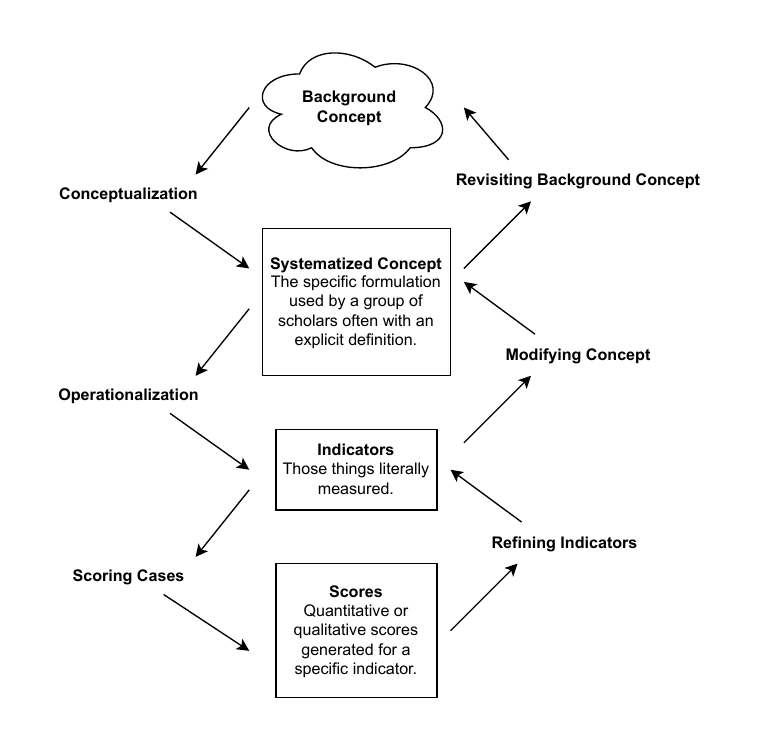}
    \caption[Levels and tasks of measurement.]{Based on an original figure from \citet{adcock_collier_2001}, levels and tasks of measurement. A background concept is conceptualized as a systematized concept which is measured based on the scores of indicators. In turn, these scores contribute to refining our understanding of the systematized and background concepts.}
    \label{fig:measurement-validity-overview}
\end{figure}

In existing work, this process is implicit for evaluations of NLP systems based on the ability to resolve coreference. By contrast, in the quantitative social sciences it is common to more explicitly describe what is intended to be measured, what indicators or direct observations will be used for the purposes of measurement, and what criteria can be used to argue the validity of the measurement. In fact, this entire process can be viewed as iterative as visualized in Figure~\ref{fig:measurement-validity-overview} summarizing an original figure from \citet{adcock_collier_2001}.

With this perspective, \citet{jacobs2021measurement} present a set of common issues that affect measurement validity. Among the taxonomy defined by \citet{jacobs2021measurement}, relevant issues of measurement which I will use in the analysis of this thesis are as follows:

\paragraph{Contestedness} Are there multiple, competing theoretical understandings of the concept being measured? That is to say, is the concept being measured a contested concept? If so, if it is not clear which definition is being used, it is possible that measurements will conflate multiple concepts, or that conclusions will not generalize more broadly. In particular, when there are various definitions or definitions are implicit, measurement may conflate distinct constructs.

\paragraph{Convergent Validity} Do measurements correlate with other measurements of the same construct? When multiple measurements of the same concept do not correlate, this indicates a potential problem in the validity of the measurements. If measurements do not correlated, this indicates that they may not be measuring the same construct, and relatedly may not be measuring the construct intended to be measured.

\paragraph{Discriminant Validity} Do measurements vary in ways that indicate they may be capturing other constructs than that intended to be measured? If so, this indicates that we cannot conclude that the measurements are capturing only that which is purported to be measured. Evidence of discriminant validity is that measurements \textit{do not} correlate with measurements of other unrelated constructs that are not intended to be measured.

\subsection{Measurement Modeling}

In the quantitative social sciences, measurement is often modeled explicitly as a measurement model (MM) in order to analyze what is being measured \citep{smelser_indicator_2001}. As I will show, this framework is also useful for discussing the measurement of coreference and CR model performance given that coreference is itself a contested theoretical concept. Whereas I will not go so far as to define an explicit model of measurement, which could be a future contribution of this line of work, understanding the process of formalizing measurement is useful for discussing the three aforementioned validity issues.

\para{Definition} \textit{\ac{MM}.} An MM is a structural model that describes the relationship between some unobservable theoretical concept and observable variables used to measure said concept. While an MM can be implicit, it is common to describe an MM as a formal probabilistic or algebraic structure in order to reason about what is being measured \citep{rasch_probabilistic_1980,bollen1989}. Following standard terminology, I will use \textit{construct} to refer to the theoretical concept being measured and \textit{indicators} to refer to the observable variables that are used to measure the construct \citep{kline_principles_2011}.

\begin{figure}[ht]
    \centering
    \includegraphics[width=0.25\textwidth]{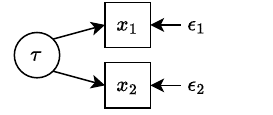}
    \caption[A graphical model representing a simple measurement model.]{A graphical model representing a simple MM with two indicators and independent error. Inspired based on an example from \citet{bollen1989}.}
    \label{fig:temperature-mm}
\end{figure}

\para{Example} Consider the following example inspired by \citet{bollen1989}. Temperature, represented as latent variable $\tau$, is a theoretical construct defined to be the average kinetic energy of molecules in a substance. Suppose that this construct is measured using readings from two distinct thermometers. Each reading is an indicator represented as latent variables $x_1$ and $x_2$, respectively. An explicit MM might then be defined to be the set of linear equations
\begin{equation}
\begin{split}
    x_1 &= w_1 \tau + \epsilon_1 \\
    x_2 &= w_2 \tau + \epsilon_2
\end{split}
\end{equation}
where here each indicator $x_i$ is assumed to be a linear function of the latent variable, weighted by $w_i$, as well as some unexplained, independent error $\epsilon_i$. This measurement model can be described as the graphical model in Figure~\ref{fig:temperature-mm}, a path diagram in this case.

Again, while I will use the general framework of measurement modeling for certain analyses, I will not use this precise, structural formulation. That being said, it is still useful to use this structural formulation in order to explain the role of MMs. Furthermore, such an explicit, structural formulation could be used to expand on the analyses in this thesis.

\section{Language Modeling}
\label{sec:lit-review-language-modeling}

The field of NLP is currently dominated by the language modeling paradigm~\citep[e.g.,][]{rogers-etal-2020-primer,zhao2023survey,zhou2023comprehensivesurveypretrainedfoundation}. In this paradigm, neural language models~\citep{bengio2000neural}---which can be thought of as computational systems trained to model the distribution of the surface form of language---are used for directly or indirectly solving most tasks. This is achieved either by further training a language model for a particular task using additional task-specific training data \citep{howard-ruder-2018-universal,Bommasani2021FoundationModels}, formulating the task as language modeling \citep{NEURIPS2020_1457c0d6}, or some combination of both of these two approaches.

The success of the language modeling paradigm across tasks has increasingly led to the study of language models as systems possibly capable of making inferences based on the latent semantics of natural language \citep[e.g.,][]{pavlick2022semantic,mahowald2024dissociating,merrill-etal-2024-learn}; and, while some dispute the plausibility that a model of the surface form of language can represent semantics \citep{bender-koller-2020-climbing,chomsky2023noam,https://doi.org/10.48550/arxiv.2301.06627}, language models are nonetheless an interesting object of study due to their strong performance when applied to tasks that require making inferences from the physical form of language---and in particular, tasks that have been argued to require inferring coreference relations, including question answering \citep{dasigi-etal-2019-quoref}, summarization \citep{liu-etal-2021-coreference}, and machine translation \citep{ohtani-etal-2019-context}.

\subsection{Preliminaries}

A corpus can be defined as a sequence of words $(w_1, w_2, \ldots, w_T)$, where $T$ is the total length of the corpus. All language modeling could be said to originate from \citet{predicting-english} who presented an early formalization of natural language as probabilistic. Shannon focused primarily on sequences of letters, but we can consider sequences of other linguistic units following the same basic principles.

Let $x_i$ be the word at position $i$ in the corpus. An $n$-gram refers to a contiguous subsequence of $n$ words in the corpus. Word $x_i$ and its' preceding $n-1$ words of context define the $n$-gram $x_{i-(n-1)}^{i} = x_{i-n+1},x_{i-n+2},...,x_{i}$. To simplify the notation, I will denote the first index of the $n$-gram as $s = i-(n-1)$. Shannon considers two probability distributions, the probability of occurrence of the $n$-gram $P(x_{s}^{i})$, and the probability of $x_i$ conditioned on its context, $P(x_i|x_{s}^{i-1})$. By the chain rule, the unconditional joint probability of the $n$-gram can be written as
\begin{align}
\label{equation:language-modeling}
     P\left(x_{s}^{i} \right) & = P\left( x_i|x_{s}^{i-1} \right) \cdot P\left( x_{s}^{i-1} \right) \\
     & = P\left( x_{s} \right) \prod_{j=s+1}^i P\left( x_j|x_{s}^{j-1} \right).
\end{align}

In this way, one can train a neural language model, popularly a Transformer-based language model \citep{NEURIPS2020_1457c0d6}, using \ac{MLE} to estimate the probability $P\left(x_{1}^{T} \right)$ by making the Markovian assumption that the probability of words only depends on a fixed context length and then training the model to predict each word given its context.

It is worth noting that the probabilities of sequences of words are believed to correlate with semantic plausibility and world knowledge. Foundationally, \citet{keller-2003} showed that MLE probabilities calculated from a large web corpus have a Spearman $\rho$ of 0.6 with human plausibility judgments for frequently occurring word pairs. However, for infrequent word pairs the correlation is close to zero.

\subsection{Transformer Language Models}

\begin{figure}[t]
    \centering
    \includegraphics[width=10cm]{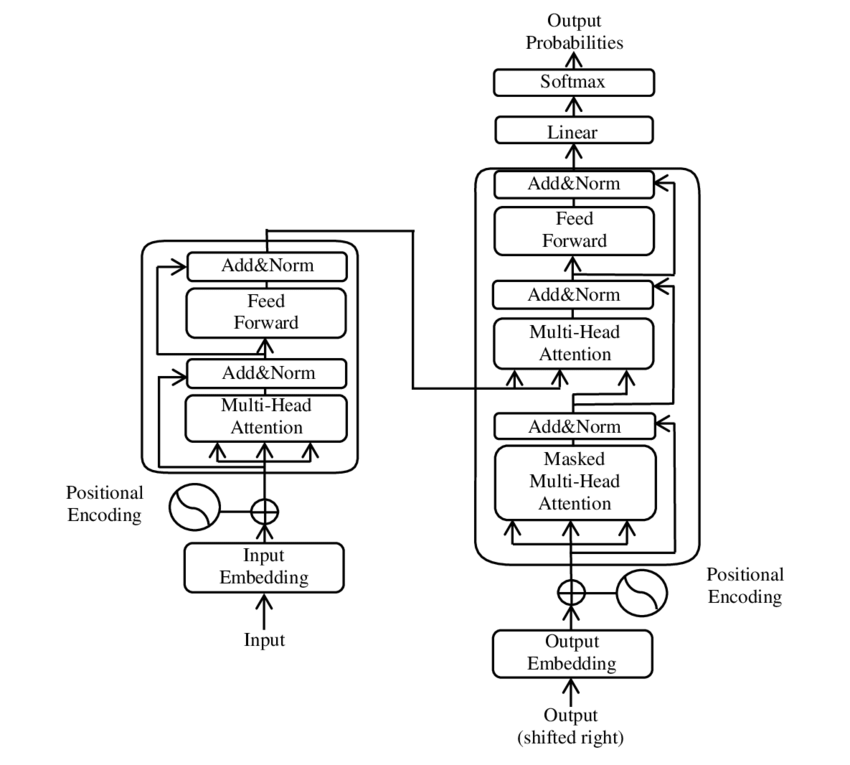}
    \caption[An illustration of the Transformer model architecture.]{An illustration of the Transformer model architecture proposed by \citet{vaswani_et_al_attention}. Reproduced from \citet{jia2019attention}. (Creative Commons Attribution-Share Alike 3.0 Unported license)\vspace{1em}}
    \label{fig:transformer-architecture}
\end{figure}

So-called ``neural'' language models function by parameterizing the language model as an artificial neural network and optimizing these parameters using gradient descent~\citep{bengio2000neural}. At a high level of abstraction, an artificial neural network is a system of linear equations with intermediate non-linear functions, and gradient descent is a supervised training technique (e.g., one that uses explicit training examples) which is based on iteratively optimizing the model to predict the estimated training data by iteratively adjusting the model parameters based on the first-order gradient of the prediction errors. The dominant architecture used for language models is the Transformer architecture shown in Figure~\ref{fig:transformer-architecture}.

While a detailed specification of the Transformer model architecture are presented in Figure~\ref{fig:transformer-architecture}, understanding of these details is not necessary for the content of this thesis. Chapter~\ref{chap:semantic-plausibility} provides slightly more detail of the architecture where relevant to the experiments.

\section{Semantic Plausibility}
\label{sec:lit-review-semantic-plausibility}

As discussed in the introduction, resolving coreference often requires systems to evaluate the \textit{semantic plausibility} of potential interpretations. Intuitively, semantic plausibility captures the relative reasonableness of events described in language—for example, ``using a fork'' is more plausible than ``eating a fork''.

Most work operationalizes plausibility via human judgments~\citep{keller-2003}. Early approaches to modeling semantic plausibility focused on n-gram statistics and information-theoretic measures such as pointwise mutual information~\citep{church-hanks-1990-word}, finding moderate correlations with plausibility ratings for frequent word pairs. Recently, large-scale neural language models based on Transformer architectures have been used to capture semantic plausibility in context-sensitive ways~\citep[e.g.,][]{zhang-etal-2019-knowledge,wang-etal-2020-semeval,liu-etal-2023-vera,kauf-etal-2024-log}. These models are relatively more accurate, but they still struggle with systematic generalization to novel or counterfactual scenarios.

Given its relevance to this thesis and the way the history of modeling semantic plausibility mirrors broader developments in NLP, I provide a more detailed overview of these methods in the remainder of this section.

\subsection{Language as Tuples of Words}

Early work often simplified the problem of modeling the plausibility of text to that of modeling the plausibility of tuples of words, most frequently pairs of words. For example, discriminating the plausibility of the text ``using a fork'' versus ``eating a fork'' might be treated as discriminating the word pairs (\textit{use}, \textit{fork}) versus (\textit{eat}, \textit{fork}). In this review, we focus on models that make this simplifying assumption as these constitute the majority of core contributions in the area. This simplification is still occasionally used in contemporary research.

On the one hand, treating natural language as fixed-length tuples of words is not a necessary simplification for the broader problem of modeling semantic plausibility. In fact, recent state-of-the-art models of semantic plausibility function over variable length sequences of words \citep{zhang-etal-2017-ordinal}, and by restricting our focus to word pairs, the models that we will review miss out on much of the nuance that defines \textit{natural} language.

On the other hand, focusing on modeling the plausibility of word pairs is a consistent assumption across early work in the area; this assumption simplifies the algorithmic design and explanation of the problem; and, even modeling the semantic plausibility of word pairs is a difficult problem that can demand complex commonsense reasoning.

Therefore, we hold this assumption throughout this review. More recent models that discriminate semantic plausibility based on variable length sequences of text are outside our scope, as are the datasets on which these models are evaluated.

It is not necessarily obvious that the occurrence of language can be thought of probabilistically; in fact, early linguistic theories of semantics (i.e. the meaning of language) considered occurrence to be a discrete problem: taking the view that either a word can or cannot occur in a given context \citep{katz-and-fodor-semantics}.

Consider again the language modeling equation~\ref{equation:language-modeling}. This perspective treats occurrence not as some discrete property, but as a distribution over words in context. The plausibility models that we review in this paper are all based on some similar view, which makes sense intuitively: the problem of modeling semantic plausibility is closely related to the problem of modeling the likelihood of language being attested. As ``using a fork'' is more semantically plausible than ``eating a fork,'' so too is the former more likely to be attested in a corpus.

That is not to say that the two problems are identical. Consider, for example, that ``a person breathing'' is more semantically plausible than ``a person dying,'' whereas ``a person dying'' is more likely to be attested in a corpus. This is because written language is used to communicate, and the event of breathing is obvious and thus rarely worth communicating (despite occurring frequently in the world). The discrepancy between what is frequent in text and what is likely in the world due to obvious events being rarely attested is known as \textit{reporting bias} \citep{Gordon:2013:RBK:2509558.2509563}.

Shannon uses maximum-likelihood estimation (MLE) to estimate these $n$-gram probability distributions. In this case, MLE corresponds to estimating likelihood by empirical counts in a corpus.

Probabilities of occurrence do correlate with semantic plausibility. \citet{keller-2003} show that MLE probabilities calculated from a large web corpus have a Spearman $\rho$ of 0.6 with human plausibility judgments for frequently occurring word pairs. However, for infrequent word pairs the correlation is close to zero.

\subsection{Pointwise Mutual Information}
\label{sec:pmi}

This leads us to the question, do information-theoretic statistics of occurrence probabilities signify any semantic relationships? In their influential paper, \citet{church-hanks-1990-word} provide an answer to this question.

They show qualitatively that pointwise mutual information seems to correlate with the semantic relatedness of words. This finding motivates the use of information-theoretic statistics in models of plausibility such as that of \citet{resnik-1997-selectional}.

Given words $x$ and $y$, \citet{church-hanks-1990-word} take pointwise mutual information to be
\begin{equation}
     \texttt{pmi}(x; y) = \log \frac{P(x,y)}{P(x)P(y)}
\end{equation}
where $P(x)$ and $P(y)$ represent the unconditional probability of the respective words occurring in some corpus, and $P(x,y)$ is the probability of $x$ and $y$ occurring together.

Church \& Hanks consider the case that the words occur in some particular syntactic relation in a sentence as a case of occurring \textit{together}. Syntactic relations are simply a class of directed relations between words in a sentence, such as \textit{subject} and \textit{direct object}. For example, in the sentence ``Jane eats the spaghetti.'' the word ``spaghetti'' is the direct object of the verb ``eat.'' By Church \& Hanks' definition of co-occurrence, we would say that ``eat'' and ``spaghetti'' occur together in this sentence.

Intuitively, pointwise mutual information then tells us how often $x$ and $y$ occur together in the given syntactic relation as compared to if they were independent and only co-occurred by chance.

Church \& Hanks extract 4 million transitive sentences from a corpus, and they represent each sentence as a triple of the subject, verb, and object, referred to as an s-v-o triple. For example, the s-v-o triple corresponding to the sentence ``Jane eats the spaghetti.'' is \textit{(Jane, eat, spaghetti)}.

For a given word $x$, they then calculate the mutual information between $x$ and each word in the vocabulary of the corpus. To do this, they estimate $P(x)$ and $P(x,y)$ by occurrence counts
\begin{equation}
     P(x) \approx \frac{\texttt{Count}(x)}{N}
\end{equation}
\begin{equation}
     P(x,y) \approx \frac{\texttt{Count}(x,y)}{N}
\end{equation}\\[0.1em]
where $N$ is the number of occurrences of any word in the corpus (i.e. the corpus length), $\texttt{Count}(x)$ is the total number of occurrences of word $x$, and $\texttt{Count}(x,y)$ is the number of times word $x$ and $y$ co-occurr in the given syntactic relation.

They find that the words $y$ that have relatively high pointwise mutual information with word $x$ are usually semantically related. For example, considering the direct object relationship, verb $x=\text{``answer''}$ has high mutual information with object $y=\text{``telephone''}$ and verb $x=\text{``drink''}$ has high mutual information with object $y=\text{``coffee.''}$.

This work importantly demonstrates that information-theoretic, count based corpus statistics appear correlated with semantics. And yet, there are still many questions left unanswered: how well does mutual information correlate with semantic plausibility in particular? High mutual information seems to imply high semantic relatedness, but does the inverse relationship hold?

I now review influential, computational models of semantic plausibility. I consider models of two broad categories: first, I describe class-based models (Section \ref{sec:class-based-models}) which focus on ways of modifying corpus probabilities using a human-labeled hierarchy of words to better correlate with human plausibility judgements. Then, I consider purely-distributional models (Section \ref{sec:purely-distributional-models}) which use no human annotations outside of raw corpus of text; instead, these models rely only on raw text corpora to learn models of plausibility. The majority of these methods further rely on vector representations of words which are called \textit{word embeddings}.

These models all focus on the simplified problem of modeling the semantic plausibility of word pairs as opposed to variable-length sequences of words. Similar to the work of \citet{church-hanks-1990-word} (Section \ref{sec:pmi}), the word pairs considered are specifically those of two words in a particular syntactic relation. For example, a model is often trained specifically to model the plausibility of verb-object pairs.

In principle, this could be any syntactic relation; therefore, for simplicity, we focus on the problem of modeling the plausibility of \textit{(verb, direct object)} pairs of words. We denote $x$ the verb and $y$ the direct object. 

In the literature, the verb and object in a dependency relation are often referred to as the predicate and argument, respectively. Also in the case of word pairs, semantic plausibility is often referred to as \textit{selectional preferences}, \textit{selectional constraints}, or \textit{thematic fit}.

The models that we consider all follow the same general form. They take as input a word pair corresponding to some particular relation, in our case verb-object, and output a scalar representing the relatively plausibility of this pair.

There are several methods for evaluation, the most popular of which is the rank correlation (e.g. Spearman's $\rho$ or Kendall's $\tau$) with human plausibility judgments.

The next most common evaluation is to use plausibility estimates to somehow improve performance of an NLU system downstream, but again, that is outside the scope of this review.

\vspace{1em}

\begin{figure}[ht]
    \centering
    \includegraphics[width=8cm]{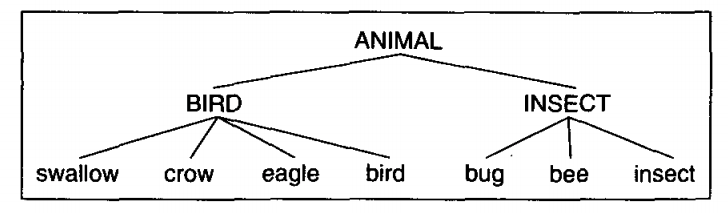}
    \caption[An example semantic hierarchy.]{An example semantic hierarchy. Reproduced from \citet{li-abe-1998-generalizing}.
    (Creative Commons Attribution-NonCommercial-ShareAlike 3.0 International License)}
    \label{fig:hierarchy}
\end{figure}

\subsection{Class-based models}
\label{sec:class-based-models}

The following class-based models make use of word classes in a semantic hierarchy. A semantic hierarchy---in some literature referred to as a lexical hierarchy, a thesaurus, or a taxonomy---is a manually-curated hierarchy, specifically a tree or a directed acyclic graph (DAG), where nodes correspond to words, and a parent-child relationship implies that the child node ``is a'' parent node. In other words, the child node is a subclass of the parent node, or alternatively child and parent are in an \textit{IsA} relationship. An example hierarchy is presented in Figure \ref{fig:hierarchy}.

The IsA relationship between words specifically is a semantic relationship known as \textit{hypernymy}. Given words $x_1$ and $x_2$ where $x_1$ IsA $x_2$, $x_1$ is referred to as the hyponym, and $x_2$ as the hypernym.

In practice, these methods that we review all use the WordNet hierarchy \citep{wordnet}, but the design decisions could apply to any semantic hierarchy of words. We denote the set of concepts in WordNet as $\mathcal{C}$.

A semantic-hierarchy is a common tool in modeling semantic plausibility for two primary reasons:
\begin{enumerate}
    \item This allows us to overcome the limitations of data sparsity. While verb-object pair \textit{(eat, meatball)} may not be attested in the corpus, likely \textit{(eat, food)} and \textit{(eat, apple)} are attested. If \textit{meatball} is a type of \textit{food} in the, we can smooth probabilities across the hierarchy to overcome this sparsity problem.
    \item Similarly, abstracting to concept classes may help account for the distinction between likelihood and plausibility caused by reporting bias. Following the same reasoning of the prior example, \textit{(breathe, nitrogen)} may not occur in the corpus, despite being a very common occurrence in the world, while \textit{(breathe, gas)} does.
\end{enumerate}

\subsubsection{Resnik}
\label{sec:resnik}

\citet{resnik-1997-selectional} presents the first model that we consider which is the basis of all future work. This model uses information-theoretic statistics to approximate plausibility judgments, and attempts to find an appropriate level of abstraction in the semantic hierarchy for the verb-object pair. The right abstraction of \textit{(eat, meatball)} might be found to be \textit{(eat, food)}, for example, in which case both pairs will take the same plausibility rating.

For verb $x$ (e.g. ``eat''), given a set of concepts $\mathcal{C}$ (e.g. the set of WordNet concepts), Resnik first defines the \textit{selectional strength} of $x$ as 
\begin{equation}
     \texttt{SelStr}(x) = D_\text{KL}({P(c|x)} \lVert {P(c)}) = \sum_{c \in \mathcal{C}} P(c|x)\log\left(\frac{P(c|x)}{P(c)}\right)
\end{equation}
where $P(c|x)$ is the probability of $x$ taking a direct object of the conceptual class $c$. Selectional strength intuitively describes how much $x$'s distribution of arguments diverges from the prior distribution; e.g., we would expect the selectional strength of ``eat'' to be high as the distribution of concepts that co-occur diverges greatly from the prior distribution of all concepts.

Resnik then defines the \textit{selectional association} of predicate $x$ and concept $c$ as
\begin{equation}
\texttt{SelAsso}(x,c) = \frac{P(c|x) \log\left(\frac{P(c|x)}{P(c)}\right)}{\texttt{SelStr}(x)}
\end{equation}
Intuitively, this describes how much concept $c$ contributes to the total selectional strength of $x$; e.g., again considering the predicate ``eat,'' selectional association would be high for concepts that co-occur in the direct object position, such as [\textsc{food}], and low for concepts that don't, such as [\textsc{fork}].

\begin{figure}[ht]
    \centering
    \includegraphics[width=10cm]{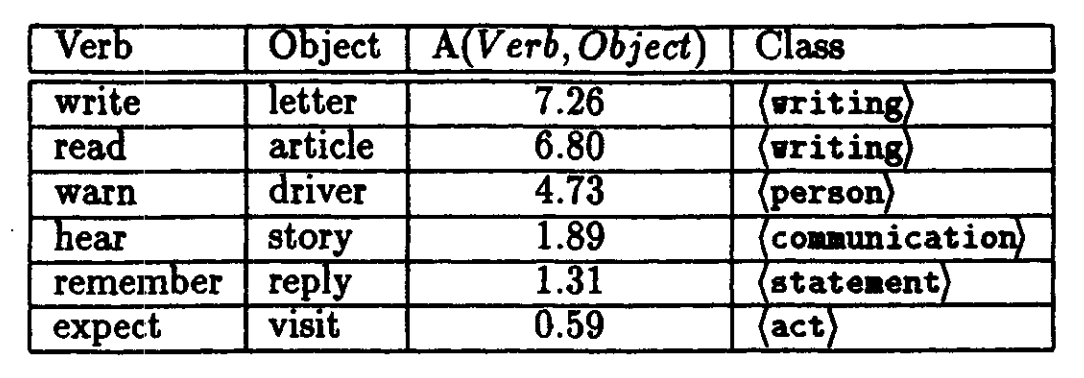}
    \caption[Example semantic plausibility scores.]{Example semantic plausibility scores and the maximizing class. Reproduced from \cite{resnik-1997-selectional}. (Creative Commons Attribution-NonCommercial-ShareAlike 3.0 International License)}
    \label{fig:resnik}
\end{figure}

Finally, the semantic plausibility of verb $x$ for direct object $y$ (e.g. ``meatball'') is taken to be
\begin{equation}
\texttt{SemPlaus}_{\texttt{Resnik}}(x,y) = \max_{\{ {c \in \mathcal{C}} \; \mid \; {y \text{ IsA } c} \}} \texttt{SelAsso}(x,c)
\end{equation}
i.e., the maximum selectional association over all hypernyms of $y$. The hope is that the maximizing concept $c$ is at the appropriate level of abstraction, and thus the plausibility of $x$ and $y$ is calculated for this abstraction and not the surface form of $y$. Figure \ref{fig:resnik} shows example values and found levels of abstractions for object-verb pairs.

Resnik empirically estimates probabilities by MLE using occurrence counts in a corpus. For a word in the corpus vocabulary, its occurrence counts are smoothed over all concepts to which it belongs. This allows for calculating $P(c|x)$ and $P(c)$.

\subsection{Purely-distributional Models}
\label{sec:purely-distributional-models}

Another class of models estimate semantic plausibility based only on co-occurrences in raw text corpora without the need for manually curated forms of knowledge.

The distributional hypothesis is that words in similar contexts have similar meanings \citep{harris_doi:10.1080/00437956.1954.11659520}. A model that learns meaning based on the distribution of co-occurrences in text is said to be a distributional model; hence, we refer to this class of models as \textit{purely-distributional models}.

\subsubsection{Semantic Roles}

The simplest distributional estimate of the semantic plausibility of verb-object pair $(x,y)$ may be the MLE estimate of $P(y|x)$. As briefly mentioned, \citet{keller-2003} has shown that for a large web corpus, and a typical pair $(x,y)$, this conditional probability does somewhat correlate with human plausibility judgements.

\cite{Pad2009APM} present a probabilistic model of semantic plausibility that is almost just as simple. The primary distinction being that they consider the semantic role of the object.

Semantic roles are similar to syntactic dependencies, but intuitively reflect higher-level relations between words in a sentence. Consider the sentence ``Jane broke the window.'' where ``window'' is the direct object of ``break,'' and the sentence ``The window broke.'' where ``window'' is the subject of ``break.'' While the syntactic relation changes between these sentences, the semantic role that the window is playing in the event is the same. In this case, we would say that the window is the patient of the verb break in both cases.

Pado et al. use a semantic role labeling system to obtain the semantic roles in their corpus. They then calculate MLE estimates of $P(y|r,x)$ and $P(r|x)$ where $r$ is the semantic role of $y$, and finally take semantic plausibility to be
\begin{equation}
    \texttt{SemPlaus}_{\texttt{Pado}}(x,y) = P(y|x) = P(y|r,x)P(r|x)
\end{equation}

\subsubsection{Exemplars}

\citet{erk-2010} present a more recent distributional model of semantic plausibility based on word embeddings. They describe this model as \textit{exemplar based} as it calculates semantic plausibility based on a set of exemplar examples taken from the training corpus.

Intuitively, this model considers how similar an object's embedding is to the embeddings of other objects that have been attested with verb $x$.

Specifically, Erk et al. calculate the semantic plausibility of word $y$ given paired word $x$ as
\begin{equation}
    \texttt{SemPlaus}(y|x) = \sum_{y' \in \texttt{Seen}(x,y')} \frac{\mu(a)}{Z} \cdot \texttt{sim}(y, y') 
\end{equation}
where $\texttt{Seen}(x,y)$ is the set of pairs attested in the training corpus, $\mu(\cdot)$ is some weighting function used to weight more frequent examples, and $Z$ is a normalizing constant.

$\texttt{sim}(y, y')$ is a similarity measure of the word embeddings of $y$ and $y'$, in this case cosine similarity.

In practice, this method could be applied to any type of word embedding. The standard word vectors used by Erk et al. are created with the following approach:
\begin{itemize}
    \item Select a set of context words, typically the $k$ most common words in the corpus (not including overly-generic words, called stop words, such as ``the'' and ``and'').
    \item For each word in the vocabulary, calculate the pointwise mutual information between this word and all context words. The vector embedding of the word $y$ then becomes a vector of these pmi statistics.
\end{itemize}

\subsection{Current limitations}

Even more recently, \citet{wang-etal-2018-modeling} demonstrate that state-of-the-art approaches to semantic plausibility of object-verb pairs still fail on many cases that require commonsense knowledge. The probabilities inferred by large language models correlate with semantic plausibility~\citep{kauf-etal-2024-log}, and are relatively accurate compared to baseline methods when finetuned for the task~\citep{liu-etal-2023-vera}, but still fail to capture the diversity of human inference ability \citep{kauf2023event}.

\chapter{Challenge Sets to Canonical Benchmarks}
\label{chap:challenge-set-assumption}

\setlength{\epigraphwidth}{0.8\textwidth}
\myepigraph{When AI can’t understand the meaning of “it,” it’s hard to believe it is poised \\ to take over the world.}{\citet{etzioni_2020}}

Now we have the general understanding needed for approaching the core arguments of this thesis. In this first core chapter, I will begin to unify the evaluations of ``canonical'' and semantic-plausibility based evaluations which will expose issues with convergent validity~\citep{jacobs2021measurement}. That is, different evaluations intended to measure similar background concepts lead to distinct and contradicting conclusions. At the same time, these evaluations may begin to expose possible brittleness in language models as capable of encoding the necessary information needed for resolving coreferences which will be revisited more thoroughly in later chapters.

With respect to evaluations based on semantic plausibility, I will consider existing challenge sets. Challenge sets such as the Winograd Schema Challenge (WSC) are used to benchmark systems' ability to resolve ambiguities in natural language. At the same time, I will consider as ``canonical'' evaluations those such as agreement with the expert annotations in OntoNotes and related datasets.

If one is to assume, as in existing work, that solving a given challenge set is at least as difficult as solving some more general task, then high performance on the challenge set should indicate high performance on the general task overall. Specifically for coreference resolution, this means that high performance on the WSC and the like should indicate that the same model exhibits high performance on examples of pronominal coreference in natural language corpora that are of a similar a form that are believed to be trivial.

However, I show empirically that this assumption of difficulty does not always hold. In particular, I demonstrate that despite the strong performance of prompted language models (LMs) on the WSC and its variants, these same modeling techniques perform relatively poorly at resolving certain pronominal ambiguities attested in OntoNotes and related datasets that are perceived to be easier.

Motivated by these findings, I propose a method for ensembling a prompted LM with a supervised, task-specific system that is overall more accurate at resolving pronominal coreference across datasets as per standard evaluation metrics. This ensembling method is not necessary intended to be an empirically useful model, but rather highlights conflicting conclusions in terms of model quality between these two evaluations.

This finding is of practical significance for how we evaluate and develop NLP systems. It implies that we should not rely exclusively on any single dataset, even if it is considered particularly difficult or challenging. Instead, comprehensive evaluation requires testing on all available datasets designed for the task, to ensure models generalize across the full range of phenomena.

Moreover, this raises further challenges for evaluation design. For example, older benchmark datasets are more likely to have been seen during pre-training of large language models, introducing potential contamination or leakage that can inflate apparent performance. Addressing such issues requires careful curation of evaluation data and more robust testing methodologies that account for training data overlap.

Finally, I emphasize that datasets involving the same linguistic phenomenon draw on distinct, but overlapping, capabilities, and evaluating on any one dataset alone does not provide a complete picture of a system's overall capability. Thus, we as a community need to consider broader meta-aspects of evaluations such as the convergent validity to achieve a fuller-picture of model performance.

\begin{figure}[t] 
\centering
  \includegraphics[width=0.95\textwidth]{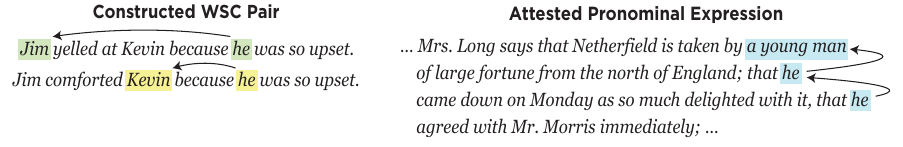}
  \caption[Examples of pronominal anaphora.]{
  \textit{Left}: An example minimal pair from the WSC. \textit{Right}: Pronouns attested in the novel \textit{Pride and Prejudice} and annotated for coreference by \citet{vala-etal-2016-annotating}.
  }
  \label{fig:csa-intro-example}
  \vspace{-5pt}
\end{figure}

\section{Overview of Methodology}
\label{sec:csa-introduction}

As outlined previously, the Winograd Schema Challenge~\citep[WSC;][]{levesque_winograd_2012} is a prominent challenge set of ambiguous \ac{PCR} problems, one of many such challenge sets used to evaluate NLP systems \citep[\textit{e.g.,}][]{isabelle-etal-2017-challenge,clark2018think,mccoy-etal-2019-right}. Challenge sets are typically constructed to consist of difficult instances of a more general task. Often, a system's performance on such a challenge set is considered in isolation from its performance on the broad range of expressions attested in natural corpora (I use the term \textit{natural corpora} to refer to text that was not explicitly constructed or elicited for research purposes. An \textit{attested} expression is one appearing in natural corpora in contrast to \textit{constructed} expressions that commonly compose challenge sets). For instance, systems are frequently evaluated on the WSC without a corresponding analysis of how they perform on a diverse range of attested pronominal expressions~\citep[][\textit{i.a.}]{kocijan-etal-2019-surprisingly,shen-etal-2021-unsupervised,eval-harness,achiam2023gpt}. This narrow evaluative focus can obscure a system's ability to generalize.

The WSC specifically consists of minimal pairs of sentences, each containing an ambiguous pronoun (Figure~\ref{fig:csa-intro-example}). These pairs are manually designed such that consistently disambiguating the pronouns is thought to require the commonsense world knowledge and reasoning abilities a human reader might employ. Following the recent success of language model (LM) based approaches at resolving WSC instances, some of the challenge's original authors have declared it to be solved~\citep{Kocijan_2023}.

Despite this success on a standard benchmark, this chapter demonstrates that the same LM-based systems exhibit fragility, proving relatively inaccurate at resolving certain ambiguous pronominal expressions attested in natural corpora from OntoNotes~\citep{hovy2006ontonotes} and related datasets. This result may seem surprising, given that ``the point of the WSC is to test programs that claim to have solved the problem of pronoun reference resolution''~\citep{Kocijan_2023} and that its instances are considered to be relatively difficult examples of PCR~\citep{peng-etal-2015-solving}.

The experiments in this chapter focus on prompted LMs as systems that perform well on Winograd schemas; among these, we primarily use the Llama family of models~\citep{touvron2023llama,dubey2024llama}, though we also present evidence that our results generalize across other LM families, including OLMo~\citep{groeneveld-etal-2024-olmo} and Mistral~\citep{jiang2023mistral}. We compare the performance of prompted LMs up to 70B parameters against state-of-the-art coreference resolution systems, such as Maverick~\citep{martinelli-etal-2024-maverick}, which are known for their accuracy on attested pronominal coreferences.

The evaluation is conducted across 11 datasets. Six of these datasets contain PCR problems from text attested in natural corpora, such as OntoNotes 5.0~\citep{ontonotes5} and OntoGUM~\citep{zhu-etal-2021-ontogum}. The remaining five datasets consist of PCR problems constructed for WSC-like challenge sets, including Winogrande~\citep{sakaguchi2021winogrande} and DPR~\citep{rahman-ng-2012-resolving}.

A comparison with unsupervised baselines shows that LMs are generally more accurate across all datasets. However, a key finding emerges: \textit{whereas supervised coreference resolution models perform relatively poorly on the WSC, these same systems are more accurate than prompted LMs at resolving certain attested pronouns}. This poor generalization for LMs is consistent across test sets of diverse annotation guidelines and textual genres.

Motivated by these results, this chapter proposes a method for ensembling a prompted LM with a task-specific system to achieve a final system that is more accurate overall. This ensembling method heuristically identifies salient discourse entities, for which coreference is disambiguated by a state-of-the-art coreference system trained on OntoNotes. The remaining instances are then disambiguated using an LM prompted with in-context examples. The resulting hybrid system is, in most cases, more accurate at resolving pronouns across both attested expressions and WSC-like challenge sets.

Ultimately, the findings presented in this chapter illustrate that datasets targeting the same linguistic phenomenon can draw on distinct capabilities. This supports the central argument that conclusions drawn from controlled evaluation settings are often limited; no single dataset can provide a complete picture of a system's generalizability. It is therefore crucial that challenge set results be considered in conjunction with evaluations that encompass a diverse range of attested expressions from varied contexts.

\section{Methodology: Background}
\label{sec:csa-related-work}

Pronominal coreference resolution (PCR) can be broadly defined as the task of determining which linguistic expressions refer to the same discourse entity as a given pronoun \citep{hobbs1978resolving}. For comprehensive surveys on this topic, see \citet{zhang-etal-2021-brief} and \citet{poesio2023computational}.

Historically, systems for resolving pronominal coreference relied on heuristic rules, often combined with unsupervised statistical patterns over handcrafted features~\citep[][\textit{i.a.}]{poon-domingos-2008-joint,charniak-elsner-2009-em,raghunathan-etal-2010-multi,lee-etal-2011-stanfords}. More recently, a variety of LM-based approaches have been introduced, including: LMs finetuned on supervised training data~\citep{zhang-etal-2019-knowledge,zhao-etal-2022-pcr4all}, weakly supervised LMs~\citep{kocijan-etal-2019-wikicrem,shen-etal-2021-unsupervised}, and the prompting of LMs by formatting PCR as either a cloze task~\citep{trinh2018simple,radford2019language} or a question-answering task~\citep{brown2020language,wang-etal-2022-super,le2023large,zhu-etal-2024-large}.

Evaluations of a system's ability to perform PCR have generally fallen into three categories: 1) assessments on collections of ambiguous pronouns attested in natural text~\citep{hobbs1978resolving,lappin-leass-1994-algorithm,webster-etal-2018-mind}, 2) evaluations on the pronominal coreference subsets of larger coreference resolution datasets~\citep{martschat-strube-2014-recall,zhang-etal-2019-knowledge,lu-ng-2020-conundrums}, and 3) performance on challenge sets composed of WSC-like instances~\citep{rahman-ng-2012-resolving,emami-etal-2019-knowref,sakaguchi2021winogrande}.

Researchers studying the general task of coreference resolution have adopted the WSC and its derivatives as challenge sets to complement canonical evaluations like OntoNotes~\citep{peng-etal-2015-solving,toshniwal-etal-2021-generalization,zhao-etal-2022-pcr4all}. Such work has consistently shown that systems designed for general-purpose coreference resolution perform poorly on the WSC. Following this line of work, this chapter views WSC-like datasets as specialized challenge sets for PCR.

Recent advances in language modeling have yielded systems that are highly accurate at resolving WSC instances, in some cases approaching human-level performance~\citep{brown2020language,wei2022finetuned,touvron2023llama}. However, a growing body of evidence shows that these same techniques are less accurate than supervised models when evaluated on established benchmarks for the general task of coreference resolution~\citep{yang-etal-2022-gpt,le2023large,zhu-etal-2024-large,gan-etal-2024-assessing-capabilities}. The analysis in this chapter diverges from these studies by focusing specifically on PCR, rather than the broader and more ambiguously defined concept of coreference~\citep{recasens-hovy-2010-coreference,can_we_fix}, to more precisely investigate this performance discrepancy.

\begin{figure}[t]
\centering
  \includegraphics[width=0.48125\textwidth]{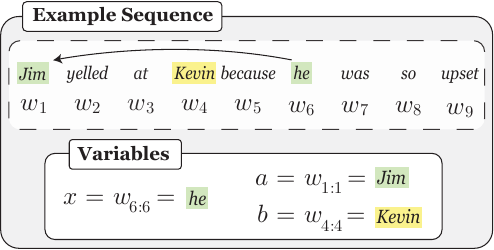}
  \caption[An example formulation of pronominal anaphora.]{
  An example instance and the corresponding variables: the pronoun $x$, antecedent $a$, and distractor candidate $b$. \vspace{1em}
  }
  \label{fig:csa-formulation-example}
\end{figure}

\section{Details of Methodology}
\label{sec:csa-method}

This section formulates the problem of pronominal coreference resolution (PCR) and provides a high-level description of how system accuracy is evaluated. It also formalizes the assumptions commonly made when evaluating on challenge sets so that these assumptions can be explicitly tested in the subsequent analysis.

\subsection{Problem Formulation}

The task of PCR is formulated as follows: given a text passage $w = (w_1, \ldots, w_t)$, the goal is to resolve a pronominal expression $x$ to its correct antecedent $a$, where $x$ and $a$ are subspans of $w$.
This chapter investigates a restricted version of this problem, framed as a binary classification task.

More explicitly, the task assumes that for a given pronoun $x$, exactly one of two candidate antecedents in $w$ is the correct resolution. This binary-choice formulation is flexible enough to accommodate both WSC-style problems and instances derived from coreference annotations occurring in natural corpora. Formally, given $w$, $x$, and a set of two candidate antecedents $\{\hat{a}_1, \hat{a}_2 \}$, the objective is to correctly identify which candidate corresponds to the true antecedent $a = w_{k:l}$. The other candidate is a distractor mention $b = w_{m:n}$ that refers to a discourse entity but does not corefer with $x$. An example of this formulation is provided in Figure~\ref{fig:csa-formulation-example}.

\subsection{The Challenge Set Assumption}

A core assumption of the WSC is that its instances are inherently more difficult to resolve than other instances of the PCR task, particularly those attested in natural corpora. This chapter formulates this "challenge set assumption" formally (Def.~\ref{def:the-challenge-set-assumption}) and proceeds to test it empirically by comparing the performance of various systems across both challenge set instances and attested pronouns.

To formalize this assumption, let $C$ be a challenge set and $D$ be some other dataset representing the same task. Furthermore, let $\theta$ and $\phi$ be two systems that are to be evaluated on this task. A function $U$ represents the performance measure of a system on a given dataset; for example, the performance of system $\theta$ on dataset $C$ is denoted as $U(\theta, C)$.

\begin{definition}[The Challenge Set Assumption]
\label{def:the-challenge-set-assumption}
The ordering of model performance on the challenge set $C$ is preserved on dataset $D$. That is, $U(\theta, C) > U(\phi, C) \implies U(\theta, D) > U(\phi, D)$.
\end{definition}

Intuitively, the assumption is that because $C$ is strictly more difficult than $D$, systems that are relatively accurate on $C$ should be relatively accurate on $D$ as well.

\subsection{Evaluating Performance}

To test the assumption that high performance on challenge sets generalizes to other contexts, systems are evaluated across multiple test sets. This section describes how the performance function $U$ is calculated for different types of data.

\paragraph{Attested Pronominal Expressions.} The evaluation on attested pronominal expressions begins with existing datasets of identity coreference relations annotated in natural corpora. From these annotations, all mentions that participate in a coreference relation and match a predefined set of pronouns are identified as a coreferring pronominal expression, $x$. (The following strings are considered pronouns: "she", "her", "he", "him", "them", "they", "it", "his", "hers", "its", "their", "theirs", "this", "that", "these", "those".) The text passage $w$ is defined as the concatenation of the sentence in which $x$ occurs with the two preceding sentences. For each expression $x$ that has a single coreferring nominal antecedent $a$ within the context $w$, and for which at least one non-coreferring expression $b$ also occurs in $w$, a test instance is created. In cases where multiple candidates for $b$ are available, one is randomly sampled. Performance is then measured based on a system's accuracy in resolving these constructed instances.

This formulation, including the predefined set of pronouns, is consistent with the conventional setup for PCR used in existing work~\citep{yang-etal-2003-coreference,ng2005supervised,li-etal-2011-pronoun,zhang-etal-2019-knowledge,zhao-etal-2022-pcr4all}.

\subsection{Differences in Formulations of PCR}
\label{sec:csa-evaluation-differences}

The formulation of PCR employed in this chapter follows the precise setup proposed by \citet{zhang-etal-2021-brief}, which was in turn based on earlier work that also considered fixed subsets of English pronouns in restricted contexts. Such restrictions are justified by the observation that most antecedents occur within the local context of a pronoun; for example, \citet{yang-etal-2003-coreference} observed that the antecedent is within the local context 95\% of the time in the MUC corpus.

To ensure a fair comparison across dataset types, the WSC-like challenge sets were formatted in the same manner. This required normalizing the data, as WSC-like datasets may initially contain pronominal expressions outside the considered set, such as \textit{one} or \textit{y'all}.

For additional details, the preprocessing scripts used to standardize all datasets are made publicly available for reproducibility.\footnote{See \url{https://github.com/ianporada/challenge-set-assumption}.}

\subsection{Additional Considerations}

This section outlines key differences in how existing work has approached PCR and clarifies the choices made in setting the scope of this analysis.

A fundamental tradeoff exists between evaluating all forms of pronominal coreference that might occur in natural language and evaluating only those forms that have been defined such that they can be reliably annotated. The goal of this analysis is more oriented towards the latter. That is, this work does not attempt to analyze all conceivable coreferences of all possible pronominal expressions. Rather, the focus is on the intersection of existing work to better understand how well models generalize across established datasets.

\paragraph{End-to-end vs. mention-linking:} As an end-to-end task, the goal of PCR is to identify which linguistic expressions a pronoun corefers with, given only the raw context and the pronoun itself. In contrast, the task can be framed as mention-linking, where candidate antecedent mentions are pre-identified, and the system's task is to select the correct one. Common approaches for mention-linking involve scoring each candidate independently or pairwise~\citep{yang-etal-2008-twin}. In this analysis, systems are evaluated according to the formulation for which they are designed.

\paragraph{One vs. many mentions:} A single discourse entity can be realized as multiple coreferring expressions, forming a coreference cluster. When multiple such expressions appear in the context of a pronoun, there are several possible interpretations of the correct antecedent. Popular approaches include selecting the most recent mention \citep{liang-wu-2003-automatic} or accepting any one of the coreferring mentions as valid. For consistency with the most common approach in existing work~\citep{zhang-etal-2021-brief}, the most recent mention is considered the valid antecedent here. Furthermore, to simplify the evaluation to binary classification, instances where multiple coreferring expressions appear within the immediate context are not included in this analysis.

\paragraph{Mention boundaries:} Datasets differ in their annotation of mention boundaries~\citep{moosavi-etal-2019-using}. For example, the antecedent noun phrase ``a young man'' in Figure~\ref{fig:intro-example} is annotated in one dataset as simply ``man,'' whereas in OntoNotes it would be annotated as the maximal span ``a young man of large fortune from the north of England.'' For end-to-end PCR models, a reasonable assumption is that the predicted mention should contain the head word of the correct antecedent and not the head word of any incorrect antecedent~\citep{crac-2022-crac}. However, optimizing for head words in this manner can lead to modeling decisions that do not align with human intuition~\citep{crac-2023-crac}. \citet{moosavi-etal-2019-using} present a method for normalizing mention boundaries to a minimal span, which could be useful for future work. For mention-linking tasks in this chapter, the dataset's provided annotated mention is used. More complex phenomena, such as split-antecedents and discontinuous mentions, are not considered in this analysis but present interesting avenues for future investigation.

\paragraph{WSC-like Challenge Sets.} WSC instances generally follow the formulation of PCR outlined above; the basic premise is that some text $w$ contains a pronoun $x$ with two candidate antecedents $\{\hat{a}_1, \hat{a}_2 \}$. Minimal formatting is therefore performed on existing WSC-like challenge sets so that examples are in the same form as those from the attested datasets. This requires tokenization and, in some cases, determining the exact span of candidate mentions, with further details provided in App.~\ref{sec:csa-evaluation-differences}. This consistent formatting allows for the direct computation of accuracy as the ratio of instances where the system predicts the correct candidate antecedent.

\section{Findings: Experiments}
\label{sec:csa-experiments}

This section describes the experimental setup in detail, which is designed to test whether the challenge set assumption holds empirically. A comparison is made between prompted LMs known to be accurate on the WSC and task-specific systems known to be accurate at resolving certain attested pronouns (\S\ref{sec:csa-models}). All systems are evaluated across 11 datasets that span both attested and WSC-like instances (\S\ref{sec:csa-datasets}).

\subsection{Systems}
\label{sec:csa-models}

\subsubsection{Prompted Language Models}

In recent years, prompted LMs have demonstrated high accuracy on the WSC. If the challenge set assumption holds, one would therefore expect such systems to be relatively accurate at resolving attested pronominal expressions. Prompted LMs function by predicting the correct antecedent span $a$ for a given problem instance $(w, x, \{\hat{a}_1, \hat{a}_2 \})$, which is formatted using a textual prompt template that may also include in-context examples.

\paragraph{Llama 3.1} The primary prompted LMs used in this analysis are from the Llama 3.1 family of models at various sizes~\citep{dubey2024llama}, which are competitive open-weights LMs. Experiments specify whether the base or instruct-tuned version is used. The instruct versions were additionally finetuned on instruction-tuning data, like the Flan collection \citep{pmlr-v202-longpre23a}, and human preference annotations. The 8B and 70B parameter model sizes are evaluated.

In experiments involving few-shot prompting, a comparison is also made against a supervised Llama 3.1 8B model, which was finetuned to resolve the WSC by training on public training sets formatted with a QA prompt.

\paragraph{Additional LMs} Performance is additionally compared with the smaller Llama 3.2 models, the fully open-source OLMo model~\citep{groeneveld-etal-2024-olmo}, and the Mistral-NeMo 12B parameter model~\citep{mistralNemo}.

\begin{figure*}[t]
\centering
  \includegraphics[width=0.95\textwidth]{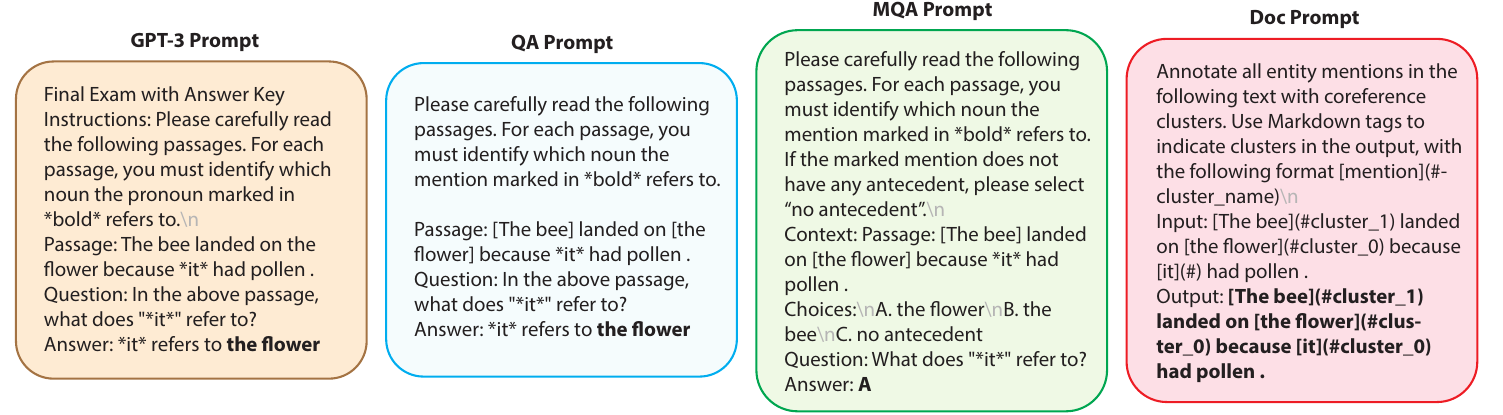}
  \caption[A training set instance from the Definite Pronoun Resolution (DPR) dataset formatted using each of the corresponding prompts.]{A training set instance from the Definite Pronoun Resolution (DPR) dataset~\citep{rahman-ng-2012-resolving} formatted using each of the corresponding prompts. Denoted in bold is the expected model output. The GPT-3 prompt~\citep{brown2020language} does not rely on gold mention span annotations. QA Prompt and Doc Prompt were presented by \citet{le2023large}. The multiple-choice QA (MQA) prompt was presented by \citet{zhu-etal-2024-large}.
  }
  \label{fig:csa-prompts}
\end{figure*}

\paragraph{Prompting Techniques.}

The goal of this chapter is not to propose new prompting techniques; therefore, the experiments utilize four existing prompt templates sourced from the literature. These templates are shown in Figure~\ref{fig:csa-prompts}. The GPT-3 prompt, used by \citet{brown2020language} for evaluating GPT-3 on the SuperGLUE WSC~\citep{wang2019superglue}, does not require gold mention annotations. For this prompt, a string match of the model output is checked, following \citet{brown2020language}. The other prompts (QA, MQA, and Doc) were proposed for using language models to perform coreference resolution and require explicit candidate mention spans. For these prompts, the candidate output with the highest probability assigned by the language model is taken to be the model's prediction. Another common approach is to formulate WSC-like instances as a cloze task~\citep{trinh2018simple,eval-harness}. This prompting technique is not considered, however, as it is incompatible with pronominal references whose resolution depends on the grammatical features of the pronoun.

Prompted LMs are evaluated in zero- and few-shot settings, depending on the comparison being made. In the zero-shot setting, the LM is prompted only with the relevant instructions and input passage. In the few-shot setting, the instruction-tuned versions of the Llama 3.1 models are used with 32 training instances prepended to the input.

\subsubsection{Task-Specific Systems}

The performance of prompted LMs is compared against the following task-specific systems, which were designed for the general problem of coreference resolution. Such models are generally believed to perform poorly on the WSC. One would therefore expect prompted LMs to outperform these task-specific systems across all PCR datasets, assuming \textit{the challenge set assumption} (Def.~\ref{def:the-challenge-set-assumption}) holds.

\paragraph{dcoref} As a representative unsupervised system, the ``Stanford Deterministic Coreference Resolution System''~\citep[dcoref;][]{lee-etal-2013-deterministic} is considered. This is a deterministic, rule-based approach to general identity coreference resolution that does not rely on supervised training. The system, which is optimized for the OntoNotes dataset, uses 10 sieves (such as string match and grammatical feature agreement) to identify potentially coreferring mentions. The most recent version implemented in Stanford Core NLP \citep{manning-etal-2014-stanford} is used. The original dcoref paper notes that approximately 30 percent of its errors on OntoNotes are related to pronominal anaphora.

\paragraph{Maverick} For a representative supervised system, the state-of-the-art Maverick coreference resolution system is used~\citep{martinelli-etal-2024-maverick}. The publicly released weights of the best-performing model are used, which consists of a DeBERTa-v3 encoder~\citep{he2021debertav3} finetuned on OntoNotes.

\subsection{Datasets}
\label{sec:csa-datasets}

Systems are evaluated on 11 datasets, including both curated datasets of pronouns attested in natural corpora, such as OntoNotes \citep{ontonotes5}, and challenge sets of WSC-like instances, such as the original WSC test set~\citep{levesque_winograd_2012} and DPR~\citep{rahman-ng-2012-resolving}.

\subsubsection{Attested Pronominal Expressions}

As noted in the methods section (\S\ref{sec:csa-method}), tests involving attested pronouns are structured by restricting annotations from more general coreference resolution datasets to create binary classification problems similar to the WSC. The datasets used for this purpose are described below. Four are datasets of nominal identity coreference in English-language, document-level passages (OntoNotes, OntoGUM, PreCo, and ARRAU). The remaining two focus exclusively on PCR (GAP and PDP).

\paragraph{OntoNotes} OntoNotes 5.0~\citep{ontonotes5} consists of seven genres, including news, conversations, and web data, annotated for coreference by two experts. This dataset has been used in prior work to explicitly evaluate PCR, both in isolation~\citep{zhang-etal-2019-knowledge,zhang-etal-2021-brief,zhang-etal-2019-incorporating} and as a failure case of more general coreference resolution systems~\citep{lu-ng-2020-conundrums}. The standard English CoNLL-2012 Shared Task version of this dataset is used~\citep{pradhan-etal-2012-conll}.

\paragraph{OntoGUM} OntoGUM~\citep{zhu-etal-2021-ontogum} is a reformatted version of the GUM corpus~\citep{10.1007/s10579-016-9343-x}, which was annotated for coreference by linguistic students. Version 9.2.0 of OntoGUM is used in this work. This dataset is designed to follow the same annotation guidelines as OntoNotes while expanding coverage to additional textual genres such as web forums and video blogs.

\paragraph{PreCo} PreCo \citep{chen-etal-2018-preco} is a large-scale dataset of English exams annotated for coreference.

\paragraph{ARRAU} ARRAU 2.1~\citep{Uryupina_Artstein_Bristot_Cavicchio_Delogu_Rodriguez_Poesio_2020} is a dataset of written news and spoken conversations annotated by experts for various anaphoric phenomena. The version formatted by \citet{xia-van-durme-2021-moving} is used here. Its annotation guidelines differ from OntoNotes, and additional phenomena have been annotated, including extensive semantic and syntactic features of mentions.

\paragraph{GAP} GAP~\citep{webster-etal-2018-mind} is a dataset of pronouns attested in English Wikipedia and annotated for coreference. Only instances where exactly one of two candidate antecedents is coreferring with the given pronoun are studied, in order to match the PCR problem formulation of this chapter.

\paragraph{PDP} PDP~\citep{morgenstern2016planning} is a collection of 80 pronoun disambiguation problems attested in text, originally used for the WSC to test systems on examples believed to be relatively easy.

\subsubsection{WSC-like Challenge Sets}

The five challenge sets evaluated in this chapter are as follows. To standardize their format, the lexical units $w_i$ are considered to be syntactic words. The raw text is split into these syntactic words using the Stanza library~\citep{qi-etal-2020-stanza}.

\paragraph{KnowRef-60K} \citet{emami-etal-2020-analysis} presented WSC-like instances created by perturbing internet forum text with heuristic rules. These instances can therefore be considered a hybrid of attested and constructed text.

\paragraph{DPR} The Definite Pronoun Resolution~\citep[DPR;][]{rahman-ng-2012-resolving} dataset contains instances in a format similar to the original WSC, but without the strict requirement that instances cannot be resolved using simple selectional preferences.

\paragraph{SuperGLUE WSC (SG-WSC)} This is the set of WSC instances used for the SuperGLUE benchmark~\citep{wang2019superglue}, which was originally modified from WSC 273 and PDP.

\paragraph{WSC 273} The original WSC~\citep{levesque_winograd_2012} consists of 273 instances. Mentions were manually annotated to fit the format used here, similar to the process in \citet{McCann2018decaNLP} and \citet{toshniwal-etal-2021-generalization}.

\paragraph{Pronominal Winogrande (P-WG)} The portion of the Winogrande test set~\citep{sakaguchi2021winogrande} containing person entities is used. The underscore is replaced with an appropriate third-person pronoun, following the methodology of \citet{porada2023investigating}.

\subsection{Input Format}
\label{sec:csa-input-format}

This section provides further details regarding how the input was formatted. It also discusses additional results under other formats. The main conclusions are found to be consistent across these formatting decisions.

\paragraph{Speaker Information.} Many datasets of coreference annotations consist of spoken language and include corresponding speaker metadata. Models were tested both with and without this metadata, and results are reported for each model in its best configuration. It was found that Maverick and dcoref perform best with speaker information, while including speaker data in the form of ``SPEAKER\_NAME: ...'' had a marginal negative effect on the performance of LMs.

\paragraph{Input Length.} The datasets of curated natural corpora considered here typically consist of long document contexts. Therefore, experiments were conducted including either only the local context $w$ or the entire document in the input. Including the full context length was found to have only a marginal effect on performance. In the unsupervised case, only the local context $w$ is used. For supervised models (both finetuned systems and few-shot LMs), the full document is included in the input.

\paragraph{dcoref} For the dcoref baseline, gold parses are not used, as not all datasets include this information. Instead, parses are predicted by the Stanford CoreNLP pipeline.

\paragraph{LinkAppend} As an additional representative example of supervised systems, the state-of-the-art LinkAppend coreference resolution system \citep{bohnet-etal-2023-coreference} is also considered. The publicly released weights of the best-performing system are used, which consists of the multilingual mT5-XXL language model (13B params) finetuned on OntoNotes.

\section{Findings: Results}
\label{sec:csa-results}

This section first presents a comparison between zero-shot prompting methods and the unsupervised dcoref system. Following this, the best-performing prompting method is compared against the supervised Maverick coreference resolution system and a supervised Llama 3.1 baseline. Across all figures, error bars represent 90\% confidence intervals. Results are presented on the corresponding test splits using the best model configuration for each system.

\begin{figure*}[t]
\centering
  \includegraphics[width=0.99\textwidth]{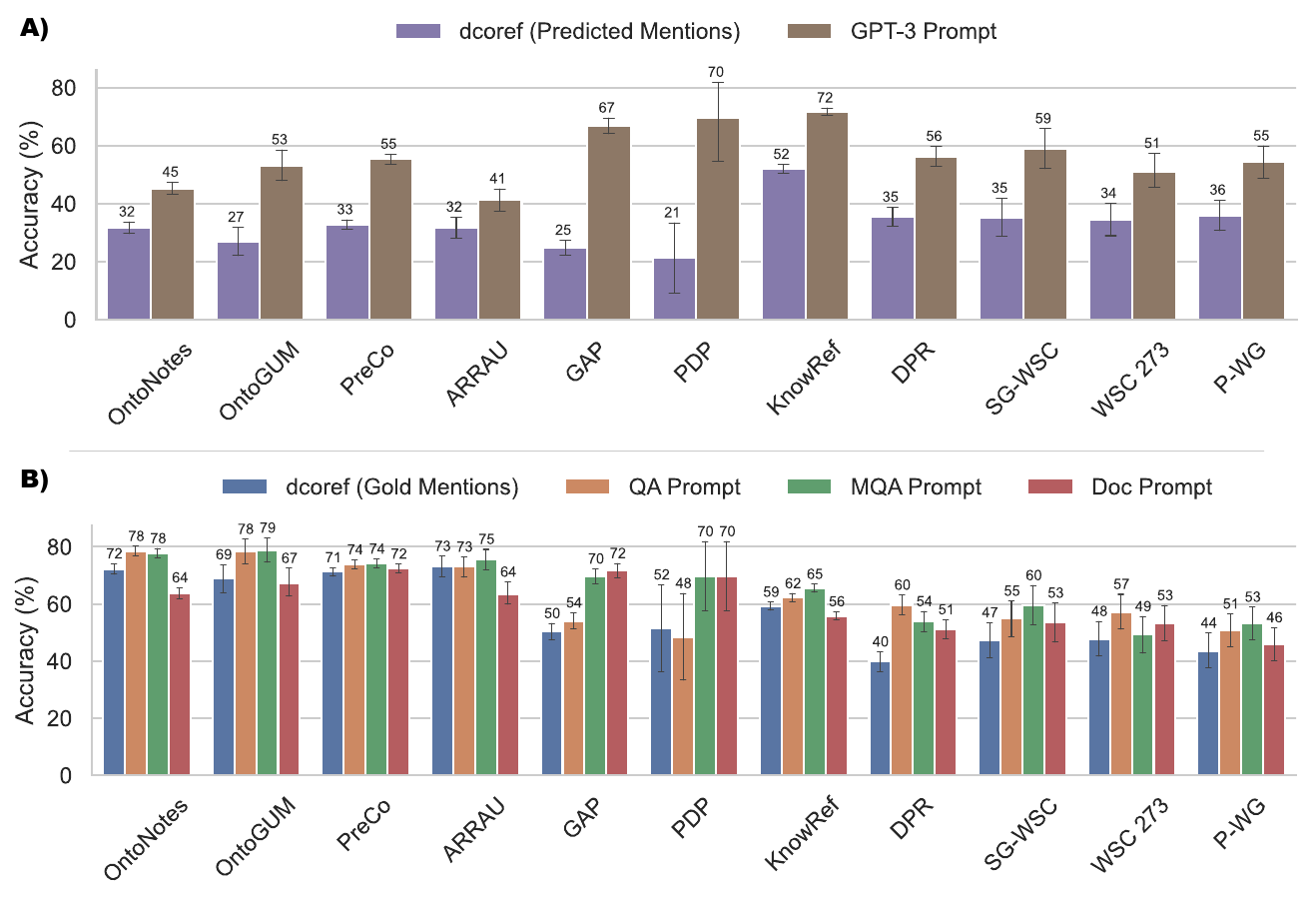}
  \caption[A comparison of the rule-based dcoref system and the Llama 3.1 8B base model prompted for PCR using various prompts.]{A comparison of the rule-based dcoref system~\citep{lee-etal-2013-deterministic} and the Llama 3.1 8B base model prompted for PCR using various prompts. \textbf{A)} Systems that do not need gold mention spans. Across datasets, Llama 3.1 with the GPT-3 prompt always outperforms the dcoref baseline. \textbf{B)} Systems that require gold mention spans as input. In general, prompted Llama 3.1 is more accurate than dcoref on both attested and constructed instances.
  }
  \label{fig:csa-unsupervised-result}
\end{figure*}

\paragraph{Comparing prompted LMs against earlier unsupervised systems, the challenge set assumption does hold.} Results for the fully unsupervised systems are presented in Figure~\ref{fig:csa-unsupervised-result}. Generally, prompted LMs, which outperform dcoref on the WSC variants, also outperform dcoref on datasets of attested pronominal expressions. The results also show that model performance is sensitive to the prompt format.

Exceptions occur on the PDP dataset, whose small size makes it difficult to draw generalizable conclusions, and in the case of the Doc Prompt, which exhibits high variance across datasets. \citet{le2023large} similarly found that Llama models did not consistently generalize with the Doc Prompt template.

\begin{figure}[ht]
\centering
  \includegraphics[width=0.45\textwidth]{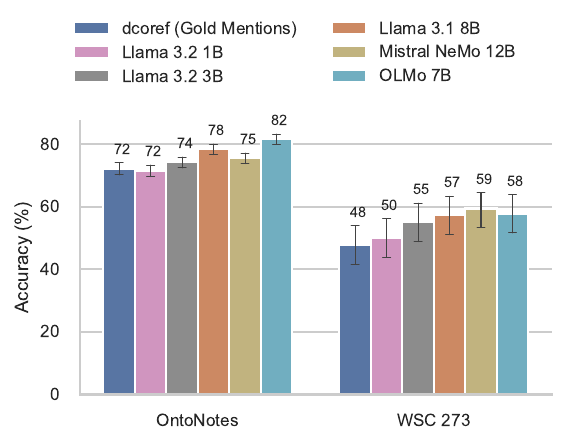}
  \caption[Accuracies of various LMs at pronominal anaphora using the QA prompt template as compared against a dcoref baseline.]{
  Accuracies of various LMs using the QA prompt template as compared against a dcoref baseline. We find that LMs generally outperform dcoref on both attested and constructed instances.
  }
  \label{fig:csa-all-llms}
\end{figure}

In Figure~\ref{fig:csa-all-llms}, the accuracies of various LMs are compared using the QA prompt template. The conclusion that prompted LMs outperform dcoref on both constructed and attested instances is consistent across LM families.

\begin{figure*}[t] 
\centering
  \includegraphics[width=0.99\textwidth]{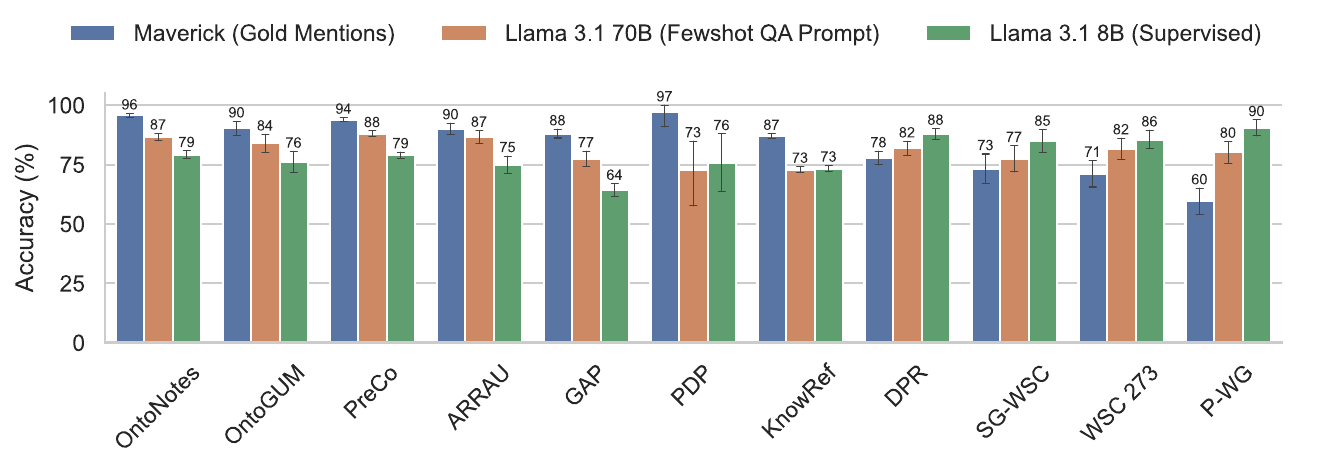}
  \caption[A comparison of the accuracy of Llama 3.1 70B instruct (32-shot) against the supervised Maverick coreference resolution system.]{
  A comparison of the accuracy of Llama 3.1 70B instruct (32-shot) against the supervised Maverick coreference resolution system. We observe that the challenge set assumption does not hold; that is, despite being generally more accurate on WSC-like datasets, the prompted LM is less accurate on datasets of pronominal expressions attested in natural corpora. From left to right, the first six datasets consist of attested examples, and the remaining five are WSC-like challenge sets.
  }
  \label{fig:csa-main-result}
\end{figure*}

\paragraph{However, when comparing prompted LMs against a supervised coreference resolution system, the challenge set assumption does not hold.}

In Figure~\ref{fig:csa-main-result}, the accuracy of Llama 3.1 70B using the QA prompt in a few-shot setting is presented and compared against a supervised coreference resolution system. While the prompted LM is more accurate across WSC-like datasets (with the exception of KnowRef), the supervised coreference resolution system is more accurate at resolving attested pronominal coreferences.

For this experiment, the instruction-tuned version of Llama 3.1 with 32 in-context examples is considered a prototypical example of an LM as evaluated on WSC-like datasets. When this is compared against a supervised Llama 3.1 base model, trained on the Winogrande and DPR training sets for 5k steps, the difference in accuracies across datasets is even more extreme.

The exceptional case of KnowRef may be due to the fact that this dataset is constructed by perturbing attested pronominal expressions and may be overall more similar to collections of attested rather than constructed linguistic expressions.

\section{Analysis}
\label{sec:csa-analysis}

The results presented thus far do not answer the question of \textit{why} the challenge set assumption does not hold. Heuristic estimates of features such as number and animacy are typically required to agree between an antecedent and a pronoun for the two to be coreferring, but these features are, by design, never required to resolve WSC instances. Therefore, one hypothesis is that prompted LMs are not sufficiently considering these features for the attested PCR problems. To test this hypothesis, an analysis was conducted to determine to what extent LMs could benefit from, or already implicitly incorporate, the use of such features. This was done by experimenting with oracle baselines that include these features in the model input as a verbalized statement.

\subsection{Verbalized Features}
\label{sec:csa-verbalized-features}

The verbalized features considered in this analysis are those annotated in the ARRAU corpus. These are: 1) grammatical gender, 2) number, 3) enamex type (i.e., semantic type: is the entity a person, organization, or location?), and 4) distance between mentions. The incorporation of a gold, annotated label was also explored as an oracle baseline.

\paragraph{Prompt.}

Verbalized features are appended to the input string in the form:

\begin{quote}
    \small
    The [FEATURE\_NAME] of ``[X]'' is [Y].
\end{quote}

For example, the passage in Figure~\ref{fig:csa-formulation-example} is prepended with verbalizations such as:

\begin{quote}
    \small
    The grammatical gender of ``Jim'' is male.
\end{quote}

Results of this experiment are presented in Table~\ref{tab:added-features-results}. A minor accuracy increase is observed from the inclusion of grammatical gender, but otherwise there is no discernible influence. Meanwhile, the oracle baseline suggests that models are capable of incorporating verbalized features when they perfectly align with the correct antecedent prediction (i.e., when using gold labels).

\begin{table}[th]
\small
\centering
\begin{tabular}{lr}
\toprule
                                           & ARRAU \\
                                           \midrule
Llama 3.1 70B (QA Prompt) & 0.86                       \\
+ gold gender                                                & 0.87                       \\
+ gold number                                                & 0.85                       \\
+ gold enamex type                                           & 0.86                       \\
+ distance between mentions                                  & 0.84                       \\
+ gold label (oracle)                                        & 0.99                       \\
\bottomrule
\end{tabular}
\caption{Results including additional features in the model input on the ARRAU validation set.}
\label{tab:added-features-results}
\end{table}

\section{Ensembling Systems for Higher Evaluation Scores}
\label{sec:csa-ensembling}

Finally, this section presents results for the proposed ensembling method. This method is motivated by the preceding findings: because the challenge set assumption does not hold, prompted LMs and task-specific systems have distinct strengths at PCR.

\subsection{Method}

This method functions by heuristically determining if $x$ corresponds to a salient discourse entity, in which case a supervised coreference resolution system is used to predict the correct antecedent $a$. Otherwise, $x$ is resolved using a prompted LM. This approach benefits from the fact that supervised coreference resolution systems are relatively accurate at resolving pronominal expressions that corefer to the most salient discourse entities. Meanwhile, prompted LMs are relatively accurate at resolving pronominal expressions referring to infrequently mentioned entities.

This relative strength of each type of system is evidenced by the ensembling method itself. The empirically higher accuracy of this hybrid approach over individual baselines demonstrates that the error distributions of supervised models and prompted LMs are on at least some subset of examples effectively orthogonal; the ensemble achieves a performance increase precisely because the prompted LM is more accurate on low-occurrence entities than the supervised system, and vice versa for high-occurrence entities.

\subsection{Implementation}

For the proposed ensembling method, pronominal coreferences are first predicted using both the supervised Maverick system and the prompted Llama 3.1 70B instruct model, as before. It is then heuristically determined if a candidate antecedent corresponds to a salient discourse entity based on the number of coreferring noun phrases predicted by Maverick in an end-to-end setup. When the number of predicted coreferring mentions is greater than two (that is, the pronoun is estimated to corefer to more than one other linguistic expression), the Maverick predictions given gold mention spans are used. Otherwise, the Llama predictions are used.

\subsection{Results}

\begin{table}[t]
\small
\centering
\begin{tabular}{lccc}
\toprule
System & OntoGUM & PreCo & WSC 273  \\
\midrule
Maverick & \textbf{0.90}	& 0.94 & 0.71 \\
Llama 3.1 70B & 0.84 & 0.88 & 0.82 \\
\midrule
Ensemble & \textbf{0.90}	& \textbf{0.95} & \textbf{0.84}\\
\bottomrule
\end{tabular}

\caption{
  Accuracy of the ensemble method compared against Maverick (supervised coreference resolution) and prompted Llama 3.1 70B instruct.
  }
  \label{tab:ensemble-results}
\end{table}

The results for the proposed ensembling method on three out-of-domain datasets of attested pronominal expressions are presented in Table~\ref{tab:ensemble-results}. The ensemble predictions are at least as accurate as the best-performing model, and in the case of PreCo and WSC 273, more accurate than the single most accurate system.

\section{Example Instances}

\subsection{Examples}

Here we present example instances from the validation sets of those datasets that include a validation split. For readability, we show only the local context $w$. Qualitatively we can observe that examples from the attested corpora are longer even given the length constraint of a maximum three sentences. In the PreCo example instance, it can be observed how grammatical gender might influence the resolution of the pronoun \textit{her}.

\begin{figure}[H]
\centering
  \includegraphics[width=0.75\textwidth]{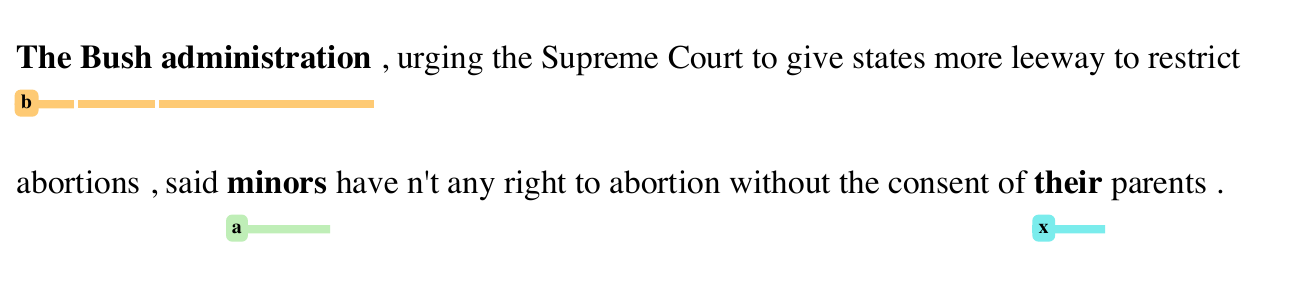}
  \caption{
  An instance from the OntoNotes dataset.
  }
  \label{fig:ontonotes-example}
\end{figure}

\begin{figure}[H]
\centering
  \includegraphics[width=0.475\textwidth]{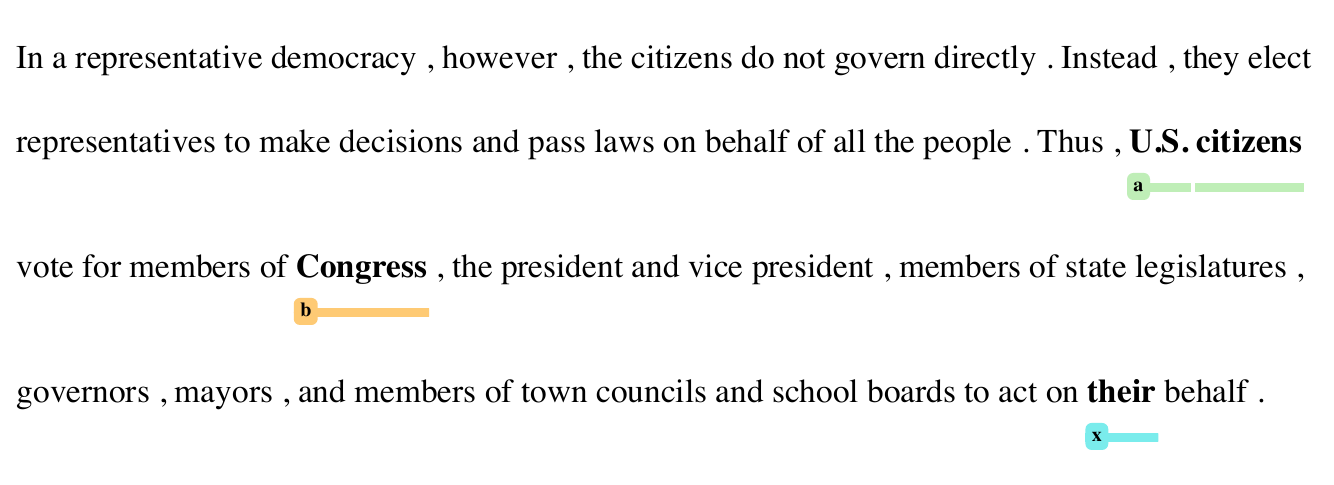}
  \caption{
  An instance from the OntoGUM dataset.
  }
  \label{fig:ontogum-example}
\end{figure}

\begin{figure}[H]
\centering
  \includegraphics[width=0.75\textwidth]{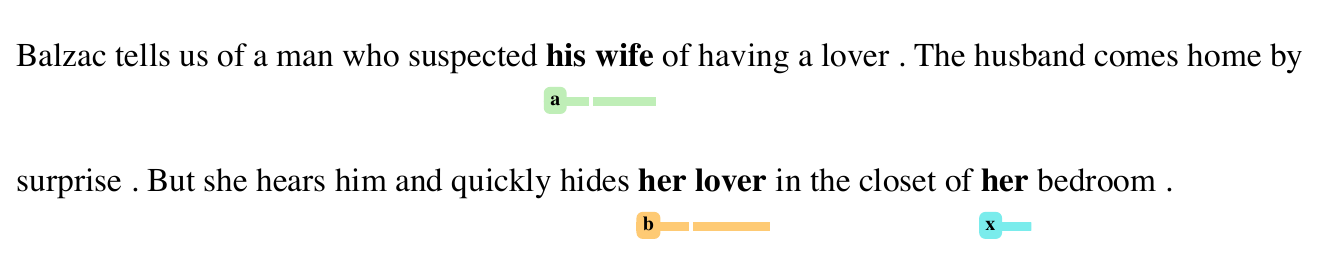}
  \caption{
  An instance from the PreCo dataset.
  }
  \label{fig:preco-example}
\end{figure}

\begin{figure}[H]
\centering
  \includegraphics[width=0.75\textwidth]{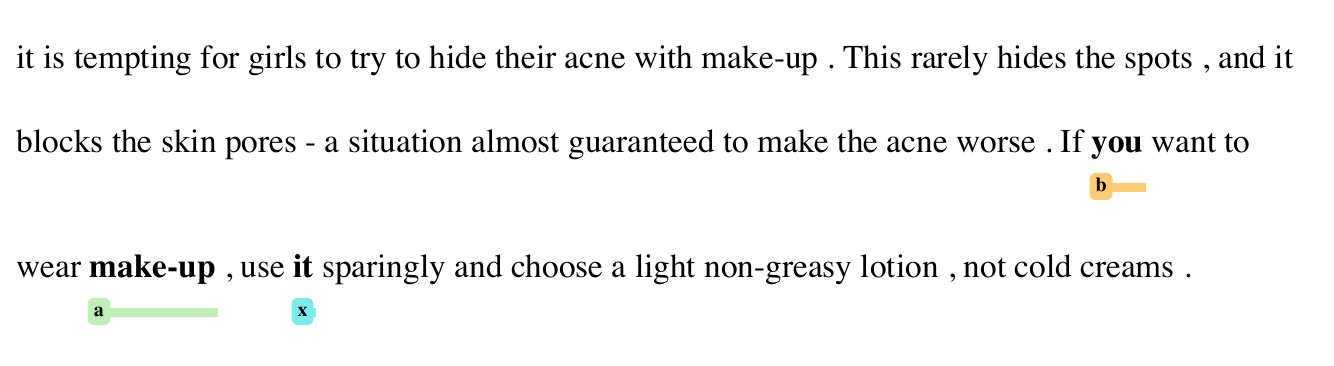}
  \caption{
  An instance from the ARRAU dataset.
  }
  \label{fig:arrau-example}
\end{figure}

\begin{figure}[H]
\centering
  \includegraphics[width=0.75\textwidth]{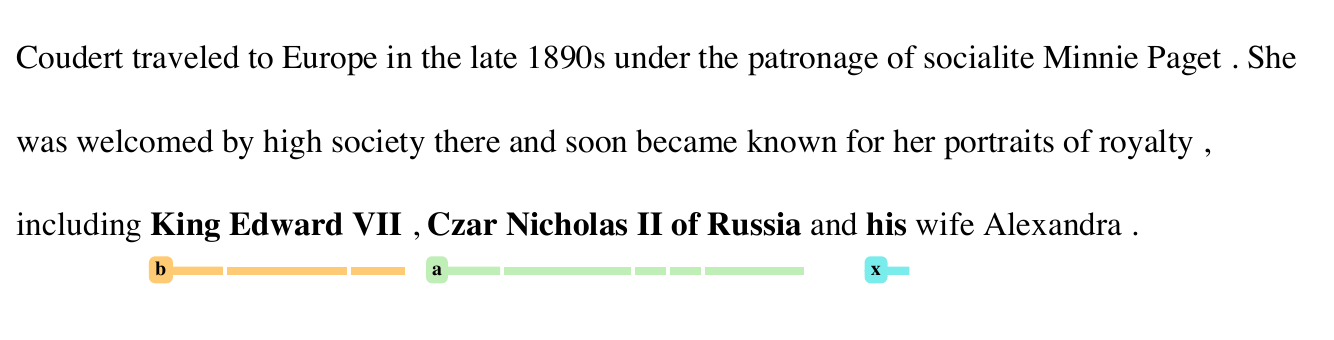}
  \caption{
  An instance from the GAP dataset.
  }
  \label{fig:gap-example}
\end{figure}

\begin{figure}[H]
\centering
  \includegraphics[width=0.75\textwidth]{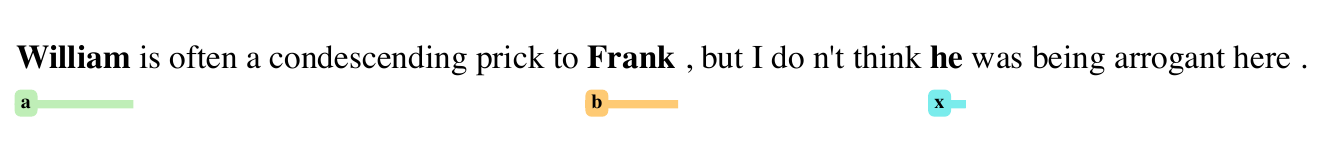}
  \caption{
  An instance from the KnowRef-60K dataset.
  }
  \label{fig:knowref-example}
\end{figure}

\begin{figure}[H]
\centering
  \includegraphics[width=0.75\textwidth]{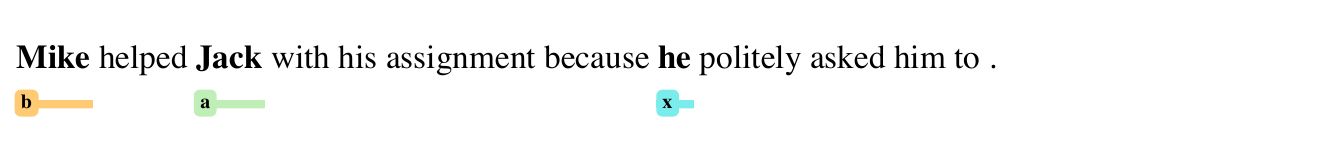}
  \caption{
  An instance from the DPR dataset.
  }
  \label{fig:dpr-example}
\end{figure}

\begin{figure}[H]
\centering
  \includegraphics[width=0.75\textwidth]{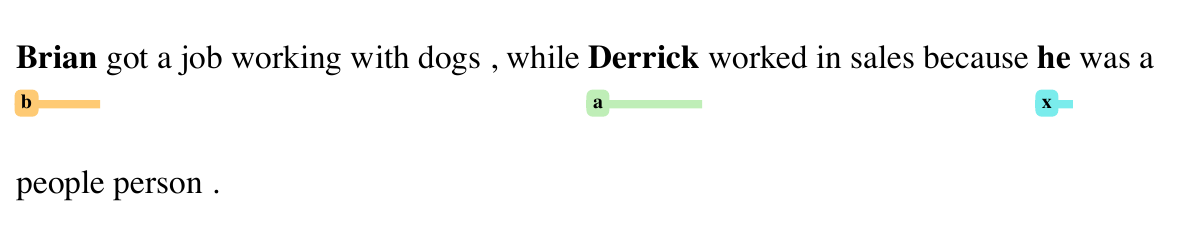}
  \caption{
  An instance from the Pronominal Winogrande (P-WG) dataset.
  }
  \label{fig:winogrande-example}
\end{figure}

\section{Findings: Discussion and Conclusion}
\label{sec:csa-discussion}

Coreference resolution systems have traditionally struggled at resolving pronouns when the resolution depends on semantic knowledge related to high-order predicate-argument relations~\citep{kehler-etal-2004-non,durrett-klein-2013-easy,zhang-etal-2019-sp}. Meanwhile, the results presented in this chapter suggest that resolving WSC instances, which are designed to explicitly rely on such knowledge, is in some ways relatively easier than other cases for prompted LMs. Therefore, the intuitions of the research community regarding what constitutes a challenging example may not always align with the actual failure cases of newer modeling paradigms. Consequently, care must be taken not to interpret high performance on challenge sets as an indication that the more general task being studied can consistently be solved by a given system.

While one may possibly argue that the proposed ``Challenge Set Assumption'' is an oversimplification of assumptions held by the research community, it captures a prevailing implicit heuristic: the expectation that high accuracy on supposedly difficult edge cases implies high accuracy on what is believed to be a more general distribution~\citep{peng-etal-2015-solving,Kocijan_2023}. Consequently, the path forward is likely not to design a single dataset that eliminates these contradictions, but rather to adopt holistic evaluation suites. By treating challenge sets and canonical corpora as complementary rather than hierarchical, we can better detect these specific cases where certain conclusions are not consistent across datasets.

\paragraph{The Solvability of PCR.}

The experiments and results in this chapter are not intended to make claims regarding the solvability of the PCR task. It may be that alternative prompting formats exist for which Llama models are relatively more accurate at resolving attested pronominal coreferences, and one would expect accuracy to increase with LM size. What these results do show, however, is that existing approaches successful on the WSC and its variants do not generalize to all attested PCR problems.

\paragraph{Coreference and Substitutability.}

By their design, WSC instances can be formatted as a cloze-style task where the correct antecedent is that which is most likely to be substituted for the ambiguous pronoun. Substitutability and coreference are related but distinct concepts, however. While WSC instances are difficult in that they cannot be solved with the agreement of features between a pronoun and a candidate antecedent, they differ from some attested PCR problems in that for WSC instances the concept of coreference is aligned with substitutability. One possible hypothesis to explain these results is that this alignment is useful for solving the WSC. This hypothesis is based on the idea that substitutability can be formatted as a cloze-style task and is therefore closely aligned with the LM pretraining objective.

\paragraph{Data Contamination.}

An open question is whether LMs are exposed to the WSC or other datasets' test instances during pretraining. \citet{elazar2023s} estimate that up to 30\% of WSC test instances may be contaminated in the training corpus of Llama and other language models. However, OntoNotes, OntoGUM, Winogrande, Knowref, and GAP are estimated to have close to zero contamination according to the Data Contamination Database~\citep{data-contamination-database}. ARRAU is not publicly distributed and also unlikely to be contaminated. Because the results presented here are consistent for datasets that are likely not contaminated, issues of data contamination are unlikely to invalidate the findings.

\paragraph{Takeaway.}

The ability to disambiguate pronominal expressions is necessary for interpreting natural language and has been used extensively as a benchmark to evaluate models of semantics and discourse.

In this work, several possible approaches to modeling pronominal coreference were studied. Across evaluations, it was found that prompting a large language model (LM) outperforms other approaches on the WSC, but underperforms on certain attested occurrences of pronouns annotated for coreference in OntoNotes and related datasets. This highlights an issue of convergent validity in existing evaluations. That is, findings are not necessarily consistent in the ways one might intuitively expect.

\chapter{Measurement Validity in Coreference Resolution}
\label{chap:measurement}

\myepigraph{It would be interesting to inquire how many times essential advances in science have first been made possible by the fact that the boundaries of special disciplines are not respected. Trespassing is one of the most successful techniques in science.\vspace{5pt}}{\citet{kohler1960dynamics}}

The experimental exploration in Chapter~\ref{chap:challenge-set-assumption} exposed initial limitations in evaluation design by bridging ``canonical'' with ``challenge set'' evaluations. This chapter dives more deeply into canonical evaluations, employing the vocabulary and concepts of measurement validity to more explicitly analyze why these limitations exist.

It is increasingly common to evaluate the same CR model across multiple datasets. A central question, however, is whether such multi-dataset evaluations allow us to draw meaningful conclusions about model generalization, or if they instead reflect the idiosyncrasies of a particular experimental setup. To investigate this, I approach CR evaluation through the lens of measurement modeling, a framework commonly used in the social sciences to assess the validity of measurements.

Adopting this perspective highlights two principal issues with canonical CR evaluations. Firstly, their \textit{content validity} is challenged by the contested nature of coreference itself; with multiple competing definitions in use, it is not always clear which specific construct is being measured. Secondly, these evaluations can lack \textit{discriminant validity}, as performance may be correlated with phenomena beyond the intended scope of measurement.

This analysis shows that across seven datasets, measurements intended to reflect CR model generalization are often confounded by differences in how coreference is both defined and operationalized. This issue limits our ability to draw reliable conclusions about how well CR models generalize along any single dimension.

I argue that the measurement modeling framework provides the necessary vocabulary for articulating the challenges surrounding what is actually being measured by CR evaluations. Thus, drawing this explicit connection to measurement validity is a primary contribution of this thesis.

\begin{table}[t]
    \vspace{1em}
    \centering
    \def\arraystretch{0.8}
    \small
    \begin{tabular}{@{}l  p{0.76\linewidth}@{}}
        Dataset & Example \\
        \midrule
        OntoNotes & Highway officials insist the ornamental railings on \hly{older bridges} aren't strong enough to prevent vehicles from crashing through. But other people don't want to lose \hly{the bridges'} beautiful, sometimes historic, features. \\[.75em]
        
        PreCo & ``Melting permafrost can also destroy \hlb{trees} and forests,'' Weller said. ``When holes in the ground form, \hlb{trees} fall into them and die.'' \\[-.25em]
        
        \makecell[l]{\\ Phrase \\ Detectives} & Within 1942-1944 \hlg{bicycles} were also added to regimental equipment pools. \ldots \ Although seeing heavy use in World War I, \hlg{bicycles} were largely superseded by motorized transport in more modern armies. \\
        \bottomrule
    \end{tabular}
    \caption[Examples of annotated coreference (highlighted) where at least one expression could be interpreted as generic.]{Examples of annotated coreference (highlighted) where at least one expression could be interpreted as generic. Generic expressions tend to be annotated differently across datasets in part due to varying definitions of coreference.}
    \label{tab:coref-examples}
\end{table}

\section{Overview of Methodology}
\label{sec:measurement-introduction}

Fundamentally, coreference is the background concept of multiple linguistic expressions referring to the same discourse entity (see Table~\ref{tab:coref-examples}). However, operationalizations of coreference and of coreference resolution (CR)---the task of identifying coreferring expressions---differ substantially across the literature~\citep[][\textit{i.a.}]{winograd_1972,kantor1977management,hirst1981anaphora}.

A common goal in CR research has been to build models that generalize across diverse settings~\citep{urbizu-etal-2019-deep,xia-van-durme-2021-moving,zabokrtsky-etal-2022-findings}. This goal has motivated the design of benchmarks ``to measure progress towards truly general-purpose coreference resolution''~\citep{toshniwal-etal-2021-generalization}, and has led to the increasingly common practice of evaluating CR models across multiple datasets~\citep[\textit{e.g.,}][]{yang-etal-2012-domain,poot-van-cranenburgh-2020-benchmark,straka-2023-ufal}. 

However, \textit{what we can learn from multi-dataset evaluations about a model's ability to generalize is necessarily limited by differences in how the datasets were constructed}.

Consider a hypothetical scenario: researcher \textbf{A} defines coreference as a relationship exclusively between noun phrases and uses crowdworkers to annotate a dataset accordingly. Independently, researcher \textbf{B} defines coreference as a relationship between any noun phrase or verb phrase, annotating a different dataset with this broader definition. If a CR model is trained on dataset \textbf{A} and evaluated on dataset \textbf{B}, any conclusions about its generalization ability will be confounded by the foundational differences in how coreference was defined and annotated across the two settings.

This chapter aims to systematically examine how such differences in the definition and annotation of coreference across datasets limit the conclusions that can be drawn about the generalization ability of CR models. To do so, I analyze common evaluation practices using the \textit{measurement modeling} framework~\citep[][\textit{i.a.}]{smelser_indicator_2001,jacobs2021measurement,blodgett-etal-2021-stereotyping}. This framework distinguishes what is being measured, the \textit{construct}, from how it is being measured, the \textit{operationalization} of that construct via a \textit{measurement model}. This distinction is crucial when working with contested constructs---those with disputed or competing definitions---as it helps differentiate between disagreements over the definition of a construct versus variations in the operationalization of a single, agreed-upon definition.

Coreference is, in fact, a contested construct~\citep{grishman-sundheim-1996-design,doddington-etal-2004-automatic,can_we_fix}. As \citet{poesio-etal-1999-mate} note, ``[a] very basic problem arising in the case of coreference is deciding what type of information is being annotated, since the term `coreference' is used to indicate different things.'' The measurement modeling framework thus helps illustrate how multi-dataset evaluations risk conflating various factors concerning what is being measured, thereby obscuring what we learn about a model's ability to generalize.

In this chapter, I conduct a multi-dataset evaluation across seven English-language coreference datasets and analyze the results through this lens. Specifically, I consider how coreference is defined and annotated within each dataset~(\S\ref{sec:measurement-datasets}) and how this affects measurements of model generalization~(\S\ref{sec:measurement-experiments} \& \S\ref{sec:measurement-results}). To this end, I identify particular types of coreference that vary between definitions---such as the treatment of generic and predicative expressions---and show, using disaggregated results, how models systematically fail when the definition of coreference in the test set differs from that of the training set.

For instance, consider the coreference between the two generic instances of ``trees'' in PreCo, highlighted in Table~\ref{tab:coref-examples}. A state-of-the-art CR model trained on OntoNotes fails to predict this link. This failure is likely attributable to a discrepancy in operationalization: under the definition of coreference used in OntoNotes, two generic nouns of this type are typically not considered coreferring.

The results suggest that in order to understand if and how CR models generalize a concept of coreference, we must \textit{first} account for inconsistencies in how that concept is defined. By applying the measurement modeling framework and using disaggregated evaluations, this work provides a blueprint for assessing out-of-domain performance in a way that controls for known variations in how coreference is defined and operationalized across datasets.

\section{Methodology: Background}
\label{sec:measurement-related-work}

\paragraph{CR Model Error Analyses.}
My work is related to prior efforts to identify and classify model errors. Traditional error analyses of CR models, however, have typically focused on performance measurements within a single dataset~\citep{uryupina-2008-error,Versley2008,durrett-klein-2013-easy}. While some analyses consider collections of datasets and find that many error types are consistent, they often do not examine out-of-domain settings where models are tested on datasets they were not trained on~\citep{lu-ng-2020-conundrums,chai-strube-2023-investigating}. Crucially, these prior studies—as well as those analyzing downstream performance~\citep{dasigi-etal-2019-quoref,chai-etal-2022-evaluating}—have not considered how the underlying definition of coreference itself might correlate with the observed errors.

\paragraph{CR Model Generalization.}
To improve generalization, several approaches have focused on training a CR model with multiple training sets, often involving techniques like \textit{active learning}~\citep[\textit{e.g.,}][]{zhao-ng-2014-domain,https://doi.org/10.48550/arxiv.2210.07602,yuan-etal-2022-adapting}. It has been observed, however, that models trained on multiple datasets do not necessarily improve in accuracy on a test set for which large amounts of in-domain training data are already available~\citep{toshniwal-etal-2021-generalization}. Other work has addressed specific aspects of annotation variance; for instance, \citet{moosavi-etal-2019-using} studied how mention span boundaries differ across datasets and proposed an algorithm for normalizing them.

Additionally, prior work has examined specific generalization challenges. Models trained on OntoNotes are known to be less accurate on textual genres or proper nouns not seen during training~\citep{moosavi-strube-2017-lexical,subramanian-roth-2019-improving,zhu-etal-2021-ontogum}, a weakness that has been addressed by incorporating explicit linguistic features or syntactic rules~\citep{zeldes-zhang-2016-annotation,moosavi-strube-2018-using}. Other studies have focused on generalization to difficult cases of pronominal coreference, noting that OntoNotes-trained models perform poorly on examples from web text~\citep{webster-etal-2018-mind,emami-etal-2019-knowref} or constructed challenge sets~\citep{rahman-ng-2012-resolving,peng-etal-2015-solving,toshniwal-etal-2021-generalization}.

\paragraph{Measurement Modeling.}
\label{sec:measurement-measurement_modeling}
The effective measurement of theoretical concepts has been studied extensively in the quantitative social sciences~\citep[\textit{e.g.,}][]{black_doing_1999,adcock_collier_2001,bhattacherjee2012social}. Because such concepts are often unobservable and cannot be measured directly, they must be inferred from observable variables. A model that describes the relationship between a theoretical concept and the variables used to infer it is called a \textit{measurement model}~\citep{smelser_indicator_2001,jacobs2021measurement}. In this paradigm, the term \textit{construct} refers to the theoretical concept being measured, while the term \textit{indicator} refers to an observable variable from which the measurement of the construct is inferred~\citep{kline_principles_2011}.

I draw on this line of work to examine how design decisions made during the construction of CR datasets can limit the conclusions we draw about a model's ability to generalize. Specifically, the measurement modeling framework provides a lens through which to analyze how choices in defining and annotating coreference affect the measurement of the constructs of \textit{coreference}, \textit{CR model performance}, and \textit{CR model generalization}.

\section{Methodology: Coreference}
\label{sec:measurement-datasets}

As a theoretical concept, \textit{coreference} is an unobservable construct. Therefore, any measurement of it must be inferred from indicators, such as the annotations of coreference provided in a dataset.

This section first describes the coreference datasets used in my analysis (\S\ref{sec:measurement-intro-datasets}). It then details the discrepancies in how the construct of coreference is defined (\S\ref{sec:measurement-constructs}) and operationalized (\S\ref{sec:measurement-annotation-guidelines}) across these datasets.

\subsection{Datasets}
\label{sec:measurement-intro-datasets}

The ultimate goal is to understand what multi-dataset evaluations reveal with respect to models' ability to generalize.
To examine this, I focus the analysis on datasets that have been used in prior work to evaluate CR model generalization. Specifically, this selection of datasets is based on prior multi-dataset evaluations of CR models \citep{toshniwal-etal-2021-generalization,xia-van-durme-2021-moving,zhu-etal-2021-ontogum,zabokrtsky-etal-2023-findings}. In total, I consider seven English-language datasets containing annotations of identity coreference at a document level (see Table~\ref{tab:dataset-stats} for dataset figures). \S\ref{sec:measurement-scope} provides more details regarding the selection of these datasets, how they have been preprocessed or formatted, and the use of these datasets within the broader coreference literature.

\para{OntoNotes} OntoNotes 5.0~\citep{ontonotes5} consists of news, conversations, web data, and biblical text in which coreference was annotated by experts. We use the English CoNLL-2012 Shared Task version of this dataset~\citep{pradhan-etal-2012-conll}. While OntoNotes 5.0 is annotated for multiple phenomena, the CoNLL-2012 Shared Task contains only so-called ``identical'' coreference, also sometimes referred to as ``anaphoric coreference.''

\begin{table}[t]
    \centering
    \small
    \def\arraystretch{0.65}
    \setlength{\tabcolsep}{1.1em}
    \begin{tabular}{lcccc}
        Dataset & Train & Dev. & Test & Total Words (K) \\
        \midrule
        OntoNotes & 2,802 & 343 & 348 & 1,632 \\[.2em]
        PreCo & 36,120 & 500 & 500 & 12,493 \\[.2em]
        Phrase Det. & 695 & 45 & 45 & 1,321 \\[.2em]
        \hdashline\\[-.5em]
        OntoGUM & 165 & 24 & 24 & 204 \\[.2em]
        LitBank & 80 & 10 & 10 & 211 \\[.2em]
        ARRAU & 444 & 33 & 75 & 348 \\[.2em]
        MMC & 955 & 134 & 133 & 324 \\
        \bottomrule
    \end{tabular}
    \caption[Number of documents for each dataset split of studied CR datasets.]{Number of documents for each dataset split.}
    \label{tab:dataset-stats}
\end{table}

\para{PreCo} PreCo \citep{chen-etal-2018-preco} consists of English comprehensive exams annotated for coreference by trained university students.

\para{Phrase Detectives} Phrase Detectives 3.0~\citep{yu-etal-2023-aggregating} consists of Wikipedia, fiction, and technical text. The training set annotations were sourced by aggregating annotations of users playing the \textit{Phrase Detectives} online game, where the users were tasked with annotating coreferences, or verifying others' annotations. The test set was annotated by experts. We use the ``CoNLL'' formatted version of the dataset.\footnote{\url{https://github.com/dali-ambiguity/Phrase-Detectives-Corpus-3.0}}

\para{OntoGUM} OntoGUM~\citep{zhu-etal-2021-ontogum} is a reformatted version of the GUM corpus~\citep{10.1007/s10579-016-9343-x}. The GUM corpus was originally annotated in an iterative process by linguistics students. OntoGUM was created by transforming GUM, using deterministic rules and annotated syntactic parses, to follow the OntoNotes annotation guidelines. We use version 9.2.0 of OntoGUM.

\para{LitBank} LitBank~\citep{ bamman-etal-2020-annotated} consists of coreference annotated in English literature by experts. Only noun phrases of ACE categories and pronouns have been annotated for coreference.\footnote{The ACE categories are: people, facilities, geo-political entities, locations, vehicles, and organizations.} We use the ``CoNLL'' formatted dataset version.\footnote{\url{https://github.com/dbamman/litbank}}

\para{ARRAU} ARRAU 2.1~\citep{Uryupina_Artstein_Bristot_Cavicchio_Delogu_Rodriguez_Poesio_2020} is a dataset of written news and spoken conversations annotated for various anaphoric phenomenon by experts.
We use all documents and the formatting procedure of \citet{xia-van-durme-2021-moving} which keeps only the coarsest-grained  annotations.

\para{MMC} Multilingual Coreference Resolution in Multiparty Dialogue (MMC)~\citep{zheng-etal-2023-multilingual} is a dataset of television transcripts. The training set was annotated by crowdworkers, and the test set by experts.
We use the English portion of the ``CoNLL'' formatted version of this dataset.\footnote{\url{https://github.com/boyuanzheng010/mmc}}

\begin{table*}[th]
    \centering
    \tiny
    \def\arraystretch{0.65}
    \setlength{\tabcolsep}{0.01em}
    \begin{tabular}{p{0.13\linewidth}@{\hskip 1em} p{0.3\linewidth}@{\hskip 1em} p{0.25\linewidth}@{\hskip 1em} p{0.25\linewidth}}
        Type & OntoNotes & PreCo & Phrase Detectives \\
        \midrule
        \textbf{Generic Mentions} \newline [dogs] can bark & Generics mentions are only annotated when they corefer with a pronoun or determinate noun phrase, or when they occur in a news headline. & All generic noun phrases and modifiers can be annotated as coreferring. & All generic noun phrases can be annotated as coreferring. \\
        \midrule
        \textbf{Verb Phrases} \newline it will [grow] & The head of a verb phrase can be annotated as coreferring with a determinant noun phrase. & Not annotated. & Not annotated. \\
        \midrule
        \textbf{Appositives} \newline [[Abe] , [the chef]] & Annotated in the dataset, but not considered coreference. & Annotated as three mentions: both noun phrases and the larger span. & Annotated in the dataset, but not considered coreference. \\
        \midrule
        \textbf{Copular Predicates} \newline [he] is [the teacher] & Not annotated. & In a copular structure, the referent and attribute are annotated as coreferring. & Annotated in the dataset, but not considered coreference. \\
        \midrule
        \midrule
        \textbf{Nesting} \newline [he [himself]] & When two nested mentions share a head, only the dominant mention is annotated. Proper nouns cannot contain nested mentions. & Appositives and mentions with shared heads are annotated as nested mentions. & The right-most mention in an appositive is considered referring to a distinct entity and can therefore be annotated as a nested mention. No restrictions on nesting of proper nouns. \\
        \midrule
        \textbf{Compound \newline Modifiers} \newline [Taiwan] authorities & Compound modifiers are annotated if non-adjective proper nouns that are not a nationality acronym. & All compound modifiers can be annotated as coreferring. & No explicit restrictions on the annotation of compound modifiers. \\
        \bottomrule
    \end{tabular}
    \caption{Noted differences in how coreference is defined and operationalized in the training datasets.}
    \label{tab:general-cr-differences}
\end{table*}

\subsection{Dataset Selection and Scope}
\label{sec:measurement-scope}

\subsubsection{Dataset Selection}

The selection of datasets is informed by the multi-dataset evaluations of \citet{toshniwal-etal-2021-generalization}, \citet{xia-van-durme-2021-moving}, \citet{zhu-etal-2021-ontogum}, and \citet{zabokrtsky-etal-2023-findings}.

From the datasets evaluated by \citet{xia-van-durme-2021-moving} (\xia), I select the same large-scale training sets, OntoNotes and PreCo, as well as the two largest English-language test sets, LitBank and ARRAU. The collection in \citet{toshniwal-etal-2021-generalization} is largely a subset of \xia, with the exception of the Friends dataset~\citep{zhou-choi-2018-exist}. For this domain, I use the more recent MMC dataset~\citep{zheng-etal-2023-multilingual}, which encompasses the Friends corpus.

From \citet{zabokrtsky-etal-2023-findings}'s evaluation, I select GUM~\citep{10.1007/s10579-016-9343-x}, as the other English dataset, ParCorFull~\citep{lapshinova-koltunski-etal-2022-parcorfull2}, consists of only 19 documents. Specifically, I use the OntoGUM formatted version of the dataset~\citep{zhu-etal-2021-ontogum}. This choice allows for a controlled comparison where two datasets (OntoNotes and OntoGUM) are intended to share the same theoretical construct but differ in aspects of their operationalization.

Finally, I include the recently released, large-scale Phrase Detectives 3.0 dataset~\citep{yu-etal-2023-aggregating} as a third major training set.

\subsubsection{Dataset Scope}

While \textit{coreference} is most often used to denote \textit{identity coreference}---a relationship between linguistic expressions referring to entities with the same identity~\citep{nedoluzhko2021coreference}---many related phenomena are sometimes also labeled as coreference. These include bridging anaphora~\citep{clark1977bridging}, discourse deixis~\citep{webber1991structure}, event coreference~\citep{lu2018event}, and split-antecedent anaphora~\citep{edes1968output}. For this thesis, I focus on datasets of identity coreference and consider these other phenomena only insofar as they are subsumed under that category by a given dataset's annotation guidelines.

Furthermore, following \citet{xia-van-durme-2021-moving}, I exclude discontinuous mention spans from my evaluations~\citep{yu-etal-2023-universal}. Such spans are incompatible with most existing CR models, which assume that mentions correspond to a continuous sequence of tokens.

The treatment of \textit{singletons} also varies. Singletons are sometimes defined as a cover term for non-referring expressions (e.g., in idioms like \textit{on the contrary}) alongside mentions that happen to appear only once~\citep{kubler-zhekova-2011-singletons}. A more common definition, however, is simply any linguistic expression referring to a discourse entity that is mentioned only once~\citep{can_we_fix}. While the handling of singletons has been considered in some multi-dataset evaluations~\citep{toshniwal-etal-2021-generalization}, I follow the common practice of focusing my evaluation on non-singleton mentions~\citep{zabokrtsky-etal-2022-findings}.

\subsubsection{Dataset Formatting}

For PreCo, I use the last 500 documents of the training set as a validation split. For Phrase Detectives, I randomly select 45 documents from its training set as a validation split. For Litbank, I use ``split\_0'' from the official repository. For OntoGUM, I use the official splits and include the Reddit data. For ARRAU, I use the existing splits for the WSJ portion and randomly partition the remaining data into 80\%/10\%/10\% for train/validation/test splits, respectively.

\subsection{Differences Across Datasets}
This section provides an overview of the differences in how coreference is defined and operationalized across the three largest datasets used as training sets: OntoNotes, PreCo, and Phrase Detectives.

\subsubsection{Differing Definitions}
\label{sec:measurement-constructs}

Because datasets differ in how coreference is conceptualized, measurements of how CR models generalize from one to another risk being confounded by these underlying differences in the construct itself.

Notable differences in how these three datasets define coreference stem primarily from whether the following phenomena are considered to be coreference relations:
\textbf{1)} multiple \textbf{generic} expressions that can be interpreted as referring to the same discourse entity;
\textbf{2)} expressions referring to the same event where at least one is a \textbf{verb phrase} (VP);
\textbf{3)} two expressions in \textbf{apposition}; and
\textbf{4)} two expressions in a \textbf{copular structure}.
Table~\ref{tab:general-cr-differences} provides an overview of how each dataset treats these phenomena.

These differences in how coreference is defined were determined based on the original documentation of the respective datasets. We considered both the original publication as well as the annotation guidelines where available. More details on the definitions of coreference for each dataset are provided below:

\para{OntoNotes}
\textit{Identical coreference} is defined in the OntoNotes annotation guidelines to be ``names, nominal mentions, pronominal mentions, and verbal mentions of the same entity, concept, or event''~\citep{bbn2007guidelines}. This construct is distinguished from appositives (``immediately-adjacent noun phrases, separated
only by a comma, colon, dash, or parenthesis'') and copular structures (a subject and predicate linked by a copula). Furthermore, identical coreference is defined to ``not include entities that are only mentioned as generic, underspecified or abstract.''

\para{PreCo}
The PreCo authors establish a conceptualization of coreference mostly by contrasting what is considered to be coreferring in PreCo with the OntoNotes definition. As the PreCo annotation guidelines are not public, we can only assess the construct intended to be measured based on the corresponding paper and data. The PreCo authors note that they follow most of the conceptualization of coreference used in OntoNotes with the exception that they exclude verb phrases and explicitly include generic mentions and mentions in appositive and copular structures which may refer to the same entity. The PreCo dataset is additionally annotated for singleton mentions, although no definition is given for the concept of singleton.

\para{Phrase Detectives}
In the case of Phrase Detectives, coreference is considered a relation strictly between noun phrases. Predication, including apposition and copular structures, is distinguished from coreference and annotated separately. Generic only mentions can be annotated as coreferring.

\subsubsection{Differing Annotations}
\label{sec:measurement-annotation-guidelines}

Beyond differences in how coreference is conceptualized, I also consider how the construct is operationalized in practice. While many operational dimensions exist, this analysis focuses on differences in datasets' annotation guidelines, particularly concerning how coreferring expressions are specified. I highlight two key, empirically testable ways in which annotation guidelines differ across the primary training datasets: \textbf{1)} the treatment of \textbf{nested mentions} and \textbf{2)} the treatment of \textbf{compound modifiers}. Table~\ref{tab:general-cr-differences} provides further detail on these and other differences.

Naturally, conceptualization influences operationalization; these two aspects are thus closely coupled. For example, whether appositives are included in the definition of coreference directly affects how nested mentions are subsequently annotated. In this analysis, I therefore often consider factors related to operationalization and conceptualization jointly, as they are not explicitly delineated in the datasets' documentation. More generally, the literature itself rarely draws a clear distinction between decisions related to conceptualizing coreference—that is, defining the theoretical construct—and decisions related to its measurement, or how a given definition is operationalized.

For instance, some annotation decisions are explicitly pragmatic, intended to increase inter-annotator agreement. In OntoNotes, ``[o]nly the single-word head of the verb phrase is included in the span, even in cases where the entire verb phrase is the logical co-referent'' for this reason~\citep{bbn2007guidelines}. However, the status of other guidelines, such as the handling of compound modifiers, is more ambiguous. One could argue such a guideline is either a practical heuristic to improve annotation consistency or a fundamental part of the definition of coreference itself. Clarifying this distinction remains an open challenge for the research community.

\section{Methodology: Details}
\label{sec:measurement-experiments}

This section first overviews the experimental setup (\S\ref{sec:measurement-disaggregated-eval}), then describes the details of the evaluation itself (\S\ref{sec:measurement-eval-metrics}) and the models considered (\S\ref{sec:measurement-models}). My goal is to understand how measurements of generalization are affected by differing definitions and operationalizations of coreference across datasets. To this end, I empirically examine performance variations on types of coreference that differ in how they are conceptualized and/or operationalized across these datasets.

\subsection{Disaggregated Evaluation}
\label{sec:measurement-disaggregated-eval}

The purpose of a disaggregated evaluation is to determine if failures in generalization, as measured by a model's accuracy across multiple test sets, are correlated with differences in how coreference is defined and operationalized in those sets. If measurements of generalization are indeed sensitive to these differences, it would indicate that such measurements are confounded by variations in the definition of coreference, rather than solely capturing a model's capacity to generalize a single, consistent conceptualization of the task.

To achieve this, I evaluate models on the test set of the dataset they were trained on, which is referred to as \textit{in-domain}, as well as on the test sets of all other datasets, referred to as \textit{out-of-domain}. My analysis highlights types of coreference that are significantly more difficult for a model to resolve out-of-domain compared to in-domain. These cases are typically interpreted as evidence of limited generalization, and I seek to examine whether this limitation is correlated with types of coreference that differ in their conceptualization and operationalization across datasets.

More formally, let $\theta$ be the parameterization of a model, $X$ be the in-domain dataset, $\hat{X}$ be some out-of-domain dataset, and $f_\theta(X)$ be a measurement of accuracy on dataset $X$. The \textit{aggregate generalization gap} (AGG) is defined as:
\begin{equation}
    \text{AGG} = \lvert f_\theta(X) - f_\theta(\hat{X}) \rvert 
\end{equation}

Further, let $X_t$ be the subset of dataset $X$ that is of type $t$, for which model accuracy is calculated as $f_\theta(X_t)$. The analysis will highlight cases where the aggregate generalization gap differs significantly from the \textit{type generalization gap} (TGG) for a given type of coreference $t$, defined as:
\begin{equation}
    \text{TGG} = \lvert f_\theta(X_t) - f_\theta(\hat{X_t}) \rvert
\end{equation}

\subsubsection{Formalizing Coreference Types}

This section clarifies the precise differences in coreference definition and operationalization considered in this study. I formalize these differences as particular \textit{types of coreference} that vary in their treatment between datasets. Since all the noted differences in annotation pertain to particular kinds of mentions within a coreference relation (Table~\ref{tab:general-cr-differences}), each \textit{type of coreference} is defined based on formally defined \textit{types of mentions}. For the moment, it is not important whether a given mention is annotated in the dataset or predicted by a model; this distinction will be clarified during the evaluations of precision and recall (\S\ref{sec:measurement-eval-metrics}).

\paragraph{Metadata.}
In practice, some mention types are identified using universal dependency relations \citep{nivre-etal-2020-universal} or part-of-speech (POS) tags. This metadata is sourced as follows: for OntoNotes, I convert the expert-annotated constituency parses into dependency parses using the Stanford CoreNLP library \citep{manning-etal-2014-stanford}. For OntoGUM, I use the existing expert-annotated dependency parses. For all other datasets, including PreCo and Phrase Detectives, I parse the text using the Stanza parser \citep{qi2020stanza}.

\vspace{8pt}

\para{Studied Coreference Types}

\begin{enumerate}[wide, labelindent=0pt]
    \item \textbf{Nested}: {\em A mention that corefers with an overlapping span (either larger or smaller).} \vspace{-2pt}
    \begin{quote}
        \small
        President Chen said, [he [himself]] has not returned to his hometown \ \ldots
    \end{quote} \vspace{-4pt}

    \item \textbf{ON Generic}: {\em A coreferring mention that is a generic noun phrase.} We use the OntoNotes definition of generic, which we refer to as an ON generic: a noun phrase is considered generic in OntoNotes if it has an indefinite article or is plural with no article. We check if there exists \textit{a} or \textit{an} as the child of a \texttt{det} relation, or there is no \texttt{det} relation and the POS tag is \texttt{NNS} (plural noun).
    \vspace{-2pt}
    \begin{quote}
        \small
        \ldots you know yeah they had [a farm] when they were first married \ldots I don't know how many years they had [a dairy farm] \ldots
    \end{quote} \vspace{-2pt}
    
    \item \textbf{Compound}: {\em A coreferring mention that is a compound modifier.} This is defined to be a mention that is the dependent of a \texttt{compound:nn} relation.
    \vspace{-2pt}
        \begin{quote}
        \small
        \ldots \ we miss our [Taiwan] compatriots even more, and \ldots \ gave a speech, expressing hopes that [Taiwan] authorities would \ \ldots
    \end{quote}  \vspace{-2pt}
    
    \item \textbf{Copular}: {\em A mention that is in a copular construction with another mention to which it corefers.} Two mentions are said to be in a copular construction if they are in an \texttt{nsubj} relation with each other and the rightmost mention is the head of a \texttt{cop} dependency relation.
    \vspace{-2pt}
    \begin{quote}
        \small
        Yet I realize that in my father's eyes, [I] will always be [his little girl].
    \end{quote} \vspace{-2pt}
\end{enumerate}

\subsubsection{Evaluation Metrics}
\label{sec:measurement-eval-metrics}

In this section, I outline the specific metrics used to calculate the accuracy measurement $f_\theta(\cdot)$. To evaluate a model, we would like to know how well said model's predicted coreference clusters agree with the datasets' annotated clusters.

\para{Notation} Let document $D = (w_1, w_2, \ldots, w_n)$ be a sequence of $n$ words. Let $S = \{ s_{i:j} \mid 1 \leq i \leq j \leq n \}$ denote the set of all spans of words in $D$, where a span is a continuous subsequence of words $s_{i:j} = (w_i, \ldots, w_j)$.
The set of annotated mentions $M \subseteq S$ denotes all spans in $S$ that are annotated as coreferring with at least one other span. The set of entities $E = \{ E_1, E_2, \ldots, E_k \}$ is a $k$-partition of $M$, i.e., a family of $k$ non-empty, pairwise disjoint subsets of $S$. Each entity $E_i \in E$ represents a set of spans that refer to the same discourse entity in the document. A CR model takes as input $D$ and outputs $E' = \{ E'_1, \ldots, E'_\ell \}$, a prediction of the true $E$, which is an $\ell$-partition of the set of model predicted mentions $M'$.

\para{Metrics} There is not a single standard approach for calculating disaggregated CR evaluation scores. I choose to the adopt the B$^3$-based method of \citet{bamman-etal-2020-annotated} due to its clear intuition, although in practice we find that our conclusions are not sensitive to the exact choice of metric.

In the aggregate case, I report agreement using B$^3$ recall, precision, and F1 \citep{bagga-baldwin-1998-entity-based}.
The probabilistic intuition behind B$^3$ recall is that it calculates, in expectation over all annotated mentions, the probability that an annotated coreferring mention is correctly recalled by the model. A similar intuition applies to precision.

More formally, B$^3$ calculates cluster agreement by averaging over all mentions in the cluster. Given mention $s$, let $\mathcal{E} = \bigcup \{E_i \mid s \in E_i\}$ be the set of all mentions annotated as coreferring with $s$, and $\mathcal{E'} = \bigcup \{E'_i \mid s \in E'_i\}$ the set of all mentions predicted as coreferring with $s$. B$^3$ calculates recall (R) and precision (P) as

{\small
\begin{equation}
    \text{R} = \frac{1}{|M|} \sum_{s \in M} \frac{|\mathcal{E} \cap \mathcal{E'}|}{|\mathcal{E}|} \qquad \text{P} = \frac{1}{|M'|} \sum_{s \in M'} \frac{|\mathcal{E'} \cap \mathcal{E}|}{|\mathcal{E'}|}
\end{equation}
}
from which F1 is calculated as the harmonic mean.

To then evaluate performance on a given type of coreference, I use a modified B$^3$ score that calculates the expected agreement only for mentions of the given type. That is, I only sum over mentions $s$ that are of the type of mention being considered. In other words, $M$ is replaced with $\{m \mid m \in M \land \text{is\_type}(m)\}$ when calculating recall, and $M'$ is replaced with $\{m \mid m \in M' \land \text{is\_type}(m)\}$ when calculating precision. Where here, $\text{is\_type}(m)$ indicates that mention $m$ is of the type being considered. I also report  CoNLL score \citep{denis2009global,pradhan-etal-2012-conll} in the aggregate case to facilitate comparison with prior work.

\subsection{Models}
\label{sec:measurement-models}

My evaluation focuses on four models: the publicly available LinkAppend model~\citep{bohnet-etal-2023-coreference} and three versions of LingMess~\citep{otmazgin-etal-2023-lingmess}, each trained on one of the three primary training datasets. These were selected as they are currently the best-performing decoder and encoder-based models, respectively, on the OntoNotes test set. Hyperparameters are available in the public repository for this thesis.

\paragraph{LinkAppend.} \citet{bohnet-etal-2023-coreference} proposed a method for formulating CR as a sequence-to-sequence task where the objective is to predict links between mentions or append them to existing clusters. They trained a 13B parameter mT5 language model \citep{xue-etal-2021-mt5} with speaker and genre metadata, achieving state-of-the-art performance on OntoNotes. I use the publicly released weights of their best-performing model, which I refer to as LinkAppend.

\paragraph{LingMess.} \citet{otmazgin2022lingmess} proposed an encoder-based CR model architecture and training procedure based on the mention-ranking paradigm. Using this recipe, they trained a state-of-the-art model named LingMess on the OntoNotes dataset, initialized with a Longformer-large encoder \citep{Beltagy2020Longformer}. For this evaluation, I train a LingMess model on each of the three training sets using the default hyperparameters and the Longformer-large encoder. Each training run was completed in under 24 hours on a single 40GB A100 GPU.

\begin{table*}[t]
    \centering
    \scriptsize
    \def\arraystretch{0.65}
    \begin{tabular}{lccccccccccccr}
    \multirow{3}{*}{\backslashbox{Test}{Train \vspace{2em}}} & \multicolumn{6}{c}{ON} & \multicolumn{3}{c}{PC} & \multicolumn{3}{c}{PD} & \\
    \cmidrule(lr){2-7}\cmidrule(lr){8-10}\cmidrule(lr){11-13}
& \multicolumn{3}{c}{LinkAppend} & \multicolumn{3}{c}{LingMess\textsubscript{ON}} & \multicolumn{3}{c}{LingMess\textsubscript{PC}} & \multicolumn{3}{c}{LingMess\textsubscript{PD}} \\
 \cmidrule(lr){2-4} \cmidrule(lr){5-7} \cmidrule(lr){8-10} \cmidrule(lr){11-13} & R & P & F1 & R & P & F1 & R & P & F1 & R & P & F1 & \quad \# Ments.  \\
\midrule
OntoNotes (CoNLL) & 83.7 & 82.9 & 83.3 & 79.2 & 82.6 & 80.9 & 64.1 & 48.8 & 55.4 & 33.5 & 61.4 & 43.3 & \\[.2em]
OntoNotes (B$^3$) & 83.2 & 82.0 & 82.6 & 77.6 & 81.6 & 79.5 & 62.5 & 45.8 & 52.9 & 28.0 & 57.5 & 37.7 & 19,764 \\[.2em]
\hdashline
\rule{0pt}{8pt}\quad Nested & 43.9 & 70.6 & 54.1 & 48.5 & 65.2 & 55.6 & 20.7 & 2.0 & \ccell{3.6} & \hphantom00.1 & \hphantom00.2 & \ccell{0.2} & 247 \\[.2em]
\quad ON Generic & 56.1 & 64.7 & 60.1 & 47.0 & 63.8 & 54.1 & 52.9 & 8.1 & \ccell{14.0} & 34.0 & 13.8 & 19.7 & 1,182 \\[.2em]
\quad Compound & 81.3 & 76.1 & 78.6 & 81.2 & 80.9 & 81.0 & 73.5 & 14.3 & 23.9 & \hphantom00.0 & \hphantom00.0 & \ccell{0.0} & 412 \\[.2em]
\quad Copular & 83.3 & 45.5 & 58.8 & 66.7 & 40.0 & 50.0 & 40.5 & \hphantom00.4 & \ccell{0.8} & \hphantom00.0 & \hphantom00.0 & 0.0 & 12 \\[.2em]
\midrule
OntoGUM (CoNLL) & 64.6 & 78.3 & 70.7 & 59.3 & 76.3 & 66.7 & 57.3 & 49.3 & 52.9 & 38.5 & 67.2 & 48.9 &  \\[.2em]
OntoGUM (B$^3$) & 65.0 & 75.6 & 69.9 & 59.3 & 73.6 & 65.7 & 58.3 & 46.0 & 51.4 & 33.0 & 61.8 & 43.0 & 2,707 \\[.2em]
\hdashline
\rule{0pt}{8pt}\quad Nested & 25.0 & 2.1 & 3.8 & \hphantom00.0 & \hphantom00.0 & \hphantom00.0 & \hphantom00.0 & \hphantom00.0 & \hphantom00.0 & \hphantom00.0 & \hphantom00.0 & \hphantom00.0 & 2 \\[.2em]
\quad ON Generic & 34.5 & 50.0 & 40.8 & 18.4 & 23.4 & 20.4 & 64.4 & \hphantom03.9 & \hphantom07.3 & 21.0 & \hphantom03.5 & \hphantom06.0 & 169 \\[.2em]
\quad Compound & 50.6 & 69.2 & 58.1 & 35.4 & 74.3 & 47.5 & 48.7 & 13.5 & 21.2 & \hphantom00.0 & \hphantom00.0 & \hphantom00.0 & 64 \\[.2em]
\bottomrule
    \end{tabular}
    \caption[Aggregated and disaggregated metrics intended to measure model performance at inferring coreference.]{
    Aggregated and disaggregated metrics intended to measure model performance. Scores are correlated with differences in how coreference is defined and operationalized across datasets, indicating that purported measurements of out-of-domain generalization also encompass these differences between datasets. Each column corresponds to a model trained on the respective training set: OntoNotes (ON), PreCo (PC), or Phrase Detectives (PD). For LinkAppend~\citep{bohnet-etal-2023-coreference}, we use the publicly released weights. For LingMess~\citep{otmazgin2022lingmess}, we train the model on each of the three training sets. Each row corresponds to the specified type of coreference. F1 scores are highlighted for types of coreference where the F1 score dropped significantly out-of-domain as compared to the overall difference in aggregate scores. “\# Ments.” refers to the number of annotated mentions.
    \label{tab:results-ontonotes}
    }
\end{table*}

\begin{table}[ht]
    \centering
    \scriptsize
    \def\arraystretch{0.65}
    \begin{tabular}{lcccr}
 & \multicolumn{3}{c}{LingMess} & \\
 \cmidrule(lr){2-4}
 Eval. & \multicolumn{1}{c}{ON} & \multicolumn{1}{c}{PC} & \multicolumn{1}{c}{PD} & \# Ments.  \\
\midrule
PreCo (C) & 64.5 & 77.1 & 50.8 & \\[.2em]
PreCo (B$^3$) & 64.2 & 76.3 & 47.3 & 25,983 \\[.2em]
\hdashline
\rule{0pt}{8pt}\quad Nested & \ccell{11.6} & 55.9 & \ccell{0.5} & 1,760 \\[.2em]
\quad ON Generic & \ccell{12.0} & 70.5 & 50.9 & 3,520 \\[.2em]
\quad Compound & \ccell{5.0} & 77.1 & \ccell{0.0} & 2,163 \\[.2em]
\quad Copular & \ccell{0.8} & 63.7 & \ccell{0.0} & 1,410 \\
\midrule
Phrase Det. (C) & 70.8 & 60.9 & 44.3 & \\[.2em]
Phrase Det. (B$^3$) & 69.8 & 56.9 & 36.8 & 3,259 \\[.2em]
\hdashline
\rule{0pt}{8pt}\quad Nested & 27.9 & 3.4 & 4.0 & 74 \\[.2em]
\quad ON Generic & \ccell{8.3} & 42.9 & 44.4 & 296 \\[.2em]
\quad Compound & 8.1 & 26.6 & \hphantom00.0 & 127 \\[.2em]
\quad Copular & \hphantom00.0 & \hphantom00.7 & \hphantom00.0 & 10 \\
\midrule
LitBank (C) & 65.5 & 50.9 & 36.6 & \\[.2em]
LitBank (B$^3$) & 66.3 & 47.0 & 32.9 & 2,267 \\[.2em]
\hdashline
\rule{0pt}{8pt}\quad Nested & 24.3 & 7.3 & \hphantom00.0 & 39 \\[.2em]
\quad ON Generic & 5.7 & 16.5 & 13.2 & 91 \\[.2em]
\quad Compound & \hphantom00.0 & \hphantom00.0 & \hphantom00.0 & 11 \\[.2em]
\quad Copular & 28.6 & 6.7 & \hphantom00.0 & 10 \\
\midrule
ARRAU (C) & 59.4 & 53.5 & 38.1 & \\[.2em]
ARRAU (B$^3$) & 57.8 & 51.2 & 33.2 & 6,132 \\[.2em]
\hdashline
\rule{0pt}{8pt}\quad Nested & 7.0 & 1.3 & 1.1 & 284 \\[.2em]
\quad ON Generic & 7.3 & 48.4 & 40.2 & 924 \\[.2em]
\quad Compound & 36.2 & 55.3 & \hphantom00.0 & 545 \\[.2em]
\quad Copular & 13.1 & 2.3 & 13.9 & 32 \\
\midrule
MMC (C) & 61.4 & 55.6 & 45.2 & \\[.2em]
MMC (B$^3$) & 57.3 & 51.0 & 40.5 & 5,338 \\[.2em]
\hdashline
\rule{0pt}{8pt}\quad Nested & 10.0 & 11.0 & \hphantom00.0 & 140 \\[.2em]
\quad ON Generic & 16.6 & 53.0 & 34.8 & 393 \\[.2em]
\quad Compound & \hphantom00.0 & 10.3 & 100.0 & 7 \\[.2em]
\quad Copular & 1.2 & 55.1 & 1.2 & 636 \\
\bottomrule
    \end{tabular}
    \caption[Disaggregated and disaggregated metrics intended to measure model performance at inferring coreference.]{Disaggregated F1 scores using the same setup as Table~\ref{tab:results-ontonotes}, as evaluated on additional test sets. F1 scores are highlighted when the difference relative to the in-domain model is significantly larger than the same difference calculated for all mentions. C indicates CoNLL F1 score.}
    \label{tab:results-all}
\end{table}

\section{Research Findings: Results}
\label{sec:measurement-results}

To examine how model performance varies across different types of coreference and different datasets, I conduct an extensive disaggregated evaluation. Summary results are presented in Table~\ref{tab:results-ontonotes} for models evaluated on OntoNotes and OntoGUM, and in Table~\ref{tab:results-all} for all other test sets.

This section illustrates how model failures correlate with differences in the way coreference is both defined and operationalized across training and test sets. In some of the datasets studied (e.g., PreCo), the coreference types central to this analysis are highly prevalent, representing up to 34\% of all annotated coreferring mentions (Table~\ref{tab:results-all}).

\paragraph{Performance varies significantly for coreference types defined differently across datasets.}
Consider the coreference types highlighted in Table~\ref{tab:results-ontonotes}. There are substantial performance variations (e.g., drops of over 40 percentage points) for generic and copular mentions when comparing the in-domain LingMess model with the out-of-domain models. While this variation could have multiple causes, a key factor is that both PreCo and Phrase Detectives (used to train the out-of-domain models) differ from OntoNotes in that these mention types are explicitly included in their definitions of coreference. As a result, for instance, while only a few copular structures are annotated as coreferring in OntoNotes, LingMess\textsubscript{PC} predicts 1,158 such structures. For example, it correctly identifies the following (according to the PreCo definition):
\vspace{-2pt}
\begin{quote}
    \small
    \ldots [One of the two honorable guests in the studio]\textsubscript{1} is [Professor Zhou Hanhug]\textsubscript{1} \ldots
\end{quote} \vspace{-4pt}

Consequently, both estimations of model performance and measurements of generalization will necessarily depend on which definition of coreference is used.

\paragraph{Performance also varies due to operationalization differences, even when definitions are consistent.}
Consider the case of OntoGUM in Table~\ref{tab:results-ontonotes}, a dataset constructed using the same definition of coreference as OntoNotes. This allows us to isolate the impact of potential differences in how that same definition was operationalized. Notably, models trained on OntoNotes perform poorly on nested mentions in OntoGUM, despite the datasets sharing the same annotation guidelines. While it is difficult to fully disentangle the cause due to the limited number of instances in OntoGUM, I conjecture based on a qualitative analysis that this may be partly due to how annotators interpreted those guidelines. For instance, the OntoGUM documentation describes the removal of i-within-i coreference relations. Upon manual inspection, many purported model errors on nested mentions in OntoGUM appear to be cases where all models correctly predict i-within-i constructions that were simply not annotated. Consider these examples, correctly predicted by LinkAppend but absent from the OntoGUM annotations:
\vspace{-2pt}
\begin{quote}
    \small
    Open-air markets, bookstores, and a [a Bart station all [its]\textsubscript{1} own]\textsubscript{1} make Rockridge another of many pleasant stops \ldots \\[1em]
    So there's nothing too special about [the bed [itself]\textsubscript{1}]\textsubscript{1}, but it does have a very important function \ldots 
\end{quote} \vspace{-2pt}

Such differences in how a single definition is operationalized by dataset creators can further distort measurements of both model performance and generalization.

\section{Discussion and conclusion}

While it may seem intuitive that measurements of CR model performance and generalization would be impacted by how coreference is defined and operationalized across datasets, existing multi-dataset evaluations have not explicitly considered how these differences might affect performance estimates and the conclusions we can draw about a model's ability to generalize.

Existing evaluations of model generalization mostly report accuracy using aggregate metrics \citep{moosavi-strube-2018-using,bamman-etal-2020-annotated,toshniwal-etal-2021-generalization}. Such aggregations, however, can obscure systematic errors that reveal deeper issues with the evaluation itself \citep{10.1145/3531146.3533233}. Understanding what is actually being measured requires a more explicit consideration of the measurement process. For this reason, I argue that measurement modeling provides a useful framework for analyzing and improving measurement in CR.

My findings highlight how relying on aggregated evaluations can obscure the reasons why models fail to generalize across datasets. This is especially true for a task like coreference, which involves multiple sub-phenomena that are inconsistently annotated across datasets. If we do not account for these inconsistencies, it will remain difficult to accurately interpret the results of performance evaluations that rely on these resources.

\subsection{Relative Ranking of Model Accuracy}

One might ask whether valid conclusions can still be drawn about the \textit{relative ranking} of models' accuracy, even if the absolute scores suffer from the measurement validity issues I have shown. Ultimately, however, ranking models based on empirical indicators is still a form of measurement and is therefore subject to the same considerations: \textit{Is there a consistent definition of the theoretical concept being measured? Are the measurements valid in that they encompass only those variables we intend to measure?}

Consider the following empirical example, comparing LinkAppend and LingMess\textsubscript{ON}. On four of the five out-of-domain test sets shown in Table~\ref{tab:results-all}, LinkAppend achieves a higher CoNLL F1 score than LingMess\textsubscript{ON}. On the LitBank test set, however, the ranking inverts: LingMess\textsubscript{ON} scores 65.5, while LinkAppend scores 64.3. A possible explanation for this reversal is that LingMess\textsubscript{ON} has a lower recall across most datasets (i.e., it predicts fewer mentions), and LitBank is annotated for only a restricted set of mention types. Therefore, a tendency to under-predict, which is penalized on other datasets, may incidentally lead to a better score on LitBank simply due to differences in how coreference is operationalized. In such a scenario, comparing the relative ranking of models has limited "convergent validity" (\textit{do measurements correlate with other measurements of the same construct?}) \citep{jacobs2021measurement} because the ranking itself is not consistent across datasets that purport to measure the same underlying ability.

\paragraph{Takeaway.}

I propose viewing the evaluation of coreference, CR model performance, and CR model generalization through the lens of measurement modeling. This perspective requires a clear distinction between the theoretical construct intended to be measured and its operationalization via a measurement model. Adopting this framework allows one to more clearly consider what is being measured and to identify potential concerns with the validity of those measurements.

Through a disaggregated, out-of-domain evaluation of CR models, I have shown that models are systematically limited in their ability to generalize to certain types of coreference, particularly those that differ in how they are defined or operationalized across datasets. This analysis also showed that models display limited generalization even when evaluated on datasets that are intended to share the same annotation guidelines. This may be attributable in part to differences in how the annotation guidelines were operationalized, for instance via direct annotation as in OntoNotes versus heuristic transformation as in OntoGUM. Further investigation based on additional qualitative analysis is a promising direction for future work.

If the research goal is to develop CR models that generalize a consistent conceptualization of coreference across datasets, then iteratively improving in-domain performance will have limited impact. A model's ability to generalize will remain inherently constrained by operationalization differences between datasets and their corresponding evaluation schemes. For valid cross-dataset evaluations, researchers must consider differences in how concepts were defined and operationalized which is not always clearly documented, such as in cases where no definition of the construct nor annotation guidelines are provided~\citep{chen-etal-2018-preco}.

\paragraph{Measurement Modeling as a Blueprint.}
Finally, the analysis in this chapter provides a blueprint for evaluating out-of-domain performance in a way that accounts for known inconsistencies in how coreference is defined and operationalized. In particular, these experiments highlight how explicitly considering the definition and operationalization of the construct being measured, combined with disaggregating performance results, can lead to a deeper understanding of what multi-dataset evaluations reveal about a model's ability to generalize. This approach is a necessary step toward meaningfully evaluating models of contested constructs across multiple datasets.

Thus, the preceding chapters have established a multi-faceted understanding of measurement validity in coreference evaluation. I have demonstrated issues of \textit{convergent validity} by bridging canonical and challenge set evaluations, and further explored the challenges of \textit{content validity} (through the lens of contestedness) and \textit{discriminant validity} by examining canonical evaluations in detail.

\NewDocumentCommand{\notsignif}{}{\makebox[0pt][l]{$^\dagger$}}

\chapter{A Controlled Re-evaluation of Coreference Resolution Systems}
\label{chap:supervised-comparison}

\myepigraph{If you listen to how people communicate with each other, they are never nice, polished conversations. Certain statements are actually interpreted differently by different people, and they often do not even realize that there is another possible interpretation.\vspace{10pt}}{\citet{poesioAiNed}}

This next chapter is a short follow-up to the previous experiments focusing on the evaluation design itself in terms of comparing models. This brief extension further solidifies issues with discriminant validity in canonical evaluations.

Given that all state-of-the-art CR models involve finetuning a pretrained language model, a critical question arises: is the superior performance (according to standard evaluation metrics) of one model over another attributable to the underlying language model or to other factors, such as the task-specific architecture? I argue that without a standardized experimental setup, it is difficult to disentangle these factors. This presents a challenge to discriminant validity, as we cannot be certain what is being measured—the effectiveness of the CR architecture or the capabilities of the base language model.

To address this ambiguity, I conduct a systematic evaluation of five CR models, controlling for key design decisions, most notably the pretrained language model used by each. The results of this controlled comparison reveal that when language model size is held constant, encoder-based CR models outperform more recent decoder-based approaches in both accuracy and inference speed. Surprisingly, among the encoder-based models, newer architectures are not consistently more accurate; in fact, the oldest model tested demonstrates the best generalization to out-of-domain textual genres.

Ultimately, I conclude that controlling for the choice of language model accounts for most, but not all, of the reported increase in F1 score over the past five years. This finding is not only unexpected but also reinforces the critical need to explicitly consider and address the discriminant validity of our evaluation methodologies, a central theme of this thesis.

\section{Overview of Methodology}
\label{sec:reeval-introduction}

In recent years, all state-of-the-art CR models have come to rely on the supervised finetuning of a pretrained language model (see \citet{poesio2023computational} for a comprehensive survey). This reliance motivates the central question of this chapter:

\vspace{0.5em}
\noindent
\textit{To what extent are recent improvements in CR due to the use of a more powerful language model, as opposed to other design decisions such as model architecture?}
\vspace{0.5em}

\noindent
To answer this question, I evaluate several existing CR models while controlling for the language model used in each. The results demonstrate that at comparable language model sizes, encoder-based CR models consistently outperform more recent decoder-based models in terms of accuracy, inference speed, and memory usage.

This performance advantage for encoder-based models holds even when scaled to sizes beyond what has been tested in existing work. For instance, when scaled to 1.5B parameters, the encoder-based LingMess model \citep{otmazgin-etal-2023-lingmess} achieves an 82.5 CoNLL F1 score on the CoNLL-2012 Shared Task test set. This matches the score of the decoder-based ASP model, which requires 11B parameters to achieve the same result \citep{liu-etal-2022-autoregressive}.

Furthermore, when controlling for additional factors like the hyperparameter search space, I find that the oldest model tested, C2F \citep{lee-etal-2018-higher}, performs competitively with all other encoder-based models. Notably, C2F generalizes better to out-of-domain textual genres than all other models of a comparable size.

\textit{Based on these results, I conclude that controlling for the choice of language model accounts for most, but not all, of the increase in F1 score reported in the past five years. Many apparent improvements in CR model accuracy may therefore be attributable to the use of a stronger underlying language model rather than to novel architectural designs.} This finding suggests that future work proposing to improve CR accuracy should carefully disentangle these factors to better understand the true source of performance gains.

The experiments in this chapter focus on five CR models: C2F \citep{lee-etal-2018-higher}, S2E \citep{kirstain-etal-2021-coreference}, WLC \citep{dobrovolskii-2021-word}, LingMess \citep{otmazgin-etal-2023-lingmess}, and Link-Append \citep{bohnet-etal-2023-coreference}. The evaluation centers on English-language, document-level, nominal-entity coreference as formulated in OntoNotes \citep{hovy-etal-2006-ontonotes}, with additional generalization tests on the GAP \citep{webster-etal-2018-mind} and OntoGUM \citep{zhu-etal-2021-ontogum} corpora.

The contributions of this chapter are therefore threefold. First, it establishes a reliable experimental baseline by reimplementing five state-of-the-art CR models and reproducing their original, published results. Second, it demonstrates that when controlling for language model choice and hyperparameter search, more recent CR models are not always more accurate, challenging the narrative of simple linear progress. Finally, it shows that encoder-based CR models can be scaled to 1.5B parameters, where they outperform decoder-based models of a comparable or even larger size, highlighting a promising direction for future efficient and accurate models.

\section{Background}
\label{sec:reeval-related-work}

This chapter follows in the tradition of existing work that has presented controlled comparisons of published models, particularly in the domain of language model pretraining \citep{melis2018on,nityasya-etal-2023-scientific}. While motivated by a similar scientific principle, my focus is specifically on coreference resolution (CR), with a particular emphasis on disentangling the effects of the base language model from the task-specific architecture.

Previous attempts to understand state-of-the-art CR models have primarily involved targeted error analyses \citep{stoyanov-etal-2009-conundrums,lu-ng-2020-conundrums} or studies of model generalization across different datasets \citep{toshniwal-etal-2021-generalization,zabokrtsky-etal-2022-findings,porada2023investigating}. However, there has been limited work that systematically evaluates the design of existing models in a controlled setting. \citet{martschat-strube-2015-latent} performed a comprehensive evaluation of CR models, but their analysis was confined to pre-neural architectures. More recent work has tended to analyze specific components of the CR pipeline rather than full model architectures in a comparative setting \citep{toshniwal-etal-2020-cross,xu-choi-2020-revealing,lai2022end}.

While many publications on new CR models include ablations that allow for pairwise comparisons with certain predecessors \citep{dobrovolskii-2021-word,liu-etal-2022-autoregressive}, such comparisons are often not on a level playing field. My study provides novel insights by training all models with the same competitive language model wherever possible, thereby creating a more standardized basis for comparison. As the results will show, this controlled approach reveals that the accuracy of older models, such as C2F, has been underestimated in existing comparisons.

\section{Method}
\label{sec:reeval-methods}

In this section, I detail each of the models included in my analysis, the experimental factors that I control for, and the datasets used for evaluation.

\subsection{Models}
\label{sec:reeval-models}

My evaluation focuses on four encoder-based models and one decoder-based model, all of which reported state-of-the-art accuracy at their respective times of publication.

\subsubsection{Encoder-based Models}
\label{sec:reeval-encoder-based-models}

\paragraph{C2F} The Coarse-to-Fine (C2F) model \citep{lee-etal-2018-higher} is an extension of the earlier E2E model \citep{lee-etal-2017-end}. The E2E model functions by encoding text spans as contextualized vectors and then scoring these representations pairwise using a task-specific head. C2F extends this approach with an intermediate bilinear scoring function that prunes the number of spans considered. While \citet{lee-etal-2018-higher} also proposed higher-order inference (HOI), I specifically consider the E2E+C2F configuration (referred to simply as C2F), as HOI is known to have only a marginal effect on performance~\citep{xu-choi-2020-revealing}. Although the original C2F model did not finetune the language model encoder, I follow the hyperparameters of \citet{joshi-etal-2019-bert}, which was the first work to do so.

\paragraph{S2E} The Start-to-End (S2E) model \citep{kirstain-etal-2021-coreference} is based on C2F, with the primary distinction that span representations are created using only the embeddings of the first and last tokens. By contrast, span representations in C2F are a weighted sum of all token embeddings in the span. S2E was motivated as a more memory-efficient approach that could maintain high accuracy.

\paragraph{WLC} The Word-Level Coreference (WLC) model \citep{dobrovolskii-2021-word} is also derived from C2F. In WLC, individual tokens are scored as candidate mentions, each representing the headword of a span. WLC was motivated as being faster than C2F and more accurate when using a RoBERTa language model \citep{liu2019roberta}.

\paragraph{LingMess} \citet{otmazgin-etal-2023-lingmess} proposed LingMess as a direct extension of S2E, primarily by increasing the number of task-specific, span-scoring heads. LingMess achieved a significant improvement in accuracy when using a Longformer language model \citep{Beltagy2020Longformer} as the encoder.

\subsubsection{Decoder-based Models}
\label{sec:reeval-decoder-based-models}

Two competitive decoder-based CR models have been proposed in the literature: ASP \citep{liu-etal-2022-autoregressive} and Link-Append. Of these, I evaluate Link-Append, which reported a higher F1 score on OntoNotes, although I compare against ASP using published results where possible.

\paragraph{Link-Append} \citet{bohnet-etal-2023-coreference} proposed Link-Append, a method for finetuning a language model for CR purely as a sequence generation task. For each sentence, the model is trained to generate a string representing all coreferring mentions within that sentence. When used to finetune the 13B-parameter mT5 language model \citep{xue-etal-2021-mt5}, Link-Append demonstrated state-of-the-art accuracy.

\subsubsection{Implementation}

I reimplement all models using Hugging Face's Transformers library \citep{wolf-etal-2020-transformers}, adhering to each model's original hyperparameters. This reimplmentation was in collaboration in Xiyuan Zou as noted in the contribution of authors section of this thesis.\footnote{Our code is available at \url{https://github.com/ianporada/coref-reeval}} This standardized setting allows for a direct comparison of empirical inference speed and memory usage. Furthermore, it facilitates controlling for factors such as the language model, avoiding the manual preprocessing steps often required by the original implementations. My reimplementations are based on the original model code where available.

\begin{table}[t]
    \small
    \centering
    \begin{tabular}{llccc}
        \toprule
        Model & Language Model & Prior & Ours \\
        \midrule
        C2F & BERT-large & 76.9 & 77.1 \\
        S2E & Longformer-large & 80.3 & 80.4 \\
        WLC & RoBERTa-large & 81.0 & 81.0 \\
        LingMess & Longformer-large & 81.4 & 81.4 \\
        Link-Append & mT5-XXL & 83.3 & 83.3 \\
        \bottomrule
    \end{tabular}
    \caption[Test set performance on OntoNotes (CoNLL-2012) as reported in prior work and for our reimplementation of the model.]{\label{tab:reimplementation-comparison}
        Test set performance on OntoNotes (CoNLL-2012) as reported in prior work (``prior'') and for our reimplementation of the model (``ours''). For each model, we use the best configuration of the respective prior work.
    }
\end{table}

To verify my reimplementations, I compare their performance against the best published configurations (Table~\ref{tab:reimplementation-comparison}). As I lack the resources to train the 13B-parameter Link-Append model, I verify my data processing by running inference using the publicly released model weights and have also clarified preprocessing details with the original authors.

\subsection{Controlled Factors}
\label{sec:reeval-controlling-for-confounds}

To determine whether the improved accuracy of a given CR model is due to its proposed design versus other factors, my evaluation controls for several variables. The primary factor is the language model used to initialize each CR model. Additionally, because more recent models have often been trained with different hyperparameters (\textit{e.g.,} more epochs), I also control for these potential confounds by evaluating all models over a consistent hyperparameter search space.

\subsubsection{Language Model}

\paragraph{Encoder} For the encoder-based CR models, I train all models using the same competitive language model, DeBERTa \citep{he2021deberta}, at both base and large sizes. I selected DeBERTa due to its strong performance on the SuperGLUE benchmark \citep{NEURIPS2019_4496bf24}.

\paragraph{Encoder vs. Decoder} A direct comparison between encoder- and decoder-based CR models using the exact same language model is not feasible, as these model classes require different underlying LM architectures. Instead, I control for language model size. I train the best-performing encoder-based model, LingMess, using DeBERTa encoders ranging from 138M to 1.5B parameters. I then compare its performance against the Link-Append and ASP decoder-based models, which were trained with language models ranging from 300M to 13B parameters.

\subsubsection{Hyperparameter Search Space}

\paragraph{Finetuning Epochs} More recent models have tended to be trained for longer. For example, \citet{joshi-etal-2019-bert} trained C2F for 20 epochs, whereas LingMess was trained for 129 epochs, and the 100k steps used for Link-Append correspond to roughly 160 epochs. To control for this as a source of improved accuracy, I perform a search over training epochs in $\{25, 50, 125\}$ for each model.

\paragraph{Task-specific Head Size} Encoder-based models after C2F have used a larger hidden-layer size for their task-specific heads (\texttt{ffnn\_size}). To control for the potential impact of this increase, I search over \texttt{ffnn\_size} $\in \{1024, 2048, 3072, 4096\}$ for every encoder-based model, a space that includes all sizes used in prior work.

\paragraph{Input Size} During training, the original C2F implementation reduced document length to a small, fixed number of sentences to manage memory usage. Other models do not have this constraint. I remove this limitation by training C2F on the maximum document length that fits into memory, ensuring a fairer comparison.

\subsection{Datasets}
\label{sec:reeval-datasets}

\paragraph{OntoNotes (ON)} As is standard in existing work, I train and evaluate all models on the coreference annotations from the English-language portion of the OntoNotes 5.0 dataset \citep{ontonotes5}. Specifically, I use the CoNLL-2012 Shared Task v4 dataset splits \citep{pradhan-etal-2012-conll} and the official CoNLL-2012 scorer for evaluation. The train, validation, and test splits contain 1940, 343, and 348 document parts, respectively.

\paragraph{OntoGUM (OG)} To assess generalization to out-of-domain genres, I also evaluate models on the OntoGUM (OG) dataset \citep{zhu-etal-2021-ontogum}. This dataset contains coreference annotations from the English-language GUM corpus \citep{Zeldes2017}, which were heuristically transformed to follow the OntoNotes annotation guidelines \citep{zhu-etal-2021-anatomy}. When evaluating on OG, I use all 213 documents in the dataset.

\paragraph{GAP} Finally, for a targeted evaluation of pronominal anaphora resolution, I evaluate models on the 2,000-example validation set of the GAP dataset \citep{webster-etal-2018-mind}. GAP consists of pronouns from English Wikipedia, each annotated for coreference with respect to two preceding noun phrases. I score model performance using the official GAP scorer.

\section{Experimental Findings}
\label{sec:reeval-experiments}

\begin{table}[t]
    \small
    \centering
    \begin{tabular}{lccccc}
        \toprule
        \multicolumn{6}{c}{\bf DeBERTa-base (138M)} \\
        \midrule
        Method & ON & OG & GAP & Mem. & Time \\
        \midrule
        C2F & 77.3 & 63.5 & 77.4 & 6.2 & 71.6 \\
        S2E & 78.6 & 63.8 & 78.9 & \textbf{1.3} & \textbf{20.9} \\
        WLC & 79.1 & 64.1 & 78.7 & 1.4 & 25.9 \\
        LingMess & \textbf{79.5} & \textbf{64.3} & \textbf{79.4} & 1.7 & 54.7 \\
        \midrule
        \midrule
        \multicolumn{6}{c}{\bf DeBERTa-large (405M)} \\
        \midrule
        Method & ON & OG & GAP & Mem. & Time \\
        \midrule
        C2F & 80.8 & \textbf{66.9} & 79.1 & 8.7 & 129.3 \\
        S2E & 80.5 & 65.9 & \textbf{79.9} & \textbf{2.7} & \textbf{45.3} \\
        WLC & 81.0 & 66.2 & 79.1 & 2.8 & 47.4 \\
        LingMess & \textbf{81.7} & 66.4 & 79.4 & 3.1 & 81.6 \\
        \bottomrule
    \end{tabular}
    \caption[Dev. set performance on OntoNotes using a DeBERTa encoder of the given size.]{\label{tab:encoder-comparison}
        Dev. set performance using a DeBERTa encoder of the given size. ON and OG are evaluated by CoNLL F1 score, GAP by F1 score. Mem. is max memory at batch size of one (GB), and time is inference speed in ms/doc at max batch size, both calculated w.r.t. inference on ON using a single 80GB A100 GPU.
    }
\end{table}

Here I present the results from the experiments controlling for language model size and the hyperparameter search space. My findings show that controlling for these factors considerably narrows the reported gap in accuracy between models.

\paragraph{More Recent Encoder-based Models Are Not Always More Accurate}

First, I test all CR models in their original configurations, with the minimal change of substituting a standardized DeBERTa encoder (Table~\ref{tab:encoder-comparison}). When using DeBERTa-large, the difference in CoNLL F1 score on OntoNotes between all models reduces to less than one absolute point. Furthermore, the older models (C2F and S2E) prove to be the most accurate on the out-of-domain datasets. In this setup, every encoder-based model also performs at or better than its best configuration reported in the literature.

\begin{table*}[t]
    \small
    \centering
    \begin{tabular}{llc @{\hskip 3em} ccc @{\hskip 3em} cc}
        \toprule
        Model & LM & Size & ON & OG & GAP & Mem. & Time \\
        \midrule
        C2F & \multirow{4}{*}{DeBERTa$_\text{L}$} & 452M & 81.2\notsignif & \textbf{66.9} & 78.1 & 8.7 & 129.3 \\
        S2E & & 465M & 80.5 & 65.9\notsignif & \textbf{79.9} & \textbf{2.7} & \textbf{45.3} \\
        WLC & & 664M & 81.1\notsignif & 65.8\notsignif & 78.8 & 2.8 & 47.4 \\
        LingMess & & 561M & \textbf{81.7} & 66.4 & 79.4 & 3.1 & 81.6 \\
        \midrule
        \multirow{2}{*}{Link-Append} & mT5$_\text{B}$ & 582M & 66.2 & 45.8 & 63.3 & 5.9 & 5.2$\mathrm{e}$4 \\
        & mT5$_\text{L}$ & 1.23B & 66.5 & 45.5 & 64.9 & 12.3 & 1.4$\mathrm{e}$5 \\
        \bottomrule
    \end{tabular}
    \caption[Dev. set accuracy on OntoNotes when controlling for multiple confounding factors.]{\label{tab:cr-models-comparison}
        Dev. set accuracy when controlling for all factors. ON and OG are evaluated by CoNLL F1 score, GAP by F1 score. Mem. is the max memory usage for inference at batch size one (GB), and time is inference speed (ms/doc) at max batch size, both w.r.t. ON inference using a single 80 GB A100 GPU.
        F1 scores without $\dagger$ are statistically significantly different from all other scores based on a permutation test with $\alpha=0.1$ and 10k permutations following \citet{chinchor-1995-statistical}.
        }
\end{table*}

Next, I control for the hyperparameter search space by comparing all models in their optimal hyperparameter configuration (Table~\ref{tab:cr-models-comparison}). Under this fully controlled setting, the performance difference between the oldest (C2F) and most recent (LingMess) encoder models shrinks to just 0.5 absolute points on OntoNotes. For all models in Table~\ref{tab:cr-models-comparison}, full precision, recall, and test set results are available in my project's GitHub repository, as are human-readable model predictions in CoNLL format.

\paragraph{Encoder-based Models Are More Accurate than Decoder-based Models at Comparable Sizes}

Table~\ref{tab:cr-models-comparison} also shows the accuracy and inference time for all encoder-based models versus the Link-Append model trained at approximately the same size. At a comparable model size, the encoder-based models are substantially more accurate and faster than Link-Append.

\begin{figure}[ht]
    \centering
    \includegraphics[width=0.8\textwidth]{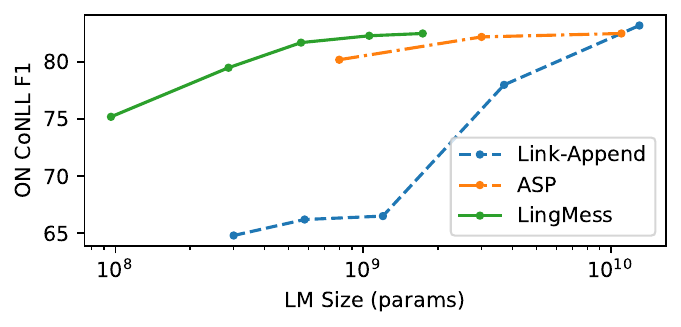}
    \caption[OntoNotes (ON) test set CoNLL F1 score for CR models trained with language models at multiple scales.]{OntoNotes (ON) test set CoNLL F1 score for CR models trained with language models at multiple scales. Link-Append is trained with mT5, ASP with Flan-T5, and LingMess with DeBERTa. ASP scores, as well as the two largest Link-Append model scores, were reported by the respective authors. \citet{bernd-crac2023-keynote} also noted that Link-Append has an unexpected, apparently ``emergent'', scaling.}
    \label{fig:scaling}
\end{figure}

The accuracy of models trained over a range of sizes is shown in Figure~\ref{fig:scaling}. Up to the largest size I trained, a 1.5B-parameter LingMess model, the encoder-based approach outperforms the decoder-based methods.

\begin{figure*}[ht]
    \centering
    \includegraphics[width=1.0\textwidth]{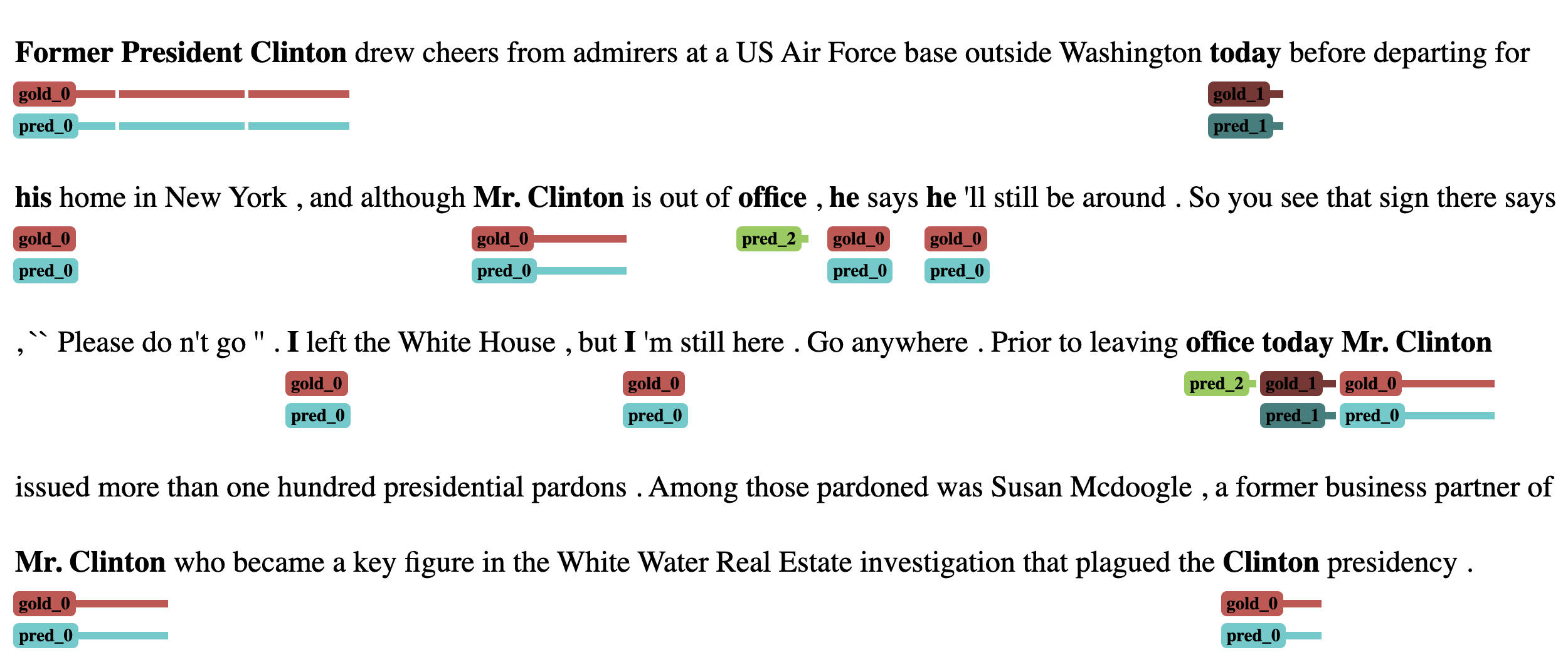}
    \caption[An example model output on the OntoNotes development set.]{An example model output on the OntoNotes development set. ``Gold'' spans denote the ground-truth annotations, and ``pred'' spans denote predictions by the Link-Append mT5-XXL model.}
    \label{fig:example-output}
\end{figure*}

\subsection{Discussion and Conclusion}

My results may be surprising because CR models more recent than C2F, such as WLC and S2E, are often reported to be more accurate in published comparisons \citep{chai-strube-2022-incorporating,liu-etal-2022-autoregressive,bohnet-etal-2023-coreference}; yet, I found the opposite to be true when controlling for the choice of language model. I suggest that future work explicitly control for these factors when presenting comparisons of different CR architectures to ensure that performance gains are correctly attributed.

In some cases, less accurate models are substantially more efficient in terms of memory or time (Table~\ref{tab:cr-models-comparison}). While these dimensions have been considered in previous work \citep{kirstain-etal-2021-coreference}, even these efficient models have largely been motivated by their raw accuracy. To this end, I argue for more holistic evaluations of coreference resolution systems that look beyond a single accuracy metric, especially given that a high OntoNotes F1 score does not necessarily translate to strong out-of-domain performance.

In my controlled comparison, the WLC model is never Pareto-optimal when considering accuracy, runtime, and memory usage simultaneously. In contrast, all other encoder-based models are optimal in at least one of these three dimensions.

\paragraph{Takeaway.}

In this chapter, I have reevaluated existing CR models, controlling for confounding factors such as the choice of language model and the number of training epochs. This controlled analysis revealed that among encoder-based models, the relative performance gap on OntoNotes narrows to just 0.5 absolute CoNLL F1 score, and the oldest model tested (C2F) generalizes best to out-of-domain textual genres. At all scales tested---up to 1.5B parameters---encoder-based CR models outperform decoder-based models of comparable size; notably, the LingMess model trained with a 1.5B DeBERTa encoder matches the accuracy of decoder-based models that are over seven times larger.

These findings provide strong support for this thesis's central argument: that conclusions from coreference evaluations are often limited by unexamined design choices. Specifically, this analysis highlights a critical issue of \textit{discriminant validity}, where the conflation of the base language model's power with the CR model's architectural merit has obscured the true sources of performance gains. The fact that model rankings change between in-domain and out-of-domain settings further underscores issues of \textit{convergent validity}, demonstrating that high performance on a standard benchmark does not guarantee robust generalization. By adopting a more controlled evaluation setting, we can move beyond existing benchmark performance and begin to understand to a slightly higher degree what makes a CR model effective and generalizable.

\chapter{Inferences Depending on Semantic Plausibility}
\label{chap:semantic-plausibility}

\myepigraph{The more I research into language and common sense, the weirder it gets. Common sense, in my view, is the dark matter of intelligence and language.}{\citet{yejin_choi_commonsense}}

The preceding chapters have connected ``challenge set'' and ``canonical'' evaluations of coreference resolution, highlighting issues of convergent validity between these two styles of evaluation. I have also explored related issues of contestedness and discriminant validity within canonical evaluations. Furthermore, these analyses have provided initial evidence supporting the hypothesis that language models make inferences that lack the kind of consistent, intuitive generalization characteristic of human reasoning.

This final experimental chapter, therefore, shifts the focus specifically to challenge sets designed to evaluate the inference of semantic plausibility. This investigation serves two primary purposes. First, it demonstrates that a more controlled experimental design can help overcome some of the previously discussed measurement validity issues by being more precise about what is being measured. Second, through a novel evaluation methodology, I show more formally that language models can make inferences that exhibit conceptual inconsistencies and a failure to generalize across semantically related cases in a manner that humans would not.

The premise for these experiments is that resolving coreference, and understanding language more broadly, requires commonsense reasoning, a key component of which is the ability to discern the plausibility of events. While recent approaches---most notably pre-trained Transformer language models---have demonstrated significant improvements in modeling event plausibility, their performance still falls short of human capabilities.

In this chapter, I demonstrate that Transformer-based plausibility models are markedly inconsistent across the conceptual classes of a lexical hierarchy. For instance, a model might infer that ``a person breathing'' is plausible while simultaneously inferring that ``a dentist breathing'' is not. I find this inconsistency persists even when models are provided with lexical knowledge, and I present a simple post-hoc method for enforcing model consistency that improves correlation with human plausibility judgments.

Crucially, this highly controlled approach considers to some extent the measurement validity issues previously mentioned: I focus on a more narrow phenomenon for which results appear consistent across datasets. This enhanced validity, however, comes at the cost of generalizability; whereas the findings of canonical evaluations might apply to natural language broadly, the findings of this chapter are specific to a controlled phenomenon within a particular domain of language.

\section{Overview of Methodology}

A human reader can easily discern that of the following events, (\ref{itm:typical}) and (\ref{itm:atypical}) are semantically plausible, while (\ref{itm:implaus}) is nonsensical.

\begin{enumerate}
\setlength\itemsep{-2pt}
    \item \label{itm:typical} The person breathes the air.
    \item \label{itm:atypical} The dentist breathes the helium.
    \item \label{itm:implaus} The thought breathes the car.
\end{enumerate}

This ability to judge semantic plausibility is critical for understanding natural language. Specifically, modeling \textit{selectional preference}---the semantic plausibility of predicate-argument structures---is known to be an implicit requirement for discriminative tasks such as coreference resolution \citep{hobbs1978resolving, ido-anaphora-res, zhang-etal-2019-knowledge}, word sense disambiguation \citep{resnik-1997-disambig, mccarthy-disambig}, textual entailment \citep{zanzotto-etal-2006-discovering, pantel-etal-2007-isp}, and semantic role labeling \citep{gildea-jurafsky-2002-automatic, zapirain-etal-2013-selectional}. More broadly, modeling semantic plausibility is a necessary component of generative inferences such as conditional commonsense inference \citep{Gordon2011SemEval2012T7, zhang-etal-2017-ordinal}, abductive commonsense reasoning \citep{Bhagavatula2020Abductive}, and commonsense knowledge acquisition \citep{ijcai2020-554, hwang2020cometatomic}.

\begin{figure}
\centering
  \includegraphics[width=7.25cm]{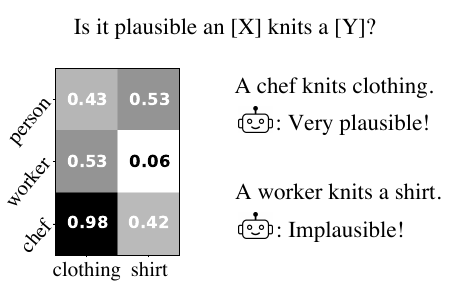}
  \caption[The relative plausibility score for the event ``an X knits a Y'' as output by a RoBERTa model.]{
  Elements in the matrix are the relative plausibility score for the event ``an [X] knits a [Y]'' as output by a RoBERTa model fine-tuned to model plausibility. [X] and [Y] correspond to the label of the row and column, respectively. Model scores are inconsistent with respect to the two events shown on the right.
  }
  \label{fig:intro-example}
\end{figure}

However, learning to model semantic plausibility is a difficult problem. Language is sparse, so most events are not directly attested in a corpus. Furthermore, plausibility relates to likelihood in the real world, which is distinct from the likelihood of an event's textual description. Ultimately, plausibility reflects human intuition, and modeling it fully requires ``the entire representational arsenal that people use in understanding language, ranging from social mores to naive physics'' \citep{resnik_1996}.

A key property of plausibility is that it is generally consistent across appropriate levels of conceptual abstraction. For example, events of the form ``the [\textsc{person}] breathes the [\textsc{gas}]'' are consistently plausible because people understand that similar concept classes share similar affordances. In this chapter, I investigate whether state-of-the-art plausibility models, which are based on fine-tuning Transformer language models, likewise exhibit this conceptual consistency. To facilitate this investigation, I first propose two novel measures for quantifying the consistency of plausibility estimates across conceptual abstractions.

As the results will demonstrate, conceptual inconsistency is a significant issue in existing models, leading to erroneous predictions (see Figure~\ref{fig:intro-example}). To address this, I explore two methods that endow Transformer-based plausibility models with knowledge from a lexical hierarchy, hypothesizing that these methods might correct conceptual inconsistency without causing over-generalization. The first method provides lexical knowledge as an additional input to the model, making no \textit{a priori} assumptions about how the model should generalize. The second method explicitly enforces conceptual consistency by defining the plausibility of an event as a maximum over the plausibility of all its conceptual abstractions.

My findings show that simply injecting lexical knowledge does not resolve the issue, but the second method, which presents a post-hoc technique for generalizing plausibility estimates, is necessarily consistent and improves correlation with human plausibility judgments. This result encourages future work that forces Transformer models to make more discrete abstractions to better model plausibility.

The analysis in this chapter focuses on simple events in English represented as subject-verb-object (s-v-o) triples. Models are evaluated by their correlation with two datasets of human plausibility judgements. The baseline models build on RoBERTa \citep{Liu2019RoBERTaAR}, a pre-trained Transformer masked language model, and use WordNet 3.1 \citep{wordnet} hypernymy relations as a lexical hierarchy.\footnote{My implementation and data are available at \url{https://github.com/ianporada/modeling_event_plausibility}}

\section{Background and Framework}

\begin{figure}[ht]
\centering
  \includegraphics[width=5cm]{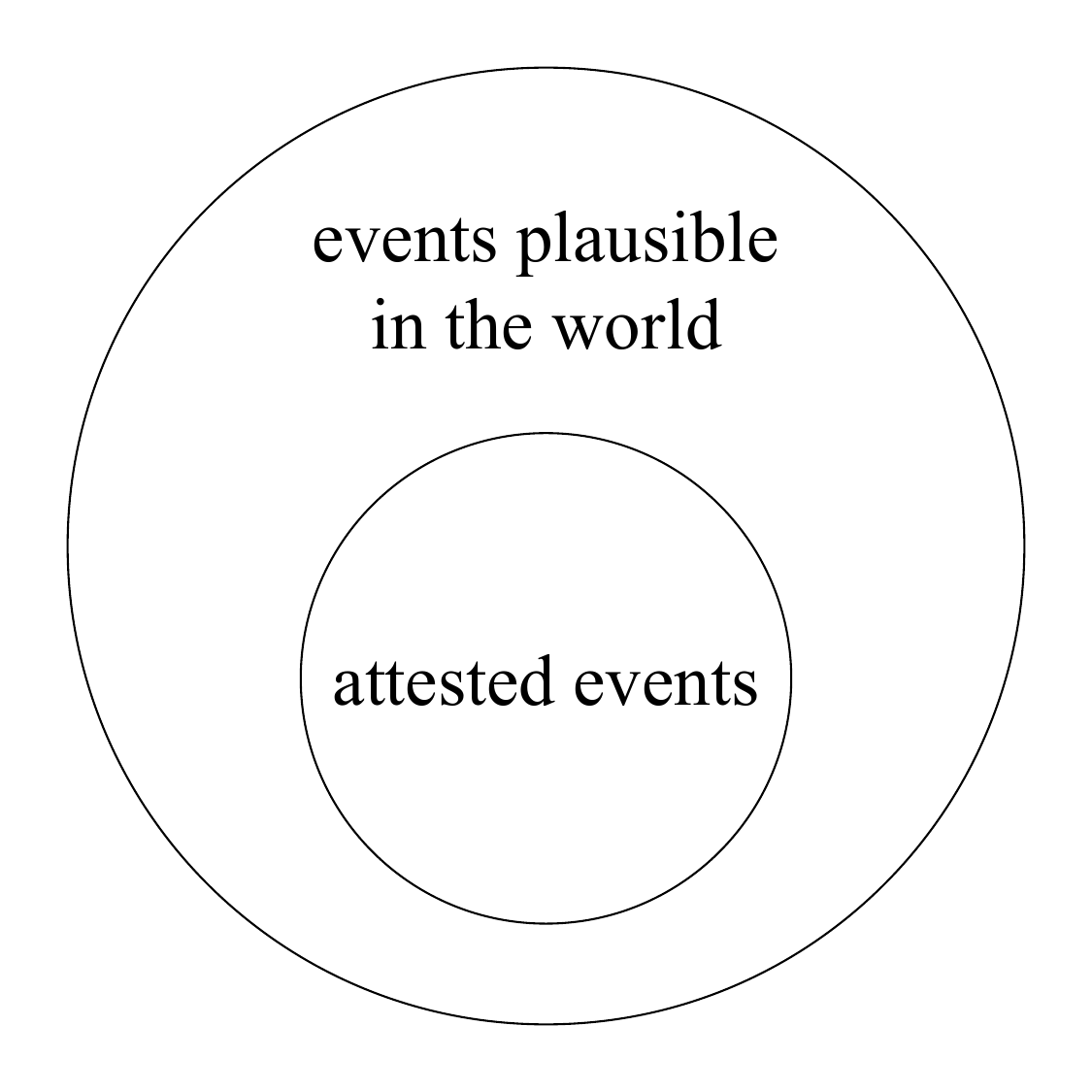}
  \caption[A visualization of the plausible versus possible distinction.]{An attested event is necessarily plausible in the world, but not all plausible events are attested. By ``the world'' we refer to some possible world under consideration---in this sense, plausibility is an epistemic modality.}
  \label{fig:plausibility}
\end{figure}

While plausibility is difficult to define precisely, I adopt the following useful distinctions from the literature:

\begin{itemize}
\setlength\itemsep{0.2em}
    \item Plausibility is a matter of degree \citep{WILKS197553, resnik_selection_1993}. I, therefore, evaluate models by their ability to estimate the relative plausibility of events.
    \item Plausibility describes non-surprisal conditioned on some context \citep{resnik_selection_1993, Gordon2011SemEval2012T7}. For example, conditioned on the event ``breathing,'' it is less surprising to learn that the agent is ``a dentist'' than ``a thought'' and is thus more plausible.
    \item Plausibility is dictated by the likelihood of occurrence in the world rather than in text \citep{zhang-etal-2017-ordinal,wang-etal-2018-modeling}. This discrepancy is due to reporting bias---the fact that people do not state the obvious \citep{Gordon:2013:RBK:2509558.2509563, shwartz-choi-2020-neural}; e.g., ``a person dying'' is more likely to be attested in a corpus than ``a person breathing'' (Figure \ref{fig:plausibility}).
\end{itemize}

The problem formulation used in this chapter follows that of \citet{wang-etal-2018-modeling}, who show that static word embeddings lack the necessary world knowledge for modeling plausibility. The current state-of-the-art approach is to take the conditional probability of co-occurrence, as estimated by a distributional model, as an approximation of event plausibility \citep{ijcai2020-554}. The fine-tuned RoBERTa baseline used in this chapter follows this paradigm.

Similar in spirit to this chapter's investigation, \citet{He2020OnTR} extend this baseline method by creating additional training data using the Probase taxonomy \citep{probase} to improve conceptual generalization. Specifically, for each training example, they swap an event's arguments with its hypernym or hyponym and treat this new, perturbed example as an implausible event.

There is also a body of recent work focusing on monotonic inferences in semantic entailment \citep{yanaka-etal-2019-neural,goodwin-etal-2020-probing,geiger-etal-2020-neural}. Plausibility contrasts with entailment in that it is not strictly monotonic with respect to hypernymy/hyponymy relations: the plausibility of an event involving an entity is not sufficient to infer the plausibility of its hyponyms (i.e., not downward-entailing: it is plausible that a \textit{person} gives birth but not that a \textit{man} gives birth) nor its hypernyms (i.e., not upward-entailing: it is plausible that a \textit{baby} fits inside a shoebox but not that a \textit{person} does).

Non-monotonic inferences have recently been explored in the context of defeasible reasoning \citep{rudinger-etal-2020-thinking}---inferences that may be strengthened or weakened given additional evidence. The change in plausibility between an event and its abstraction can be formulated as a type of defeasible inference, and the findings of this chapter may contribute to future work in this area.

\subsection{Selectional Preference}
\label{sec:semplaus-selpref}

Modeling the plausibility of single events is also studied as \textit{selectional preference}---the semantic preference of a predicate for an argument in a particular dependency relation \citep{evens_semantic_1975, resnik_selection_1993, erk-etal-2010-flexible}; e.g., the relative preference of the verb ``breathe'' for the noun ``dentist'' as its nominal subject.

Models of selectional preference are sometimes evaluated by correlation with human judgements \citep{o-seaghdha-2010-latent,zhang-etal-2019-sp}. The primary distinction between such evaluations and those of semantic plausibility, as in this chapter, is that plausibility evaluations emphasize the importance of correctly modeling atypical yet plausible events \citep{wang-etal-2018-modeling}.

Closely related to my work are models of selectional preference that use the WordNet hierarchy to generalize co-occurrence probabilities over concepts. These include the work of \citet{resnik_selection_1993}, related WordNet-based models \citep{li-abe-1998-generalizing,clark-weir-2002-class}, and a more recent experiment by \citet{o-seaghdha-korhonen-2012-modelling} combining distributional models with WordNet. Notably, these methods make a discrete decision as to the right level of abstraction---if the most preferred subject of ``breathe'' is found to be ``person,'' for example, then all hyponyms of ``person'' will be assigned the same selectional preference score.

\subsection{Conceptual Abstraction}
\label{sec:semplaus-conceptual-abstraction}

The second method I propose can be thought of as finding the right level of abstraction at which to infer plausibility. This problem has been broadly explored by existing work.

\citet{van-durme-etal-2009-deriving} extract abstracted commonsense knowledge from text using WordNet, obtaining inferences such as ``A [\textsc{person}] can breathe.'' They achieve this by first extracting factoids and then greedily taking the WordNet synset that dominates the occurrences of factoids to be the appropriate abstraction.

\citet{gong-verb-concepts-2016} similarly abstract a verb's arguments into a set of prototypical concepts using Probase and a branch-and-bound algorithm. For a given verb and argument position, their algorithm finds a small set of concepts that has high coverage of all nouns occurring in that position.

Conceptual abstractions are believed to be captured to some extent in pre-trained language models' induced representations \citep{ravichander-etal-2020-systematicity,weir-etal-2020-probing}.

\section{Problem Formulation}

Given a vocabulary of subjects $\mathcal{S}$, verbs $\mathcal{V}$, and objects $\mathcal{O}$, let an event be represented by the s-v-o triple $e \in \mathcal{S} \times \mathcal{V} \times \mathcal{O}$.

Let $g$ be a ground-truth, total ordering of events expressed by the ordering function $g(e) > g(e')$ if and only if $e$ is more plausible than $e'$. The objective is to learn a model $f:\mathcal{S} \times \mathcal{V} \times \mathcal{O} \rightarrow \mathbf{R}$ that is monotonic with respect to $g$, i.e., $g(e)>g(e') \implies f(e)>f(e')$. This simplification follows previous work \citep{wang-etal-2018-modeling}, and the plausibility score for a given triple can be considered the relative plausibility of the event marginalized over all contexts and realizations.

While the meaning of an utterance is sensitive to small linguistic perturbations, this chapter is interested in cases where one event is more plausible than another regardless of context. For instance, the event \textit{person-breathe-air} is more plausible than \textit{thought-breathe-car} regardless of the choice of determiners or the tense of the verb.

In practice, $f$ must be learned without supervised training data, as collecting a sufficiently large dataset of human judgements is prohibitively expensive \citep{zhang-2020-aser}, and supervised models often learn dataset-specific correlations \citep{levy-etal-2015-supervised,gururangan-etal-2018-annotation,poliak-etal-2018-hypothesis,mccoy-etal-2019-right}. Therefore, the model $f$ is trained with distant supervision and evaluated by its correlation with human ratings of plausibility, which represent the ground-truth ordering $g$.

\subsection{Lexical Hierarchy}
\label{sec:semplaus-lexical-hierarchy}

Let $\mathcal{C}$ be the set of concepts in a lexical hierarchy---in this study, \textit{synsets} in WordNet---with some root concept $c^{(1)} \in \mathcal{C}$. The \textit{hypernym chain} of a concept $c^{(h)} \in \mathcal{C}$ at depth $h$ in the hierarchy is defined as the sequence of concepts $\alpha(c^{(h)}) = (c^{(1)}, c^{(2)}, \ldots, c^{(h)})$, where for all $i$, $c^{(i)}$ is a direct hypernym of $c^{(i+1)}$. A lexical hierarchy may be an acyclic graph in which concepts can have multiple hypernyms; in this case, the hypernym chain $\alpha(c^{(h)})$ is taken to be the shortest such chain to the root.

\subsection{Consistency Metrics}
\label{sec:semplaus-consistency-metrics}

Based on the intuition that plausibility estimates should be consistent across abstractions in a hypernym chain, this chapter proposes two quantitative metrics of \textit{inconsistency}, \ac{CCD} and \ac{LER}. These metrics provide insight into the degree to which a model's plausibility estimates are inconsistent.

\subsubsection{Concavity Delta}
\label{sec:semplaus-concav-delta}

For a given event, as one traverses up the hypernym chain to higher conceptual abstractions, plausibility is expected to increase until it reaches a maximally appropriate level of abstraction, and then decrease thereafter. In other words, consistent estimates are expected to be concave across a sequence of abstractions.

For example, in the sequence of abstractions ``penguin flies'' $\to$ ``bird flies'' $\to$ ``animal flies,'' plausibility first increases and then decreases. The intuition is that plausibility increases as the abstraction approaches the most appropriate conceptual level, then decreases beyond this level. A concave sequence is defined as a sequence $(a_1, a_2, a_3, ...)$ where for all $i$, $2a_i > a_{i-1} + a_{i+1}$.

Let $a_{i-1}$, $a_i$, and $a_{i+1}$ be the plausibility estimates for three sequential abstractions of an event. The \textit{divergence from concavity} is defined as:
\begin{equation*}
    \delta = 
\begin{cases} 
      \frac{1}{2}(a_{i-1} + a_{i+1}) - a_i & \text{if } 2a_i < a_{i-1} + a_{i+1} \\
      0 & \text{otherwise} \\
   \end{cases}
\end{equation*}
The \textit{Concavity Delta}, \ccd{}, is then the average $\delta$ across all triples of conceptually sequential estimates. Ideally, a model's estimates should have a low \ccd{}. A higher \ccd{} reflects the extent to which a model's estimates violate this concavity intuition. A potential limitation of this metric is that naturally longer hypernym chains will have more opportunities to diverge from concavity. Thus, one might expect longer hypernym chains to have relatively higher \ccd{} as compared to relatively short hypernym chains.

\subsubsection{Local Extremum Rate}

The \ac{LER} metric describes how often a conceptual abstraction is a local extremum in terms of its plausibility estimate. In most cases, the change in plausibility between sequential abstractions should be monotonic. For example, from ``bird flies'' $\to$ ``animal flies'' $\to$ ``organism flies,'' plausibility consistently decreases. The majority of abstractions will not be the most appropriate level and therefore should not be a local extremum.

As in \S\ref{sec:semplaus-concav-delta}, all triples of conceptually sequential estimates of the form $a_{i-1}$, $a_i$, and $a_{i+1}$ are considered. Formally, \ler{} is the number of triples where $a_i > \max(a_{i-1}, a_{i+1})$ or $a_i < \min(a_{i-1}, a_{i+1})$, divided by the total number of triples.

A high \ac{LER} signifies that plausibility estimates have few monotonic subsequences across abstractions. Therefore, a more consistent model should have a lower \ac{LER}. There are, of course, exceptions to this intuition, and this metric is most insightful when it varies greatly between models.

\section{Models}

The models considered in this chapter are all of the same general form. They take an event as input and output a relative plausibility score.

\subsection{RoBERTa}
\label{sec:semplaus-roberta}

The models I propose are structured on top of a RoBERTa baseline. I use RoBERTa in the standard sequence classification framework. An event is formatted in the raw form as \texttt{`\small [CLS] subject verb object [SEP]'} where the s-v-o triple is tokenized using a byte pair encoding.\footnote{Technically, RoBERTa's \texttt{[CLS]} and \texttt{[SEP]} tokens are \texttt{<s>} and \texttt{</s>}, respectively.} These tokens are used as input to a pre-trained RoBERTa model, and a linear layer is learned during fine-tuning to project the final-layer \texttt{[CLS]} token representation to a single logit, which is passed through a sigmoid function to obtain the final output, $f(e)$.

I use the HuggingFace Transformers library PyTorch implementation of RoBERTa-base with 16-bit floating-point precision \citep{wolf-etal-2020-transformers}.

\subsection{\conceptinject{}}
\label{sec:semplaus-concept-inject}

\conceptinject{} is an extension of existing state-of-the-art plausibility models. In addition to an event, this model takes the hypernym chains of the synsets corresponding to each argument in the event as input. I propose this model to explore how injecting a simple awareness of a lexical hierarchy affects plausibility estimates.

\conceptinject{} is similar in principle to Onto-LSTM \citep{dasigi-etal-2017-ontology}, which provides the entire hypernym chains of nouns as input to an LSTM for selectional preference, and also to K-BERT \citep{Liu_Zhou_Zhao_Wang_Ju_Deng_Wang_2020}, which injects knowledge into BERT during fine-tuning by including relations as additional tokens in the input. K-BERT has demonstrated improved performance over Chinese BERT on several NLP tasks.

The model extends the vanilla RoBERTa baseline (\S\ref{sec:semplaus-roberta}). I add an additional token embedding to RoBERTa for each synset $c \in \mathcal{C}$ and initialize the embedding of $c$ as the average embedding of the sub-tokens of $c$'s lemma.\footnote{I refer to the name of a synset as its lemma; e.g., the lemma of the synset [dog.n.01] is "dog." For synsets that correspond to multiple lemmas, one is randomly sampled.} I refer to RoBERTa's positional embedding matrix as the $x$-position and randomly initialize a second positional embedding matrix, the $y$-position.

The model input format follows that used for RoBERTa (\S\ref{sec:semplaus-roberta}), with the critical distinction that tokens for the hypernyms of the subject and object are also included as additional input.

\begin{figure*}
\centering
  \includegraphics[width=16cm,height=7cm]{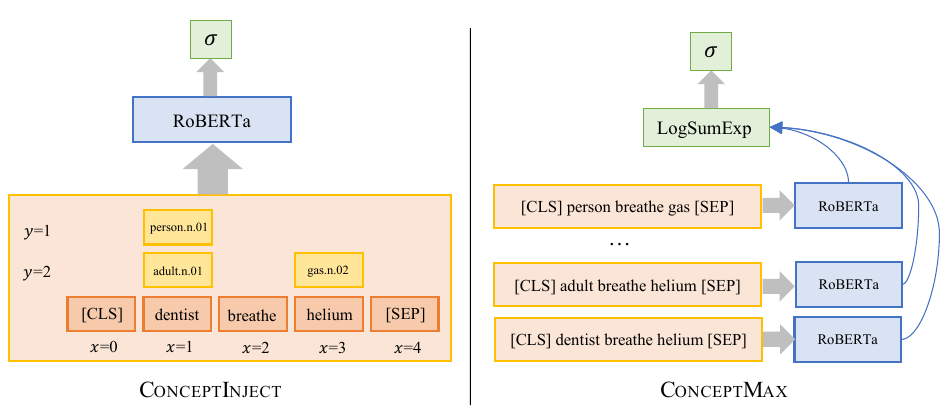}
  \caption[Proposed models for semantic plausibility inference.]{Left: The general formulation of \conceptinject{}; this model takes as input an event and the full hypernym chains of each argument. Right: \conceptmax{}, which calculates a plausibility score for each abstraction of an event using RoBERTa, and then takes the ultimate output to be the maximum of these scores. The symbol $\sigma$ represents an element-wise sigmoid function.}
  \label{fig:concept-inject}
\end{figure*}

For the subject $s$, I first disambiguate its synset $c$ using BERT-WSD \citep{yap-etal-2020-adapting}. Then, for each hypernym $c^{(i)}$ in the hypernym chain $\alpha({c})$, the token of $c^{(i)}$ is included in the model input. This token takes the same $x$-position as the first sub-token of $s$ and takes its $y$-position to be $i$, its depth in the lexical hierarchy. Finally, the $x$-position, $y$-position, and token embeddings are summed for each token to compute its initial representation (Figure~\ref{fig:concept-inject}).

The hypernyms of the object are included by the same procedure. Non-synset tokens have a $y$-position of zero. \conceptinject{} thus sees an event and the full hypernym chains of its arguments when computing a plausibility score.

\subsection{\conceptmax{}}
\label{sec:semplaus-concept-max}

\conceptmax{} is a simple post-hoc addition to the vanilla RoBERTa model (\S\ref{sec:semplaus-roberta}). I compute a score for all abstractions of an event $e$ and take the final plausibility $f(e)$ to be a soft maximum of these scores. This method is inspired by that of \citet{resnik_selection_1993}, which takes selectional preference to be a hard maximum of some plausibility measure over concepts.

Again, I use BERT-WSD to disambiguate the synset of the subject, $c^{(h)}_s$, and the synset of the object, $c^{(l)}_o$. Using RoBERTa as described in \S\ref{sec:semplaus-roberta}, I then compute a plausibility score for every triple of the form $(c^{(i)}_s,v,c^{(j)}_o)$, where $c^{(i)}_s$ and $c^{(j)}_o$ are hypernyms in the hypernym chains $\alpha(c^{(h)}_s)$ and $\alpha(c^{(l)}_o)$, respectively. Synsets are represented by their lemma when used as input to RoBERTa. Finally, I take the LogSumExp, a soft maximum, of these scores to be the ultimate output of the model (Figure \ref{fig:concept-inject}).

During training, I sample only three of the abstractions $(c^{(i)}_s,v,c^{(j)}_o)$ to reduce time complexity; thus, only four total scores need to be computed instead of $h \times l$. At inference time, I calculate plausibility with a hard maximum over all triples.

\subsection{Additional Baselines}

\paragraph{\robertazs{}} I use MLConjug\footnote{\url{https://pypi.org/project/mlconjug/}} to realize an s-v-o triple in natural language with the determiner "the" for both the subject and object, and the verb conjugated in the indicative, third-person tense; e.g., \textit{person-breathe-air} $\longrightarrow$ "The person breathes the air." I first mask both the subject and object to compute $P(o|v)$ as the LM estimate of the probability of the object occuring conditioned on the verb and a masked subject.  I then mask just the subject to compute $P(s|v,o)$. Finally, I calculate $f(e)=P(s,o|v)=P(s|v,o) \cdot P(o|v)$. In cases where a noun corresponds to multiple tokens, I mask all tokens and take the probability of the noun to be the geometric mean of its token probabilities.

\paragraph{\mlp{}} The selectional preference model of \citet{van-de-cruys-2014-neural} initialized with GloVe embeddings \citep{pennington-etal-2014-glove}.

\paragraph{n-gram} A simple baseline that estimates $P(s,o|v)$ by occurrence counts. I use a bigram model, as trigrams were found to correlate less with human judgments.
\begin{equation}
    P(s,o|v) \approx \frac{\texttt{Count}(s,v) \cdot \texttt{Count}(v,o)}{\texttt{Count}(v)^2}
\end{equation}

\begin{table}
\centering
\begin{tabular}{cc}
\toprule
$e$ & $e'$ \\
\midrule
\textit{animal-eat-seed} & \textit{animal-eat-area} \\
\addlinespace[3pt]
\textit{passenger-ride-bus} & \textit{bus-ride-bus} \\
\addlinespace[3pt]
\textit{fan-throw-fruit} & \textit{group-throw-number} \\
\addlinespace[3pt]
\textit{woman-seek-shelter} & \textit{line-seek-issue} \\
\bottomrule
\end{tabular}
    \caption[Training examples extracted from Wikipedia.]{
    Training examples extracted from Wikipedia. Event $e$ is an attested event taken to be more plausible than its random perturbation $e'$.
    }
    \label{table:wikipedia}
\end{table}

\section{Methodology: Training}
\label{sec:semplaus-training}

All models are trained with the same objective: to discriminate plausible events from less plausible ones. Given a training set $\mathcal{D}$ of event pairs $(e,e')$ where $e$ is more plausible than $e'$, I minimize the binary cross-entropy loss:
\begin{equation}
\label{eq:bce-loss}
    L = - \sum_{(e,e') \in \mathcal{D}} \log(f(e)) + \log (1-f(e'))
\end{equation}

In practice, $\mathcal{D}$ is created without supervised labels. For each ${(e,e') \in \mathcal{D}}$, $e$ is an event attested in a corpus with subject $s$, verb $v$, and object $o$. The event $e'$ is a random perturbation of $e$, uniformly of the form $(s',v,o)$, $(s,v,o')$, or $(s',v,o')$, where $s'$ and $o'$ are arguments randomly sampled from the training corpus by their occurrence frequency. This is a standard pseudo-disambiguation objective. My training procedure follows recent works that learn plausibility models with self-supervised fine-tuning \citep{kocijan-etal-2019-surprisingly,He2020OnTR,ijcai2020-554}.

For the models that use WordNet, I use a filtered set of synsets: I remove synsets with a depth of less than four, as these are too broad to provide useful generalizations \citep{van-durme-etal-2009-deriving}. I also filter out synsets whose corresponding lemma did not appear in the training corpus. These synsets were removed after sequences were formed.

The WordNet models also require sense disambiguation. I use the raw triple as input to BERT-WSD \citep{yap-etal-2020-adapting}, which outputs a probability distribution over senses, and I take the argmax to be the correct sense.

I train all models with gradient descent using an Adam optimizer, a learning rate of 2e-5, and a batch size of 128. I train for two epochs over the entire training set of examples with a linear warm-up of the learning rate over the first 10,000 iterations. Fine-tuning RoBERTa takes five hours on a single Nvidia V100 32GB GPU. Fine-tuning \conceptinject{} takes 12 hours, and \conceptmax{} takes 24 hours.

\subsection{Training Data}
\label{sec:semplaus-training-data}

I use English Wikipedia to construct the self-supervised training data. As a relatively clean, definitional corpus, plausibility models trained on Wikipedia have been shown to correlate with human judgements better than those trained on similarly sized corpora \citep{zhang-etal-2019-sp,porada-etal-2019-gorilla}.

I parse a dump of English Wikipedia using the Stanford neural dependency parser \citep{qi2018universal}. For each sentence with a direct object, no indirect object, and noun arguments (that are not proper nouns), I extract a training example $(s,v,o)$. I take $s$ and $o$ to be the lemma of the head of the respective relations (\texttt{nsubj} and \texttt{obj}), and $v$ to be the lemma of the head of the root verb. This process results in some false positives, such as the sentence ``The woman eats a hot dog.'' being extracted to the triple \textit{woman-eat-dog} (Table \ref{table:wikipedia}).

I filter out triples that never occur and those where a word occurred less than 1,000 times in its respective position. I do not extract the same triple more than 1,000 times so as not to over-sample common events. In total, I extract 3,298,396 triples (representing 538,877 unique events).

\section{Evaluation: Predicting Human Plausibility Judgements}
\label{sec:semplaus-evaluation}

I evaluate the models by their correlation with human plausibility judgements. Each dataset consists of events that have been manually labelled as plausible or implausible (Table \ref{table:human-plaus-judge}). I use AUC (area under the receiver-operating-characteristic curve) as an evaluation metric, which intuitively reflects the ability of a model to discriminate a plausible event from an implausible one.

These datasets contain plausible events that are both typical and atypical. While a distributional model should be able to discriminate typical events given that they frequently occur in text, discriminating atypical events (such as \textit{dentist-breathe-helium}) is more difficult.

\begin{table}
\centering
\begin{tabular}{ccc}
\toprule
Topic & Question & Answer \\
\midrule
cat & Does it lay eggs? & never \\
\addlinespace[3pt]
carrot & Can you eat it? & always \\
\addlinespace[3pt]
cocoon & Can it change shape? & sometimes \\
\addlinespace[3pt]
clock & Can I touch it? & always \\
\bottomrule
\end{tabular}
    \caption[Example triples from the 20 Questions commonsense dataset.]{
    Example triples from the 20 Questions commonsense dataset. These are those specific examples that contain a simple question with a single s-v-o triple and no modifiers.
    }
    \label{table:twentyquestions}
\end{table}

\begin{table}
\centering
\begin{tabular}{c@{\hskip 2em}cc}
\toprule
\multirow{4}{*}{\pep{}} &  {\it chef-bake-cookie} & \raisebox{0pt}{\small \Checkmark} \\
& {\it dog-close-door} & \raisebox{0pt}{\small \Checkmark} \\
& {\it fish-throw-elephant} & \raisebox{-2pt}{\small \XSolidBrush} \\
& {\it marker-fuse-house} & \raisebox{-2pt}{\small \XSolidBrush} \\
\addlinespace[2pt]
\hline
\addlinespace[2pt]
\multirow{4}{*}{20Q} & {\it whale-breathe-air} & \raisebox{0pt}{\small \Checkmark} \\
& {\it wolf-wear-collar} & \raisebox{0pt}{\small \Checkmark} \\
& {\it cat-hatch-egg} & \raisebox{-2pt}{\small \XSolidBrush} \\
& {\it armrest-breathe-air} & \raisebox{-2pt}{\small \XSolidBrush} \\
\bottomrule
\end{tabular}
    \caption[Representative examples taken from the validation splits of the two plausibility evaluation datasets.]{Representative examples taken from the validation splits of the two plausibility evaluation datasets, \pep{} and 20Q. For simplicity, we present human judgments as plausible ({\small \Checkmark}) or implausible (\raisebox{-1pt}{\small \XSolidBrush}). Details are provided in \S\ref{sec:semplaus-evaluation}. 
    }
    \label{table:human-plaus-judge}
\end{table}

\subsection{\pep{}}

\textsc{Pep-3k}, the crowdsourced \textbf{P}hysical \textbf{E}vent \textbf{P}lausbility ratings of \citet{wang-etal-2018-modeling}, consists of 3,062 events rated as physically plausible or implausible by five crowdsourced workers. Annotators were instructed to ignore possible metaphorical meanings of an event. I divide the dataset equally into a validation and test set following a random split which was originally proposed in a workshop version of this work \citep{porada-etal-2019-gorilla}. The motivation for including a validation set is to allow hyperparameter tuning in future work. The class distribution is equally plausible and implausible.

The evaluation relies on the assumption that all events labeled physically plausible are necessarily more plausible than all those labeled physically implausible.

\subsection{20Q}

The 20 Questions commonsense dataset\footnote{\url{https://github.com/allenai/twentyquestions}} is a collection of 20 Questions style games played by crowdsourced workers. I format this dataset as plausibility judgments of s-v-o triples similar to \pep{}.

In the game 20 Questions, there are two players---one who knows a given topic, and the other who is trying to guess this topic by asking questions that have a discrete answer. The dataset thus consists of triples of topics, questions, and answers where the answer is one of: always, usually, sometimes, rarely, or never (Table \ref{table:twentyquestions}).

I parse the dataset using the Stanford neural dependency parser \citep{qi2018universal}. I then extract questions that contain a simple s-v-o triple with no modifiers where either the subject or object is a third person singular pronoun. We replace this pronoun with the topic, and otherwise replace any occurrence of a personal pronoun with the word ``person.'' I filter out examples where only two of three annotators labelled the likelihood as never. Finally, I take events labelled ``never'' to be less plausible than all other events. This process results in 5,096 examples equally divided between plausible and implausible. I split examples randomly into equal sized validation and test sets.

\begin{table}
\centering
\begin{tabular}{@{\extracolsep{0mm}}
                @{}l
                @{\hspace{5mm}} c
                @{\hspace{5mm}} c
                @{\hspace{7mm}} c @{}}
\toprule   
Model & \pep{} & 20Q & Avg. \\
\midrule
n-gram & .51 & .52 & .52 \\
\mlp{} & .55 & .52 & .53 \\
\robertazs{} & .56 & .57 & .56 \\
RoBERTa & .64 & .67 & .66 \\
\midrule
\conceptinject{} & .64 & .66 & .65 \\
\conceptmax{} & \textbf{.67} & \textbf{.74} & \textbf{.70} \\
\bottomrule
\end{tabular}
\caption[Test set results for predicting human plausibility judgements.]{Test set results for predicting human plausibility judgements. Performance is evaluated with AUC with respect to the ground-truth, manually labeled plausibility ratings.}
\label{table:results}
\end{table}

\subsection{Quantitative Results}

Despite making a discrete decision about the right level of abstraction, \conceptmax{} achieves a higher AUC score on both evaluation sets compared to \conceptinject{} and the vanilla RoBERTa baseline (Table~\ref{table:results}). The fact that the \conceptmax{} model aligns more closely with human judgments than the baselines supports the hypothesis that enforcing conceptual consistency improves plausibility estimates.

\conceptinject{} performs similarly to the RoBERTa baseline, even though this model is aware of the WordNet hierarchy. My hypothesis is that the self-supervised learning signal does not incentivize the use of this hierarchical information in a way that would increase correlation with human plausibility judgements. However, I do find that \conceptinject{} attends to the hypernym chain by qualitatively observing its self-attention weights.

All fine-tuned RoBERTa models correlate better with plausibility judgements than the \robertazs{} baseline. The n-gram baseline performs close to random, which is perhaps to be expected, as very few of the evaluation triples occur in the Wikipedia training data.

\subsection{Qualitative Analysis}

To better understand the performance of these models, I manually inspected 100 examples from each dataset. I found that RoBERTa rarely assigns a high score to a nonsensical event, although this did occur in five cases (e.g., \textit{turtle-climb-wind} and \textit{person-throw-library}). RoBERTa also rarely assigns a low score to a seemingly typical event, although this was somewhat more common (in cases such as \textit{kid-use-handbag} and \textit{basket-hold-clothes}, for example). This finding confirms the expectation that discerning the typical and the nonsensical should be relatively easy for a distributional model.

Examples not at the extremes of plausibility are harder to categorize; however, one common failure mode appears when the plausibility of an event hinges on the relative size of the subject and object, such as in the case of \textit{dog-throw-whale}. This observation is similar to the limitations of static word embeddings noted by \citet{wang-etal-2018-modeling}.

\begin{figure*}%
    \centering
    \subfloat[\centering RoBERTa]{{\includegraphics[width=5cm]{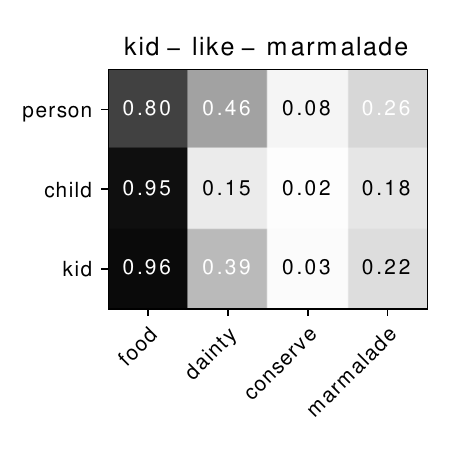} }}%
    \subfloat[\centering \conceptinject{}]{{\includegraphics[width=5cm]{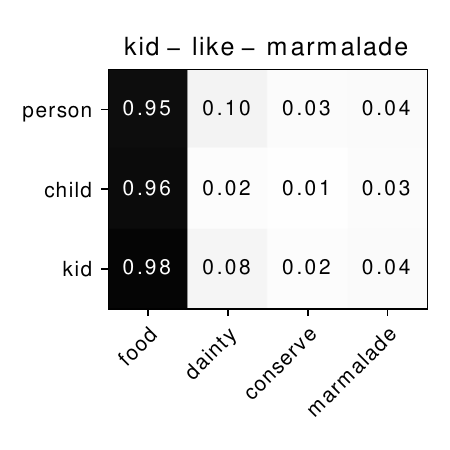}}}
    \subfloat[\centering \conceptmax{}]{{\includegraphics[width=5cm]{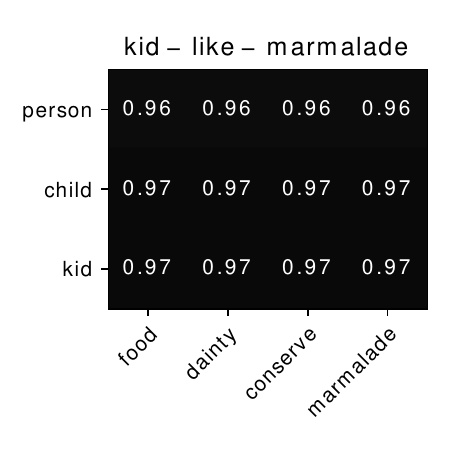}}}%
    \caption{Outputs across conceptual abstractions for the event \textit{kid-like-marmalade} from the 20Q dataset. This event is taken to be relatively plausible as the ground-truth label was ``usually.''}%
    \label{fig:roberta-results}%
\end{figure*}

\section{Research Findings: Consistency Evaluation}
\label{sec:semplaus-consistency}

For every event $e$ in the evaluation sets of human plausibility judgments (\S\ref{sec:semplaus-evaluation}),  $e$ is disambiguated using BERT-WSD and then calculate models' estimates for the plausibility of every possible abstraction of $e$ (Figure \ref{fig:roberta-results}). Based on these estimates, one can analyze the consistency of each model across abstractions.

\begin{table}
\centering
\begin{tabular}{@{\extracolsep{0mm}}
                @{}l
                @{\hspace{5mm}} c
                @{\hspace{\lengtha}} c
                @{\hspace{\lengthb}} c
                @{\hspace{\lengtha}} c @{}}
\toprule   
{} & \multicolumn{2}{c@{\hspace{\lengthc}}}{\pep{}} & \multicolumn{2}{c}{20Q} \\
 \cmidrule(r{\lengthb}){2-3}
 \cmidrule(){4-5}
Model & \ccd{} & \ler{} & \ccd{} & \ler{} \\ 
\midrule
n-gram & .06 & .50 & .07 & .50 \\
\mlp{} & .03 & .61 & .03 & .49 \\
\robertazs{} & .13 & .70 & .12 & .65 \\
RoBERTa & .09 & .52 & .08 & .51 \\
\midrule
\conceptinject{} & .08 & .52 & .07 & .51 \\
\conceptmax{} & .02 & .00 & .02 & .00 \\
\bottomrule
\end{tabular}
\caption[Evaluation of inconsistency for predicting semantic plausibility.]{Evaluation of inconsistency. \ccd{} describes the degree to which sequences of estimates across a hypernym chain diverge from a concave sequence. \ler{} describes how often conceptual abstractions are local extrema with respect to plausibility.}
\label{table:inconsistency}
\end{table}

\subsection{Quantitative Results}

I use the proposed metrics of consistency (\S\ref{sec:semplaus-consistency-metrics}) to evaluate the extent to which each model's estimates are consistent across a hypernym chain (Table \ref{table:inconsistency}).

The \robertazs{} model, which correlates with plausibility the least of the RoBERTa-based models, has by far the highest inconsistency.

The fine-tuned RoBERTa and \conceptinject{} estimates are also largely inconsistent according to my metrics. For these models, half of all estimates are a local extremum in the lexical hierarchy. As shown in Figure \ref{fig:roberta-results}, the space of plausibility estimates is rigid for these models, and most estimates are a local extremum with respect to the plausibility of the subject or object of the event.

By contrast, \conceptmax{} is almost entirely consistent by these metrics, which is to be expected as this model explicitly uses the same WordNet hierarchy that I am using for evaluation. I also evaluated consistency using the longest rather than the shortest hypernym chain in WordNet but did not find a significant change in the results. This is likely because for the consistency evaluation, I am using the hypernym chains that were filtered as described in \S\ref{sec:semplaus-lexical-hierarchy}.

\subsection{Qualitative Observations}

I qualitatively evaluate the consistency of the models by observing the matrix of plausibility estimates for all abstractions, as shown in Figure \ref{fig:roberta-results}.

In agreement with the quantitative metrics, I observe that RoBERTa estimates are often inconsistent, varying greatly between two abstractions that should have similar plausibility. Surprisingly, however, it is also often the case that RoBERTa estimates are similar or identical between abstractions. In some cases, this may be the result of the model being invariant to the subject or object of a given event.

I also observe the individual examples with the highest \ccd{}. In these cases, it does appear that the variance of the model's estimates is unreasonable. In contrast, the \ac{LER} metric is sometimes high for an example where the estimates are reasonably consistent. This highlights a limitation of the \ac{LER} metric: it does not take into account the degree of change between estimates.

Finally, I observe that the sense chosen by BERT-WSD is often different from what a human annotator, primed to rate plausibility, would assume. For example, in the case of \textit{dog-cook-turkey}, BERT-WSD takes "dog" to be a hyponym of "person" (as in, a contemptible person). While this is a reasonable sense disambiguation in some contexts, it results in a different plausibility profile than the one annotated by humans, who likely assumed the animal sense of "dog."

\section{Discussion and Conclusion}

The results of this chapter demonstrate that language model-based approaches to modeling semantic plausibility---a necessary component of resolving coreference---are not necessarily consistent or generalizable. The models are sensitive to perturbations in ways that one would not expect to interfere with a human's judgment, particularly when generalizing across a semantic hierarchy.

While the state-of-the-art in modeling plausibility has improved in recent years, current models still fall short of human ability. This chapter has shown that model estimates are inconsistent with respect to a lexical hierarchy: they correlate less with human judgments compared to model estimates that are forced to be consistent, and they do not satisfy the intuitively-defined quantitative measures of consistency proposed herein. Furthermore, simply injecting lexical knowledge into a model is not sufficient to correct this limitation. Conceptual consistency appears to require a more discrete, hierarchical bias.

These findings suggest several interesting questions for future work, including: 1) can a \textit{non-monotonic}, yet conceptually consistent model of plausibility be designed that better correlates with human judgments? and 2) can a hierarchy of abstractions be induced from data rather than relying on a manually created lexical hierarchy like WordNet?

More recent work has evaluated larger scale language models, such as GPT-3.5, on the two plausibility datasets studied and found among other conclusions that larger-scale model accuracy may in certain cases surpass human agreement~\citep{guo2023evaluatinglargelanguagemodels,gupta-etal-2023-editing}. Given the limited scope of the datasets, however, this could motivate further data curation to clarify the scope of these findings.

\paragraph{Takeaway.} This chapter provides a bookend to the investigations of the preceding chapters. The thesis began by highlighting a lack of convergent validity between ``challenge set'' and ``canonical'' coreference evaluations and then explored issues of contestedness and discriminant validity within those canonical evaluations. Those chapters demonstrated that conclusions from broad, coreference-based evaluations are often limited by their design.

The study of semantic plausibility is not isolated to these narrow cases; it is foundational to resolving many instances of coreference. Consider again, for instance, the earlier mentioned Winograd schema: ``The trophy would not fit in the brown suitcase because it was too big.'' A human reader most typically resolves `it' to `the trophy' because the resulting proposition---that the trophy's large size is the reason it will not fit---is the most plausible interpretation. Changing a single word to ``small'' flips the antecedent to `the suitcase,' again based on which interpretation is more plausible. This reliance on plausibility highlights why the conceptual inconsistencies explored in this chapter are so critical. If a model's inferences of plausibility is fragile—as demonstrated by its inconsistent judgments across a semantic hierarchy—then its ability to reliably perform the commonsense reasoning needed for coreference resolution is expected to also be limited.

In contrast, this final experimental chapter has shown that a more controlled evaluation setting can overcome some of these measurement validity issues. By precisely defining the construct being measured (semantic plausibility) and narrowing the experimental scope, the findings here do not suffer from the same validity concerns to the same extent in that results are consistent across datasets. However, this more controlled setting reveals a different facet of the thesis argument: it provides formal evidence that large language models remain fragile, exhibiting poor generalization to varied contexts. The conceptual inconsistency demonstrated in this chapter is a clear example of this fragility, supporting the conclusion that high performance on some benchmarks does not necessarily indicate robust, human-like reasoning.

\chapter{Discussion}
\label{chap:discussion}

\myepigraph{All language use can be thought of as a way of activating procedures \\ within the hearer.\vspace{10pt}}{\citet{winograd_1972}}

Having presented the detailed empirical evidence of this thesis across four core experimental chapters (Chapters~\ref{chap:challenge-set-assumption} to \ref{chap:semantic-plausibility}), this final discussion now steps back to interpret these findings and explore their broader implications. The preceding chapters considered the technical specifics of coreference-based evaluations, analyzing issues of measurement validity and the performance of modern language models. The purpose of this chapter is to synthesize these results into a coherent narrative, making the central arguments of this thesis accessible to a wider audience, including those who may not be specialists in natural language processing.

The fundamental questions this thesis has sought to address are: What can we reasonably conclude from the ways we currently evaluate systems' abilities to make inferences based on the meaning encoded by language? And, what do these evaluations tell us about the capabilities of the dominant systems in \ac{NLP} today, namely large language models? This discussion will unpack the answers that have emerged, bridging the gap between the technical contributions of this work and their significance for the future of the field in its goal of creating systems capable of performing complex tasks believed to require natural language understanding.

\section{Revisiting the Central Argument}

The thesis statement put forward in the introduction is composed of two interconnected claims: first, that conclusions drawn from coreference-based evaluations are often limited by their design, and second, that in a more controlled settings, related evaluations provide evidence that large language models are simultaneously powerful and yet fragile. By this I mean to say, somewhat more formally, that these models are improvements over previously systems based on standard evaluation metrics but are still incapable of making inferences human readers of natural language are clearly capable of. The experimental work of this thesis has substantiated both claims. Below, I revisit how the findings from each chapter build this argument.

\subsection{The Limits of Evaluation}

The first major contribution of this thesis has been to demonstrate that our measurement tools---the evaluations themselves---are often not as reliable or straightforward as we might assume. By applying the framework of \textit{measurement validity} from the social sciences, this work has systematically identified critical weaknesses in canonical coreference evaluations. These are not minor technicalities; they are fundamental issues that constrain the generalizability of our findings and can lead to misleading conclusions about model progress. This is not to say that there was no prior awareness of these issues; quite the contrary notable examples include \citet{recasens2014coreference} and \citet{can_we_fix} who have argued for additional clarity in defining the scope of coreference resolution evaluations. That being said, I believe the framework of measurement validity provides a useful vocabulary, previously not applied to the area, that will continue to help to make progress in developing more meaningful evaluations.

\paragraph{Contestedness}
The experiments in Chapter~\ref{chap:measurement} brought a foundational issue to the forefront: coreference is a \textit{contested construct}. There is no single, universally accepted definition of what counts as a coreferential relationship. As shown, datasets like OntoNotes, PreCo, and OntoGUM, while all ostensibly annotated for ``coreference,'' operate under subtly different conceptualizations. For instance, the treatment of generic nouns (\textit{e.g.,} ``trees'' in general vs. specific trees) or predicative structures (\textit{e.g.,} ``\textit{He} is \textit{a doctor}'') varies significantly.

This is not merely a minor detail. Rather, as the disaggregated results in Chapter~\ref{chap:measurement} demonstrated, a state-of-the-art model trained on one definition (OntoNotes) systematically fails on phenomena that are central to another definition (PreCo). This means that a model's poor ``generalization'' score may not reflect an inability to reason about language, but rather its adherence to the specific definition of the task it was trained on. Without a clear understanding of the construct being measured, aggregated F1 scores risk conflating a model's ability to generalize with its ability to adapt to a different, unstated task definition. This confirms the first part of the thesis statement: the conclusions are limited by the design choices of the evaluation itself.

\paragraph{Convergent Validity}
A second, related issue is the lack of \textit{convergent validity}. If different methods for measuring the same underlying ability produce conflicting results, our confidence in any single result must be diminished. This was the central finding of Chapter~\ref{chap:challenge-set-assumption}. We saw that the landscape of coreference evaluation is split between broad, \textit{canonical} evaluations on naturally occurring text (like OntoNotes) and narrow, \textit{challenge set} evaluations designed to test specific reasoning skills (like the Winograd Schema Challenge).

Intuitively, one might assume that a model that excels at the difficult, reasoning-intensive challenge sets would also perform well on the more general, canonical task. Our findings showed this assumption to be false. Prompted large language models, which have effectively ``solved'' many WSC-like benchmarks, were shown to be less accurate than supervised, task-specific systems on certain types of attested pronouns in canonical datasets. The relative ranking of systems changes depending on the evaluation chosen. This divergence is a classic sign of poor convergent validity. It suggests that these two styles of evaluation, while both related to coreference, are measuring different, though overlapping, capabilities. High performance on a challenge set cannot be interpreted as a sign of having solved the broader problem, further reinforcing the limitations of our evaluation practices.

\paragraph{Discriminant Validity}
Finally, Chapter~\ref{chap:supervised-comparison} highlighted a failure of \textit{discriminant validity}---the concern that an evaluation may be measuring confounding factors rather than the specific capability it purports to measure. The recent history of coreference resolution research has seen a succession of new model architectures reporting state-of-the-art results. The implicit claim is that these improvements are due to superior architectural design.

However, our controlled experiments revealed that when the underlying pretrained language model is held constant, much of the reported progress vanishes. The performance gap between a five-year-old architecture (C2F) and the most recent models narrowed to less than a point of F1 score, and the older model even generalized better to out-of-domain genres. This indicates that the evaluations were not necessarily only measuring architectural improvements; they were possibly confounded by the ever-increasing power of the base language models. Progress attributed to clever model design was possibly an artifact of using a better off-the-shelf encoder. This finding illustrates how conclusions about \textit{why} a model succeeds can be restricted by an experimental setup that fails to isolate the variable of interest.

\subsection{The Fragility of Language Models}

While the first half of the thesis argument focused on the limitations of our evaluations, the second half pivots to what we can learn when we design evaluations to mitigate some of these issues. By moving to a more controlled setting, we can begin to probe the nature of the reasoning capabilities of large language models. The findings here support the second part of the thesis statement: LLMs achieve high performance but remain fragile, failing to generalize in ways that align with human intuition.

Chapter~\ref{chap:semantic-plausibility} provided the clearest evidence for this fragility. The chapter moved away from the contested, broad construct of ``coreference'' to a more tightly defined, necessary component of the phenomenon: \textit{inferring the plausibility of events}. As demonstrated with the recurring Winogrand schema example, resolving a pronoun often requires judging which potential referent would result in a more plausible state of affairs. This more controlled evaluation setting avoids the definitional ambiguity of the canonical tasks and allows for a more focused test of semantic consistency.

The key finding was that language models exhibit profound \textit{conceptual inconsistency}. A model might correctly infer that ``a person breathing'' is a plausible event but simultaneously infer that ``a dentist breathing'' is implausible. From a human perspective, this is a failure of basic generalization. A dentist is a type of person, and the property of breathing should generalize from the superclass to the subclass. The fact that state-of-the-art models fail at this level of consistency reveals a fundamental brittleness. Their knowledge, derived from statistical patterns in text, does not appear to be organized along the same systematic, hierarchical lines as human conceptual knowledge.

In analysis-oriented exmperiments, this property I have described as fragility appeared not easily fixed. Simply providing the model with explicit hierarchical knowledge (the \texttt{ConceptInject} model) did not resolve the inconsistency. Only a post-hoc method that explicitly forced consistency by reasoning over the hierarchy (\texttt{ConceptMax}) improved both consistency and correlation with human judgments. (And the strong constraints of this approach might render it not useful for tasks more broadly.) This suggests that the models' internal representations are not conducive to this type of systematic inferences, and that their high performance on some benchmarks may be masking these deeper failures of generalization. This finding rounds out the overall argument of the thesis: when we use a possibly more conceptually valid, controlled evaluation, we find evidence that LLMs, despite their successes, lack a key component of robust, human-like understanding.

\section{Broader Implications for Natural Language Processing}

The findings of this thesis, while rooted in the specific domain of coreference resolution, have significant implications for the wider field of \ac{NLP}. They call for a re-evaluation of how we measure progress, how we build models, and how we conceptualize the very goal of ``language understanding.''

\subsection{Measurement-Aware Evaluation}
The most direct implication is that the \ac{NLP} community must become more sophisticated in its approach to evaluation. The ``leaderboard-driven'' paradigm, where progress is measured by a single aggregate score on a benchmark dataset, is insufficient and can be misleading. This thesis advocates for a \textit{measurement-aware} approach to evaluation, which involves several shifts in practice:

\begin{enumerate}
    \item \textbf{Explicit Construct Definition:} Researchers proposing new tasks or datasets should be explicit about how they are defining the theoretical construct they intend to measure. As seen with coreference, ambiguity here is a possible source of conclusions not generalizing in possibly expected ways.
    \item \textbf{Disaggregated Analysis:} Instead of relying solely on a single F1 score, evaluations should include disaggregated results on different phenomena, genres, or subsets of the data. As shown in Chapter~\ref{chap:measurement}, this is crucial for understanding \textit{why} a model fails to generalize.
    \item \textbf{Controlled Comparisons:} When comparing models, particularly architectures, it is essential to control for confounding variables like the size and type of the base language model. Chapter~\ref{chap:supervised-comparison} serves as a blueprint for how such controlled studies can yield more reliable conclusions about what drives performance.
    \item \textbf{Testing for Convergent Validity:} Progress on one benchmark should not be considered in isolation. The field should actively seek to evaluate models across multiple, diverse evaluations of the same underlying construct to test whether findings converge. Divergence, as seen in Chapter~\ref{chap:challenge-set-assumption}, is a critical signal that our understanding of the task is incomplete.
\end{enumerate}

\section{Limitations of This Thesis}

No single work can be exhaustive, and it is important to acknowledge the boundaries and limitations of this thesis to situate its contributions properly. The experiments presented are subject to their own constraints which affect the extent to which any conclusions might apply more broadly. This section aims to make the limitations of which I am aware more explicit.

First, the empirical scope, while broad, is necessarily constrained. The experiments focused exclusively on the English language. A significant limitation is that this scope precludes analysis of phenomena more prominent outside of English (\textit{e.g.,} zero-anaphors) or that do not exist in English (\textit{e.g.,} switch reference wherein morphological markers indicate whether a subject is shared across clauses, and obviation, which grammatically distinguishes between multiple third-person referents based on discourse prominence.). The dynamics of coreference and the nature of measurement validity may differ substantially in morphologically richer or typologically different languages. Within English, the investigation focused on a limited formulation of pronominal coreference resolution; one could expand upon these results by considering additional proform expressions (\textit{e.g.,} first and second person or reflexive pronouns) or by more explicitly distinguishing identity coreference from related phenomena such as bridging. Furthermore, the analyses were centered on a specific set of widely-used datasets (OntoNotes, Winogrande, etc.) and models (Llama, DeBERTa, etc.). While representative of the state of the art, the findings may not generalize to all possible models or data regimes.

Second, the framework of measurement validity provides a powerful lens for critique but does not offer a simple prescription for an ideal evaluation. The empirical study in this thesis only establishes \textit{correlations} between model performance and types of coreference; it does not prove causation. While this work focused on a select few types of coreference to study their interaction with dataset operationalization in detail, future work could explore other dimensions such as data processing choices (\textit{e.g.,} tokenization), task format, and the annotation procedure itself, including noise and ambiguity. A promising direction might be to define such differences using grounded theory \citep{strauss1997grounded}. A key takeaway is that a holistic understanding of how coreference is operationalized is necessary to guide future modeling and evaluation decisions. It is not as simple as taking the intersection of all coreference types annotated in existing corpora; such an approach does not solve the fundamental problem of needing to clearly define the construct being measured. Ultimately, the process of defining a construct and ensuring validity is an ongoing, iterative process for a research community, not a problem that can be definitively solved. This thesis provides a blueprint and a vocabulary for this process, but the work itself is far from complete.

Finally, the novel evaluation in Chapter~\ref{chap:semantic-plausibility}, while more controlled, is still a proxy for the broader complexity of human reasoning. Semantic plausibility is itself a complex construct, and our metrics of conceptual consistency (\texttt{CCD} and \texttt{LER}) are based on specific intuitions about how plausibility should behave across a hierarchy. While these metrics proved effective at revealing model fragility, they are not an exhaustive definition of systematic reasoning. They capture one important aspect of it, but human cognition is undoubtedly more multifaceted. The findings from this chapter, therefore, point toward a specific type of model brittleness but should be interpreted as one piece of a much larger puzzle.

\section{Future Directions}

The findings and limitations of this work point toward several promising avenues for future research that can build upon its foundations. Possible directions might include:

\textbf{Applying the Measurement Modeling Framework to Other NLP Tasks:} Coreference is not the only NLP task with a contested construct. Areas like natural language inference, question answering, and summarization could all benefit from a systematic analysis through the lens of measurement validity.

\textbf{Iterative Model and Evaluation Co-development:} The traditional pipeline where a static dataset is created and models then compete on it for years is part of the problem. A future paradigm might involve a more dynamic, adversarial process where new evaluation suites are continuously developed to expose the failures of current models, thereby driving the development of more robust systems in a virtuous cycle.

\section{Ethics Statement}

We have focused on evaluation datasets of CR, but we did not quantify possible downstream implications of these findings. Different types of CR might also be more prevalent in certain types of corpora that might be about or written by minoritized groups, or that might cover sensitive topics.  
CR models have been shown to exhibit biases, inferring coreferences disparately for distinct social groups \citep{webster-etal-2018-mind,Kocijan_Camburu_Lukasiewicz_2021,hossain-etal-2023-misgendered}. Similarly, CR and other NLP datasets are also known to contain biased, stereotypical, or in other way problematic context~\citep{cao2021toward,selvam2022tail}.

PCR systems are known to perform disparately on subgroups which has ethical implications particularly for potential real-world use cases~\citep{zhao-etal-2018-gender,rudinger-etal-2018-gender,webster-etal-2018-mind,hossain-etal-2023-misgendered}. We therefore do not recommend or endorse the use of these systems for downstream purposes such as real-world, commercial applications; rather, our experiments are focused solely on the validity of certain assumptions of existing datasets.

\section{Thoughts on the role of coreference-based evaluation}

Why do we focus on coreference when there are many important aspects related to understanding natural language? I hope to further motivation this scope within this section.

On the one hand, one might view natural language as corresponding to a formal logic \citep{wittgenstein2023tractatus,richard1970english} in which coreferences can be explicitly represented as, for example, predicates in Lambda calculus \citep{bekki2014representing}; however, this formulation can be seen as restrictive, as it assumes that the full diversity of natural language semantics can be captured within such a formal system.

\begin{figure}[t]
    \centering
    \includegraphics[width=15cm]{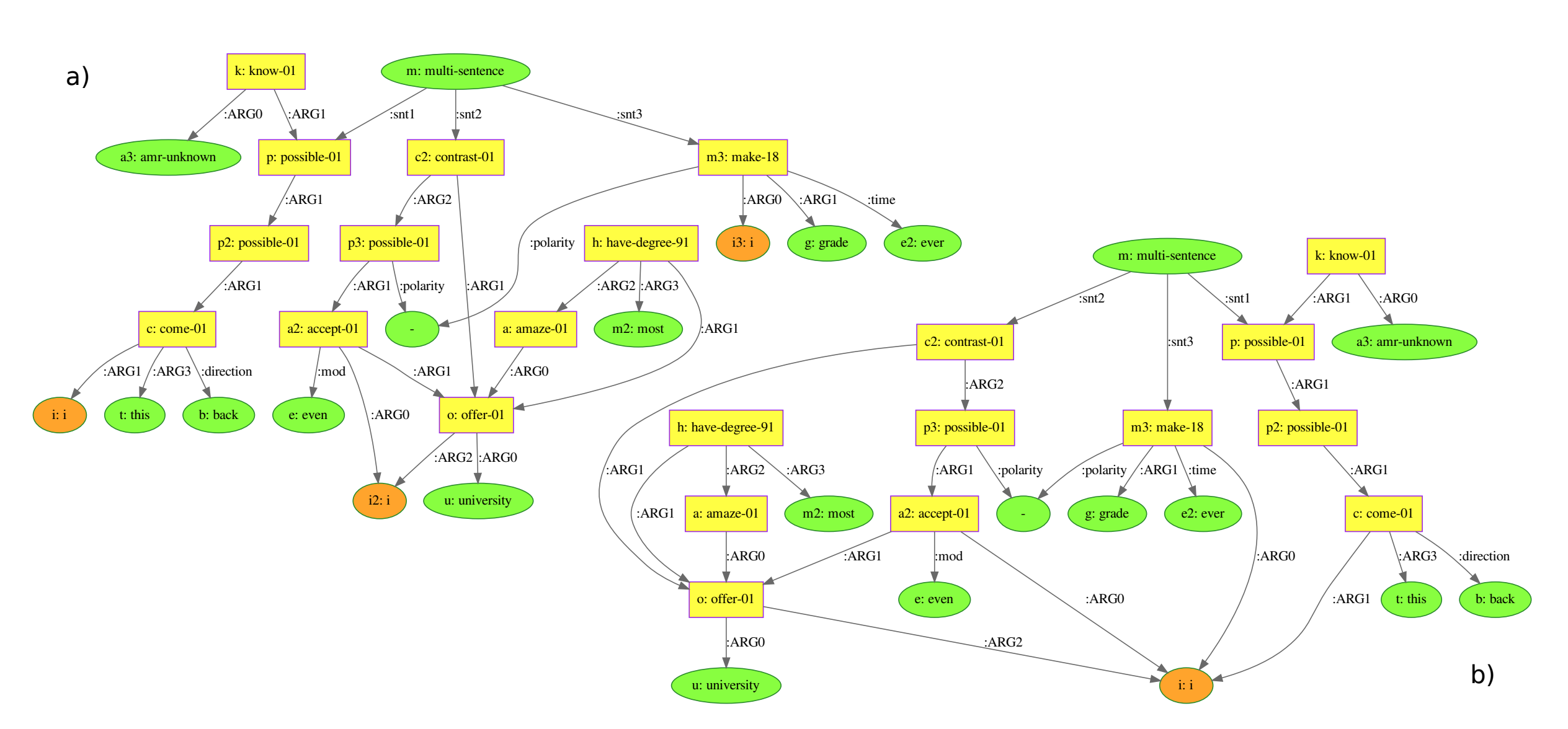}
    \caption[An example instance of an Abstract Meaning Representation (AMR).]{An example instance of an Abstract Meaning Representation (AMR) including annotated coreference relations. The original text for which this AMR was built is ``\textit{Maybe I can come back from this, who knows. I’ve got the most amazing university offers, but I can’t even accept them - I’ll never make the grades.}'' Reproduced from \citet{anikina-etal-2020-predicting}. (Creative Commons Attribution 4.0 International License) \vspace{1em}}
    \label{fig:amr-parse}
\end{figure}

The \ac{AMR}~\citep{banarescu-etal-2013-abstract} was developed in part to support practical NLP applications such as machine translation by providing a somewhat more flexible formalization of natural language semantics---and can encode coreferences explicitly \citep{anikina-etal-2020-predicting}; see Figure~\ref{fig:amr-parse} for an example. However, AMR itself must encode coreferences and thus cannot side-step the issue of CR while also introducing a particular formulation. Meanwhile, coreference-based evaluations can be relatively agnostic to the representation of meaning itself.

\subsection{Coreference-based evaluation as behavioral evaluation}

The field of natural language processing has mostly taken a behavioral approach to the study of meaning and semantics~\citep[e.g.,][]{nilsson1998artificial,potts2022could} as opposed to arguments based on innateness and intrinsic properties of a model that one could possibly use to reason about systems' capabilities~\citep{chomsky-aspects-1965,putnam1967innateness}. (In other words, the study of representing the meaning of language, popularly referred to as ``natural language understanding,'' has largely focused on the exclusive ability of a system to perform end-to-end tasks that are believed to require understanding. See \S1.2.8 of \citet{russell2016artificial} for a brief history of this distinction in the field.)

Following from this behavioral perspective, the capacity of language models to encode the meaning of language is commonly studied and described in terms of ``understanding'' the meaning of language~\citep[\textit{i.a.}]{efrat2020turking,jiang-etal-2021-know,merrill-etal-2021-provable,webson-pavlick-2022-prompt,y2022large}. But in this thesis, I am specifically focusing on the capacity for language models or any system to make inferences based on representations that somehow encode the semantics of language.

\section{Takeaways}

The literal meaning encoded by natural language, broadly referred to as semantics, is an unobservable theoretical concept that cannot be directly measured. How semantics can be represented and inferred has long been the subject of philosophical debate (the details of which are beyond the scope of this thesis, but see \citet{sep-meaning} for a general survey). Among existing approaches, computational methods for inferring semantic representations have the desirable property that they can be empirically implemented and evaluated~\citep{katz-and-fodor-semantics,charniak1976computational,blackburn2005representation}. These methods operationalize aspects of meaning in ways that allow them to be tested, refined, and ultimately deployed in practical NLP applications.

Within this computational perspective, coreference resolution (CR) emerges as an implicit but necessary component of representing the semantics of natural language. This is because resolving coreference is essential for maintaining coherent representations of meaning across spans of text—a requirement for any NLP task that aims to model semantics in a meaningful way. Consequently, CR is necessary in all NLP tasks that require semantic modeling, including applications such as machine translation, question answering, and text summarization.

Furthermore, CR models are commonly used as downstream components in practical systems for tasks such as information extraction \citep{huang-etal-2009-solving} and dialogue systems \citep{xu-choi-2022-online}. In these settings, the ability to correctly identify which expressions refer to the same entities or events is crucial for extracting structured knowledge from text and for maintaining coherent and contextually appropriate interactions with users.

\chapter{Conclusion}
\label{chap:conclusion}

\myepigraph{Natural language understanding is not solved, and it is such a hard problem that we need every tool in the toolbox. We can’t afford to throw anything away.\vspace{10pt}}{\citet{palmer-2024-big}}

This thesis began by questioning what we can truly learn from our evaluations of a system's ability to resolve coreference. For decades, coreference resolution has served as a core task in NLP used for assessing progress in natural language understanding, offering a practical way to approach the evaluation of the unobservable process of semantic interpretation. Yet, as this work has argued, the clarity offered by this approach is often obscured by our evaluations themselves, with their implicit assumptions and design limitations, can restrict and sometimes distort the conclusions we are able to draw.

Through a broad investigation spanning canonical benchmarks, challenge sets, and novel controlled experiments, a clear, two-part picture has emerged. First, this thesis demonstrated that current evaluation practices for coreference resolution suffer from significant issues of measurement validity. The very construct of coreference is contested across datasets, leading to a lack of convergent validity where model rankings are inconsistent. Furthermore, poor discriminant validity often confounds progress, attributing performance gains to architectural novelty when they may stem from the scale of the underlying language model. These findings confirm the first part of the thesis statement: conclusions drawn from many coreference-based evaluations are indeed limited in their generalizability.

Having established the limitations of our measurement tools, this thesis then used a more controlled and rigorously defined evaluation setting to probe the capabilities of state-of-the-art systems. This led to the confirmation of the second part of the thesis statement, revealing the second main point: modern large language models are both powerful and fragile. They are impressively capable of learning the statistical patterns present in natural language corpora and can be used to achieve relatively high scores on standard benchmarks. Yet, they lack the systematic, robust generalization that is possibly a hallmark of human linguistic competence. Specifically, their failure in the studied cases to maintain conceptual consistency when inferring event plausibility is a demonstrative example of this fragility, showing that relative benchmark improvements do not necessarily imply human-like understanding.

The ultimate conclusion of this thesis is a direction for greater clarity in how we as a field measure progress. By understanding the limitations of both our models and our methods for evaluating them, we can chart a more effective path forward. The aspirational goal remains the same: to build machines that can genuinely comprehend and reason with natural language. This thesis has argued that achieving this goal will require not only more powerful models but also better measurements.

\bibliography{references,anthology}
\bibliographystyle{iclr2020_conference}

\end{document}